\documentclass[preprint,authoryear,3p]{elsarticle}

\usepackage{graphicx}

\usepackage{amssymb,amsthm,amsmath}
\usepackage{subfig}
\usepackage{float}
\usepackage{multirow}
\usepackage{longtable,lscape}
\usepackage{cases}
\usepackage{url}
\usepackage{enumitem}
\usepackage{diagbox}

\usepackage[usenames, dvipsnames]{color}

\usepackage{fancyhdr}
\theoremstyle {plain}

\theoremstyle{definition}
\newtheorem{defn}{Definition}[section]

\theoremstyle{remark}
\newtheorem{rem}{Remark}[section]
\usepackage{titlesec}
\usepackage{float}
\titleformat{\paragraph}
{\normalfont\normalsize\bfseries}{\theparagraph}{1em}{}
\titlespacing*{\paragraph}
{0pt}{3.25ex plus 1ex minus .2ex}{1.5ex plus .2ex}

\numberwithin{equation}{section}

\journal{Transportation Research Part C}

\begin{document}

\begin{frontmatter}



\title{A Survey on Autonomous Vehicle Control in the Era of Mixed-Autonomy: From Physics-Based to AI-Guided Driving Policy Learning}


\author[cu,dsi]{Xuan Di\corref{cor}}
\ead{sharon.di@columbia.edu}
\author[cu]{Rongye Shi}

\cortext[cor]{Corresponding author. Tel.: +1 212 853 0435;}

\address[cu]{Department of Civil Engineering and Engineering Mechanics, Columbia University}
\address[dsi]{Center for Smart Cities, Data Science Institute, Columbia University}

\begin{abstract}
    This paper serves as an introduction and overview of the potentially useful models and methodologies from \textit{artificial intelligence} (AI) into the field of \textit{transportation engineering} for autonomous vehicle (AV) control in the era of mixed autonomy. 
It is the first-of-its-kind survey paper to comprehensively review literature in both transportation engineering and AI for mixed traffic modeling. 
We will discuss state-of-the-art
applications of AI-guided methods, 
identify opportunities and obstacles, 
raise open questions, 
and 
help suggest the building blocks and areas where AI could play a role in mixed autonomy. 
We divide the stage of autonomous vehicle (AV) deployment into four phases: the pure HVs, the HV-dominated, the AV-dominated, and the pure AVs. 
This paper is primarily focused on the latter three phases. 
Models used for each phase are summarized, encompassing game theory, deep (reinforcement) learning, and imitation learning. 
While reviewing the methodologies, we primarily focus on the following research questions: 
(1) What scalable driving policies are to control a large number of AVs in mixed traffic comprised of human drivers and uncontrollable AVs? 
(2) How do we estimate human driver behaviors? 
(3) How should the driving behavior of uncontrollable AVs be modeled in the environment? 
(4) How are the interactions between human drivers and autonomous vehicles characterized? 
Hopefully this paper will not only 
inspire our transportation community to rethink the conventional models that are developed in the data-shortage era, 
but also reach out to other disciplines, in particular robotics and machine learning, to join forces towards creating a safe and efficient mixed traffic ecosystem. 
\end{abstract}


\end{frontmatter}

\section{Introduction}


We are transitioning 
into a big data era from a data-shortage era, 
thanks to the popularity of ubiquitous sensors, such as 
GPS  \citep{onboard_hecker2018end,onboard_hecker2018learning,onboard_hammit2018evaluation,onboard_lidar_flores2018cooperative,onboard_lidar_zhang2018background}, 
blue tooth \citep{bluetooth_allstrom2014calibration}, 
and smart phones \citep{herrera2010evaluation}. 
Autonomous vehicles (AV), mounted with sensors like camera and LiDAR, 
will potentially provide exploding volumes of transportation data \citep{cv_sas}. 
While moving from a data-sparse to a data-rich era, 
we, the transportation community, urgently need a methodological paradigm shift  
from physics-based models to 
\textit{artificial intelligence} (AI)-guided methods,  
which can project future traffic dynamics comprised of AVs driving alongside human-driven vehicles (HV) 
and assist in socially optimal policy-making. 
\textbf{Physics-based} (or rule-based \citep{zhou2019longitudinal}) models refer to all the scientific hypotheses about the movement of cars or traffic flow, including traffic models on micro-, meso-, and macro-scale; 
while \textbf{AI-guided} methods refer to cutting-edge models that mimic human intelligence, leveraging deep neural networks, reinforcement learning, imitation learning, and other advanced machine learning methods. 

This paper serves as an introduction and overview of the potentially useful models and methodologies from \textit{AI} into the field of \textit{transportation engineering} in the era of mixed autonomy. 
We will discuss the state-of-the-art
applications of AI-guided methods to AV controls, 
identify opportunities and obstacles, 
raise open questions, 
and 
help suggest the building blocks and areas where AI could play a role in mixed autonomy. 
It is the first-of-its-kind survey paper to comprehensively review literature in both transportation engineering and AI for mixed traffic modeling. 
Hopefully this paper will not only 
inspire our transportation community to rethink the conventional models that are developed in the data-shortage era, 
but also reach out to other disciplines, in particular robotics and machine learning, to join forces towards creating a safe and efficient mixed traffic ecosystem. 

Vehicles' driving choices contain three levels: 
operational level (including pedal and brake control, turn signal), 
tactical level (including lane-changing, lane-keeping), 
and strategic level (including routing). 
This paper is mainly focused on the operational and tactical controls of AVs. 
The operational and tactical controls can be further categorized into 
longitudinal control (i.e., car-following, lane-keeping) 
and lateral control (i.e., lane-change). 
Longitudinal control has been studied in various scenarios, including: 
platooning \citep{gong2016constrained,zhou2017rolling,wei2017dynamic,li2018nonlinear}, 
speed harmonization \citep{ma2016freeway,malikopoulos2018optimal,arefizadeh2018platooning}, 
longitudinal trajectory optimization \citep{wei2017dynamic,li2018piecewise}, 
and eco-approach and departure  
at signalized intersections \citep{altan2017glidepath,hao2018eco,yao2018trajectory}. 
Most of the existing studies are limited to a single AV navigating along a highway or dense with human drivers, or all AVs dominate the road with negligible interactions with HVs \citep{katrakazas2015real}. 

 \subsection{Modeling complexity} 

To date, the vast majority of existing research has focused - perhaps unsurprisingly - on two polar scenarios, 
where either a single AV navigates in an ecosystem dense with human drivers, 
or a platoon of AVs move along a highway, with negligible interaction with human-controlled counterparts. 
Much less attention has been accorded to the far more realistic, yet challenging transition path between these two scenarios. 
However, it is precisely this hybrid human-machine space that deserves our concerted attention now, so-called ``mixed autonomy" \citep{wu2017emergent}. 

\begin{figure}[H]
	\centering
	\includegraphics[width=0.9\linewidth,height=\textheight,keepaspectratio]{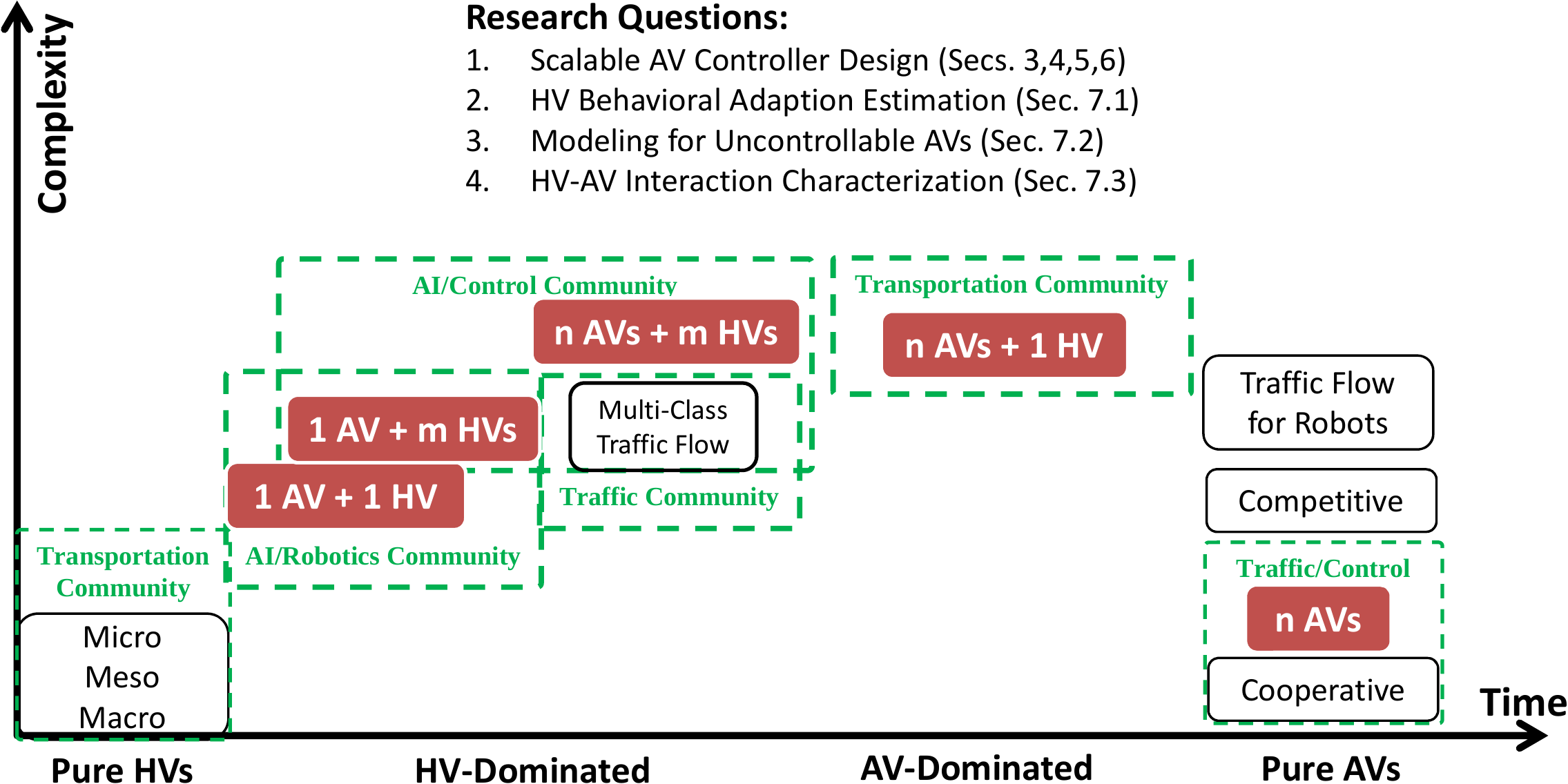}
	\centering 
	\caption{Modeling complexity at each stage (Four research questions will be answered through models of each phase. White boxes indicate the model types for each phase. Research communities are enclosed by dashed green boxes.)
	}
	\label{fig:complex}
\end{figure}

We divide the stage of AV deployment into four phases: 
the pure HVs, the HV-dominated, the AV-dominated, and the pure AVs. 
This paper focuses on the latter three phases. 
Figure~(\ref{fig:complex}) demonstrates the modeling complexity for each phase. 
It is most challenging to model the \emph{HV-dominated} and \emph{AV-dominated} phases, in other words, mixed autonomy. 
This is an understudied phase due to the unknown and complex interactions among different types of vehicles.  
We further divide mixed autonomy by the relative proportion of AVs and HVs using the following notions (indicated in red boxes in Figure~(\ref{fig:complex})): 
\begin{itemize}
	\item 1 AV + 1 HV: one AV interacts with one HV; 
	\item 1 AV + m HVs: one AV navigates the HV-dominated traffic environment;
	\item n AVs + m HVs: multiple AVs navigate the HV-dominated traffic environment;
	
	n AVs + 1 HV: multiple AVs interact with one HV in the AV-dominated traffic environment; It is one special case of n AVs + m HVs. 
	\item n AVs: a pure AV market where all vehicles are replaced by AVs. Accordingly, AVs interact with one another.
\end{itemize}

\begin{rem}
In the process of preparing this paper, we discover another survey paper \citep{zhou2019longitudinal} on longitudinal control of AVs and their impact on traffic congestion. The main difference is that \cite{zhou2019longitudinal} focus on training a single AV on an empty highway or with a few HVs surrounded (corresponding to 1 AV + m HVs), while ours reviews a broader literature in mixed autonomy. 
\end{rem}

In Figure~(\ref{fig:complex}), the community associated with each phase is enclosed by a green dotted box. 
The transportation community has primarily focused on modeling the pure HVs, the AV-dominated, and the pure AVs, 
while AI and control communities are more focused the HV-dominated phase where a single AV or a finite number of AVs navigate the traffic environment. 
Below we will further elaborate how these communities diverge in autonomous control modeling. 

 \subsection{Divergence in the communities} 

Researchers from the transportation community and the robotics community refer to same quantities with different terminologies \citep{fernandez2019game}. Here we present all the relevant terminologies across the communities in Table~(\ref{tab:term}). 
\begin{table}[H]
	\centering\caption{Terminologies across communities (partly adapted from \cite{fernandez2019game})}
	\label{tab:term}
	\begin{tabular}{l||l| l|l}
		\hline
		Transportation & Control & AI & Game Theory \\ \hline\hline
		Traffic & System & Environment & Game \\ 
		Traffic evolution & Dynamics & Transition & Dynamics \\ 
		Traffic state & State space & State & State \\ 
		\hline
		CAV & CACC & AV & AV \\ 
		Car-driver unit & Controller & Agent & Player \\  %
		Vehicle control & Control & Action & Play \\ 
		Vehicle control & Control & Action & Play \\ 
		Vehicle control law & Control law & Policy & Strategy \\ 
		Vehicle control objective & Cost & Reward & Payoff \\ 
		Car-following & Longitudinal control & - & - \\
		Lane-change & Lateral control & - & - \\
		\hline
		Driving behavior & Driver model & Driver intent & Rationality \\  \hline
		Traffic outcome & Optimal control & Optimal policy & Equilibrium \\ \hline
	\end{tabular}
\end{table}

When it comes to the autonomous driving controller design, these two communities take two different paths. 
First of all, these two communities share different goals. 
Transportation researchers aim to understand the influence of AVs on the transportation system performance (from the SYSTEM perspective), such as traffic congestion \citep{zhou2019longitudinal}, 
while robotics researchers are primarily focused on the development of optimal driving policies for AVs to learn and adapt in a stochastic environment  (from the VEHICLE perspective).  
As a consequence, 
these communities investigate problems on different scales. 


The robotics community aims to design AV controllers with human-in-the-loop. 
In particular, researchers on human-cyber-physical systems 
design AVs that actively influence human drivers through mutual interactions, 
in order to achieve efficient driving. 
Their impact on system level performance remains unknown though. 
The transportation systems community aims to understand the influence of AVs on the transportation system performance, including travel time, traffic delay, traffic safety, and emissions. 
Because multi-class microscopic models are not scalable for the varying topology of mixed vehicle types, 
researchers are more focused on 
modeling mixed traffic using the multi-class approach on a macro scale
\citep{talebpour2016influence,levin2016multiclass,melson2018dynamic,chen2016optimal,chen2017optimal,kockelman2017assessment}. 
For example multi-class Lighthill-Whitham-Richards (LWR) model \citep{lighthill1955kinematic} has been used to capture the evolution of hybrid traffic dynamics, assuming AVs are powered by stable controllers \citep{levin2016multiclass,patel2016effects}. 
On road networks, static \citep{chen2016optimal,chen2017optimal} or dynamic \citep{dresner2007sharing,levin2015intersection,levin2016multiclass,patel2016effects,melson2018dynamic} traffic assignment models are developed to capture AVs' intersection coordination and routing behavior. 
Those models on a macroscopic scale may lack detailed interpretation of how different types of vehicles interact at the micro scale.

Second, with different goals, different AV decision-making 
frameworks have been employed. 
Academic researchers have to make various assumptions to implement AV components in their models or simulations, 
because real-world AVs are primarily developed and tested by private companies 
which are not willing to reveal
how the existing AV test fleets on public roads are actually programmed to drive and interact with other road users. 
Accordingly, different driving models lead to different driving behavior and traffic patterns. 

The transportation and the control communities assume AVs are particles or fluids following the physics-based models, including both the micro- and macroscopic traffic models that were originally developed for human drivers,   
and tailors AV behavior on that of HVs 
in which AVs are essentially human drivers but react faster, ``see" farther, and ``know" the road environment better. 
For instance, a majority of studies equate AVs 
to advanced driver-assistance system, 
Connected Adaptive Cruise Control (CACC), 
or commercial semi-autonomous functionality (e.g., Tesla's Autopilot). 
Accordingly, models of the dynamic response of these systems are used as AV driving models \citep{naus2010string,qin2013digital,shladover2015cooperative,delis2016simulation,zhou2020stabilizing}. 
Otherwise AVs are treated like humans but with modified parameters \citep{schakel2010effects,naus2010string,ploeg2011design,milanes2014cooperative,milanes2014modeling,jin2014dynamics}. 
Also, because these automated driving systems are only enabled in designated traffic scenarios, such as platooning,  
control models are thus constrained to these scenarios, not applicable to a generic traffic environment.

The robotics community, on the other hand, treats AVs like AI robots or agents who can continuously explore environments and exploit optimal actions \citep{sadigh2016planning,liu2016enabling}. 
When the environment is observable, AVs select optimal strategies based on predefined reward functions in cooperative or non-cooperative games. 
Reinforcement learning, a cutting-edge learning paradigm initially developed for optimal control of robotics, 
has been naturally deployed for AVs. 
In this framework, human drivers are modeled as part of the environment where AVs move and explore,  
using either an Markov decision process \citep{mukadam2017tactical} or a simulated model-free environment \citep{wu2017emergent,wu2017framework,wu2017flow,wu2018stabilizing,kreidieh2018dissipating}. 

The aforementioned modeling difference arises from the fundamentally different assumptions of vehicle automation levels. 
The transportation and control community is focused on Level-2 or 3 automated vehicles \citep{sae} enabled with vehicle-to-vehicle (V2V) or vehicle-to-infrastructure (V2I) communication. 
The most studied scenario is collaborative platooning, in which connected and automated vehicles (CAVs) drive in a platoon to optimize certain systematic performance measures. 
On the other hand, the robotics and AI community is focused on Level-5 automated vehicles, which are autonomous vehicles. 
A variety of generic traffic scenarios are studied such as lane-keeping, lane-change. merging, and crossing. 

Third, behavioral modeling of HVs and AVs is different. 
The transportation community differentiates HVs and AVs with different models: HVs tend to exhibit unstable, stochastic behavior, 
while AVs can overcome traffic instability with stable controller design. 
In contrast, because the robotics community does not account for collective traffic patterns, they believe human drivers are intelligent for AVs to emulate. 
Thus those studies do not usually distinguish between HVs and AVs. Instead, both HVs and AVs are modeled as AI agents. 

Fourth, there is a discrepancy in how the interactions between AVs and HVs are modeled. 
The transportation community does not formalize how AVs interact with HVs in driving processes. 
On the microscopic level, car-following models (CFM) are applied that implicitly encode how one follows its immediate or far upstream leaders. 
On the macroscopic level, usually multi-class traffic models are adopted, which do not define micro level interactions in detail. 
The robotics community tries to explicitly design microscopic interactions between one or a few AVs and one or multiple HVs. 

%

 \subsection{Driving Policy Mapping: Overview}\label{subsec:policymap}

To further demonstrate the divergence of two communities, we provide a generic mathematical form of driving policy mappings. 
Usually, a driving behavior model, or a driving control or policy is a mathematical mapping from states, i.e., observations of a traffic environment, to actions, i.e., acceleration and steering angle. 
The mapping can be a mathematical formula 
or a neural network (NN). 
\begin{equation}
\phi: \mathbf{s} \longrightarrow \mathbf{a}
\end{equation}

We summarize a variety of mapping forms for longitudinal control policies in Table~(\ref{tab:map}). 
The driving policy mapping is categorized into physics-based and AI-based. 
Physics-based mapping can be characterized by mathematical formulas, 
while AI-based mapping is usually represented by a variety of machine learning models. 
Within each mapping type, we also compare how HVs and AVs are modeled differently. 
As pointed out before, 
when both HVs and AVs are modeled by physics-based mappings, the main difference between HVs and AVs lie in the parameters, 
reflecting that AVs ``sense" better, ``see" farther, and ``react" faster. 
The AI-based mappings assume there exists a complex, highly nonlinear mapping from driving perception to machine activation. 
In these mappings, the difference of HVs and AVs may not be so notable because the goal is to train AVs to exhibit human-like performance. 
In the last column, we list communities along with sample references for each mapping category. Most of the listed references may be revisited in the rest of the paper.

\begin{landscape}
	\setlength\LTcapwidth{\textwidth} 
\small\begin{longtable}{|p{0.14 cm}|p{0.1 cm}|p{1.5cm}||p{2 cm}|p{1.5 cm}||p{1.5 cm}|p{1.5 cm}|p{1.5 cm}|p{1.5 cm}||p{3 cm}|}
		\caption{Mapping categories in the car-following scenario by communities} 
	\label{tab:map} \\ 	\hline
	    \multicolumn{3}{|c||}{\multirow{2}{*}{Models}} & \multirow{2}{*}{Goal} & \multirow{2}{*}{\parbox[t]{1.2 cm}{Input features}} & \multicolumn{4}{|c||}{Behavioral difference} & \multirow{2}{*}{\parbox[t]{3 cm}{Community (Sample references)}} \\ \cline{6-9}
		 \multicolumn{3}{|c||}{} &  &  & information & speed or acceleration & following distance & reaction time &  \\ \hline\hline
		\multirow{3}{*}{\parbox[t]{2 cm}{\rotatebox[origin=c]{90}{Physics-Based}}} & \parbox[t]{1 cm}{\rotatebox[origin=c]{90}{HV}}
		& CFM & capture traffic characteristics & $v_{i-1}-v_i, h_i$ & less, local & \parbox[t]{0.6 cm}{continuously changing} & long & long ($\sim 1.5$ sec) & Transportation \citep{newell1961nonlinear,gipps1981behavioural,treiber2000congested,kesting2010enhanced} \\ \cline{2-10}
		& \parbox[t]{1 cm}{\rotatebox[origin=c]{90}{AV}} 
		& \parbox[t]{1.2 cm}{Linear or nonlinear controllers} &  string stability, improved system performance (capacity, congestion, safety), robust to errors/perturbations & linear $h_i-h^*_i, s_i-s^*_i$, nonlinear $v_{i-1}-v_i, h_i$, others' acceleration & rich, global & set from different speed options, heterogeneous acceleration rate & short & short (0.5 sec) or heterogeneous & Transportation \& control \citep{schakel2010effects,naus2010string,ploeg2011design,milanes2014cooperative,milanes2014modeling,orosz2010traffic,jin2014dynamics,qin2017scalable,chen2019traffic} \\ \hline\hline 
 		\multirow{1}{*}{\rotatebox[origin=c]{90}{AI-Based}} & \parbox[t]{1 cm}{\rotatebox[origin=c]{90}{HV}} 
		& \parbox[t]{1.5 cm}{(Recurrent) NN} & human-like performance & (time series of) $v_{i-1}-v_i, h_i$, reaction delay & local & \multicolumn{3}{|c|}{\parbox[t]{4.5 cm}{unstable behavior, including asymmetric driving behavior, traffic oscillation}} & Transportation \citep{ann,NN_sgd_khodayari2012modified,RNN_zhou2017recurren,huang2018car,NN_DRL_zhu2018human} \\ \cline{2-10} 
		& \parbox[t]{1 cm}{\rotatebox[origin=c]{90}{AV}} & (Deep) RL & individual reward on safety and efficiency & images & local & \multicolumn{3}{|c|}{\multirow{3}{*}{\parbox[t]{4.5 cm}{end-to-end nonlinear, unintepretable controllers}}} & Robotics \citep{Lill-2015, Tianhao-2016,Sallab-2017,Perot-2017,Jaritz-2018}\\ \hline
\end{longtable}
\end{landscape}

Physics-based driving models, widely used by the transportation and control communities, assume each unit behaves like an automated particle or automaton, 
within which human cognitive process and the machine's mechanical dynamics are highly simplified. 
Car-following is the most studied driving behavior, which is 
categorized in terms of spatial resolution:
microscopic driving models and macroscopic traffic flow models. 
In microscopic models, cars are assumed to select their driving velocity and acceleration dynamically 
based on the following distance from their immediate leader, speed difference, and other features. 
The mathematical tool is ordinary differential equation.  
Some of the widely used microscopic car-following models include Newell \citep{newell1961nonlinear}, Gipps' model \citep{gipps1981behavioural}, IDM \citep{treiber2000congested,kesting2010enhanced}, and OVM \citep{orosz2010traffic,jin2014dynamics,qin2017scalable}. 
In macroscopic traffic flow models, cars are assumed to follow hydrodynamics. 
The evolution of aggregate traffic density and velocity are determined using partial differential equation.  
Popular traffic flow models include LWR \citep{lighthill1955kinematic,richards1956shock}, PW \citep{payne1971model}, and ARZ models \citep{aw2000resurrection}. 
MOBIL~\citep{Kesting-2007} introduces the ``politeness" parameter and corresponding utility function to capture intelligent driving behavior in steering and acceleration. 

These physics-based models, however, usually suffer from two major weaknesses~\citep{chen2019model}.  
First, different models have been developed for different driving behaviors. 
For example, car-following and lane-change behaviors are usually modeled separately. 
The model developed for one behavior has to be redesigned manually for different scenarios and tasks.
Second, the predefined motion heuristics usually make strong assumptions about driving behaviors with a small set of parameters, 
which may not be able to capture human's strategic planning behaviors and may not generalize well to diverse driving scenarios in a highly interactive environment. 

 \subsection{AI for decision-making of AVs} 

\emph{``It is not the strongest ... that survives, nor the most intelligent .... It is the one that is most adaptable to change."} 
Instead of hypothesizing explicitly how AVs would drive,  
we believe the futuristic AVs should be designed to act as rational, utility-optimizing agents 
that play best strategies at each level of driving choices. 
By doing so, it would allow AVs to react according to the impending traffic situations and closely mimic human drivers' intelligence. 
However, the major advantage AVs will have over human driver is its ability to access the situation promptly with a better set of information, and thereby enable AVs to react in an optimal way compared to a human driver. 
Natural traffic experiments are, however, costly and highly risky to perform. 
We thus seek an innovative \emph{AI-guided} methodological framework  
for complex multi-agent 
learning and adaptation. 

Despite a significant amount of machine learning efforts given to computer vision, the intelligence of AVs lies in its optimal decision-making at the stage of motion planning. 
We believe the key to empowering AVs' driving intelligence is AI 
or even a broader area ``Artificial general intelligence" (AGI) \citep{ramamoorthy2018beyond}. 
We are seeing a growing number of studies 
that have employed AI methods 
to discover humans' driving behaviors, 
including  
deep learning \citep{NN_moment_tanaka2013development,traj_colombaroni2014artificial,RNN_zhou2017recurrent,NN_SGD_wang2018capturing,onboard_zhu2018modeling}, 
reinforcement learning \citep{vanderwerf2001modeling}, 
and imitation learning \citep{Kuefler-2017,Bhattacharyya2018}. 
More recently, many attempts also focus on human behavior prediction, such as lane changing~\citep{Kumar-2013,Woo-2017,wei2019vision,shou2020long}, 
merging \citep{rios2016survey,bevly2016lane}, 
and stop behavior~\citep{Toru-2006}, to predict with high confidence when a human would change lanes.
However, applications of AI to the decision-making processes of AVs are still emerging and remain understudied. 

Game theory, a mature field for modeling strategic interactions of rational players, has empowered intelligence of multiple interacting machines and 
is revolutionizing the field of AI \citep{tennenholtz2002game}. 
Fortunately, we have seen a gradual convergence in the control, transportation, and AI communities  
that have employed game-theoretic models to design algorithmic decision-making processes for AVs \citep{yoo2012stackelberg,yoo2013stackelberg,kim2014game,Talebpour2015,yu2018human,huang2019stable,huang2020game,huang2020stable}. 
We also believe gaming traffic would be a key feature of future AVs to strategically interact with and navigate through a complex traffic environment.

 \subsection{Organization of the paper} 

The remainder of the paper is organized as follows: 
In Section~\ref{sec:control}, we will provide a general problem statement for AV control in mixed traffic, along with the existing knowledge gaps. 
We will then examine the existing models and methods for AV control in Sections~\ref{subsec:1AV1HV}-\ref{subsec:nAVmHV}.  
Section~\ref{sec:data} presents the methods and models of human and autonomous driving policy learning, respectively. 
Section~\ref{sec:sum} summarizes all the models that have been reviewed. 
In Section~\ref{sec:con}, we present the challenges and insights into modeling the mixed traffic with AI methods and provide potential research areas.

\section{AI-guided driving policy learning for AVs} 
\label{sec:control}

\subsection{Multi-vehicle systems (MVS) in mixed autonomy} 

A mixed traffic system is comprised of a large number of intelligent agents, which are AVs and human drivers. 
They dynamically select driving actions while interacting with the traffic environment. 
Their actions are interdependent in the sense that one's driving action depends on others', via 
either coupled reward functions, 
the common traffic environment state, 
or the action constraints. 
Due to this coupling among agents, the mixed transportation system is a multi-agent system (MAS) - a widely used term in the control and robotics community.   
Specifically, we call it a ``multi-vehicle system (MVS)."

\begin{defn} \textbf{(AV control problem statement in mixed-autonomy.)}
	In a mixed traffic system, there are $N$ controllable AVs indexed by $n\in\{1,2,\dots,N\}$ driving along a stretch of a road, 
	with initial states $s^{(1)}_{0},\dots,s^{(N)}_{0}$. 	
	Each car aims to select a sequence of optimal driving controls (e.g., acceleration or steering angle) in discrete (i.e., $a^{(n)}_{1}, \dots, a^{(n)}_{T}$) or continuous time steps (i.e., $a^{(n)}(t), t\in\left[0, T\right]$) 
	over a predefined planning horizon $\left[0, T\right]$. 
	The selected controls are solved 
	by minimizing a common  
	or an individual cost functional.  
	Human drivers and other uncontrollable AVs serve as the background traffic, which evolves according to some dynamics. 
	\emph{What is a scalable distributed control strategy for these AVs?}
\end{defn}



\subsection{Research questions}

While reviewing the methodologies, we primarily focus on the following \emph{research questions}: 

\begin{enumerate}
	\item What \textbf{scalable driving policies} are to control a large number of AVs in mixed traffic comprised of human drivers and uncontrollable AVs? (Sections~\ref{subsec:1AV1HV}-\ref{subsec:nAVmHV})
	\item How do we estimate human driver behaviors? (Section~\ref{subsec:HV})
	\item How should the driving behavior of uncontrollable AVs be modeled in the environment? (Section~\ref{subsec:AV})
	\item How are the interactions between human drivers and autonomous vehicles characterized? (Section~\ref{subsec:interact})
\end{enumerate}

Below we will present the knowledge gaps that exist in the literature with respect to each research question.

 \subsection{Roadmap to navigate literature: Control dimensions} 

 In the subsequent sections, we will give an overview of how the existing literature addresses these three gaps. 
 We categorize these studies based on how many AVs and HVs are involved. 
 They include: 
 one AV interacts with one HV (1 AV + 1 HV), 
 one AV navigates in the HV-dominated traffic environment, i.e., one AV interacts with multiple HVs (1 AV + m HVs), 
 multiple AVs interact with many HVs (n AVs + m HVs, $n<<m$), 
  multiple AVs interact with one HV (n AVs + 1 HV), 
 and a pure AV market (n AVs).

\begin{figure}[H]
	\centering
	\includegraphics[width=1.0\textwidth, keepaspectratio=true]{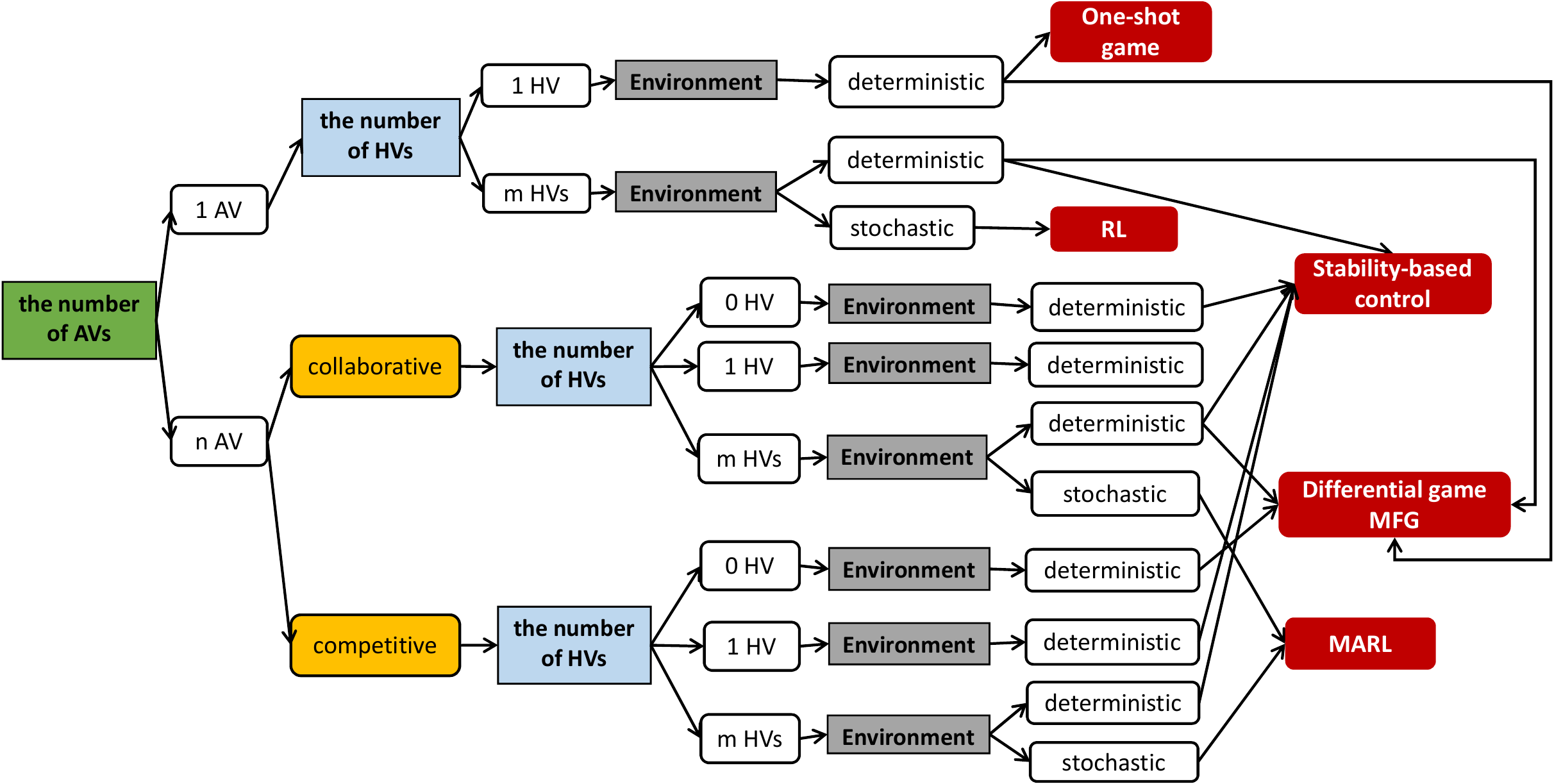}
	\caption{Roadmap to mixed traffic models}
	\label{fig:road}
\end{figure}

Other than the number of AVs and HVs,  
we further categorize the multi-AV control problem based on two dimensions: 
the first dimension is whether these controllable AVs are cooperative or not, 
while the second dimension is whether the AV control takes into account uncertainty arising from the external environment. 
Fig~(\ref{fig:road}) provides a roadmap to navigate readers to literature on AV controls in mixed traffic. 
The colored rectangles represent the classification criteria and the red round-corner rectangles (as the end nodes) represent the methodologies used to control AVs. 

\section{1 AV + 1 HV, 1 AV + 1 AV: General-sum Game-Based Control} 
\label{subsec:1AV1HV}


Game theory is a natural approach to model the non-cooperative strategic interactions among AVs or between one AV and one HV, 
who are usually taken as intelligent agents aiming to optimize an individual objective function. 
In the game theoretic framework, cars are referred to as ``agents" or ``players". 

\subsection{One-shot game}

The one-shot two-person game is applied to model two cars' strategic actions at one step. 
Driving \citep{yoo2012stackelberg}, 
merging \citep{liu2007game,yoo2013stackelberg}, 
lane-changing \citep{Talebpour2015,yu2018human,zhang2019game,yoo2020game}, 
and unprotected left-turning behavior \citep{rahmati2017towards}
is modeled as either 
a two-person non-zero-sum non-cooperative game 
\citep{liu2007game,Talebpour2015},  
a Stackelberg game \citep{yoo2012stackelberg,yoo2013stackelberg,yu2018human,zhang2019game,yoo2020game} 
or a mixed-motive game \citep{kim2014game}.
The outcome of these games can be a pure or mixed Nash equilibrium, based on the payoff bimatrix. 
The payoff of a given strategy accounts for traffic safety and efficiency, depending on current driving speed, relative positions, reaction and perception time, 
aggressiveness, and collision avoidance. 
When human driving behavior is modeled, 
human driver data is collected to estimate parameters of payoff functions, 
using 
bi-level optimization \citep{liu2007game}, 
simulated moments \citep{Talebpour2015}, 
and maximum likelihood \citep{rahmati2017towards}. 
When one of the game player is an AV, 
utility or reward needs to be designed 
while accounting for aggressiveness of surrounding drivers \citep{yoo2020game}. 
\cite{zhang2019game} further develops a game theoretic model predictive controller that solves a Stackelberg equilibrium
with multiple interacting vehicles continuously. 


\subsection{Dynamic/Continuous game: perfect information, full observability}

The one-shot game cannot model vehicles' dynamic driving actions. 
To solve for time-varying controls, dynamic optimal control \citep{wang2015game}, 
model predictive control (MPC) \citep{wang2016cooperative,gong2016constrained,gong2018cooperative}
or rolling horizon control \citep{swaroop1994comparision,wang2014rolling1,wang2014rolling2,zhou2017rolling}  
have been formulated for AVs. 
When one agent solves an optimal control problem, while its interacting agent also does so with conflicting goals,
a differential game forms \citep{wang2015game}. 
\textit{Differential game} models dynamic behaviors of interacting agents with conflicting goals, where agents' optimal strategies are obtained from optimal control problems. 
The classical differential game is primarily focused on two players and becomes intractable for an equilibrium of more than two players. 
Below we will first demonstrate the generic model between one AV and one HV 
and then present several simplification techniques developed to compute a sequence of optimal actions. 

A simultaneous differential game between one AV and one HV is formulated below: 
\begin{subequations}\label{eq:one-dynamic}
	\begin{align}
	& \dot{\mathbf{s}} = f\left(\mathbf{s},\mathbf{a}^{(AV)},\mathbf{a}^{(HV)} \right), \\
	& \mathbf{a}^{(AV)}(t) = \arg\min_{\dot{\mathbf{s}}^{(AV)} = f^{(AV)}\left(\mathbf{s},\mathbf{a}^{AV} \right)} \int_{t}^{t_f} r^{(AV)}\left(\mathbf{s},\mathbf{a}^{(AV)},\mathbf{a}^{(HV)}\right) dt, \\
	& \mathbf{a}^{(HV)}(t) = \arg\min_{\dot{\mathbf{s}}^{(HV)} = f^{(HV)}\left(\mathbf{s},\mathbf{a}^{HV} \right)} \int_{t}^{t_f} r^{(HV)}\left(\mathbf{s},\mathbf{a}^{(AV)},\mathbf{a}^{(HV)}\right) dt.
	\end{align}
\end{subequations}	
where, \\
$t_f$: a predefined planning horizon; \\
$\mathbf{s}$: the system state including the states of both the AV and the HV;\\
$f$: the state dynamic function for continuous-time, or the state transition function for discrete-time;\\
$\mathbf{a}^{(AV)},\mathbf{a}^{(HV)}$: the dynamic driving actions or policies of the AV and the HV, respectively; \\
$r^{(AV)},r^{(HV)}$: the reward function of the AV and the HV, respectively.

If the AV were to be able to predict the HV's strategy in the entire planning horizon, it then optimize its own objective function that depends on both its own current and future strategies as well as the HV's current and future strategies to generate a continuous sequence of control strategies along this horizon and implement it. 
The same process holds for the HV. 
Due to the dynamic coupling, it is challenging to solve this equilibrium.

To simplify, several techniques have been applied. 
\cite{sadigh2016planning,lazar2018maximizing} have simplified the original two-player differential game to a leader-follower game (or Stackelberg game) played at discretized time steps. 
In this game, the AV takes actions first. 
Then the HV observes actions taken by the AV and predicts the AV's future action based on the AV's historical actions, 
maximizes its own objective and calculates its own future actions for a short period of time. 
Then the AV maximizes its own objective using the HV's future actions 
and replans repeatedly using MPC at each iteration. 
In other words, in a leader-follower scheme, 
the AV directly solves an optimization based upon its prediction of human driver actions rather than human's actual strategies. 
The advantage of the Stackelberg game is that the AV can be designed beforehand to influence uncontrolled HVs via a carefully selected reward function \citep{sadigh2016planning}. 
The reward function contains two parts: one controls the AV's driving efficiency and safety, while the other determines the influence the AV would like to impose to neighboring HVs. 
\cite{lazar2018maximizing} extends this framework to a Stackelberg game between one AV and multiple HVs, but assumes that one AV only influences one HV and the actions of others HVs are fixed. 
%
%

\cite{fisac2019hierarchical} further develops a hierarchical game-theoretic planning scheme, where 
the strategic planner solves a closed-loop dynamic game with approximate dynamics in a relatively long planning horizon (e.g., 5 second), 
while the tactical planner solves an open-loop trajectory optimization 
with high-fidelity vehicle dynamics over a shorter planning horizon (e.g., 0.5 second). 
On the strategic planner level, the AV and the HV still play a feedback Stackelberg dynamic game 
in which their driving actions are recursively solved through successive application of dynamic programming. 
The solved optimal Q-value obtained from the strategic level is then introduced to the objective function of the tactical level as a guiding terminal reward representing an optimal reward-to-go. 
On the tactical planner level, the trajectory of the AV is iteratively optimized 
using a nested optimization problem that estimates the human's best trajectory response to each candidate plan in the short-term planning horizon. 
The hierarchical game-theoretic model is tested on two scenarios with merging and overtaking maneuvers: 
one on a straight empty multi-lane highway with only two-vehicle interaction 
and one with the presence of a third vehicle (i.e., a truck with a slower moving speed). 

\cite{li2018game,tian2018adaptive,tian2019game} assume two different game structures for HVs and AVs, respectively. 
Human drivers play a game based on hierarchical reasoning. 
A level-$0$ play ignores the interaction of other players, 
while a level-$1$ player assumes all other players are level-0 players. 
Similarly, a level-$k$ player assumes that all other players act according to level-$(k-1)$ models. 
In other words, in a two-driver scenario, 
when the ego driver is level-$1$, 
the action of her opposite driver is assumed a level-$0$ driver whose actions are solved without accounting for the vehicle interaction.  
Then the ego driver selects her actions based on the fixed actions of the opposite driver. 
The simultaneous game is reduced to solving two optimal control problems sequentially. 
When the ego driver is level-$2$, the action of this ego driver depends on that of the opposite vehicle, which in turn depends on that of ego vehicle. This becomes an embedded game like proposed in \cite{sadigh2016planning}. 
Accordingly, the AV plays an adaptive game against HVs 
in which the AV predicts the opponent vehicle's actions
based on the opponent vehicle's driver type, 
and update its belief of the driver type using the difference between the actual action and the predicted action and then update its own actions.   
This game method is tested on multi-lane highways and unsignalized intersections (including four-way, T-shape, and roundabout). 


\section{1 AV + m HV 
} 
\label{subsec:1AVmHV}

In this section, we will first briefly mention stability-oriented controls, and then 
introduce two types of AI-based modeling approaches: game based and reinforcement learning based. 
There exist only a few studies using game-theory based control, partly due to high dimensionality of the coupled game system. 
Accordingly, a majority of studies employ reinforcement learning based AV control, which will occupy the most space in this section. 

\subsection{Deterministic stability-oriented control} 
When the environment is deterministic, 
the traffic control community aims to understand how one AV can stabilize a HV platoon using linear \citep{cui2017stabilizing,wang2018infrastructure} or nonlinear controllers \citep{jin2014dynamics,jin2018connected}, 
based on the concept of head-to-tail stability (i.e., stability from the first vehicle to the last vehicle in a platoon \citep{jin2014dynamics}). 
Field experiments have also demonstrated the feasibility of using one AV to stabilize HVs  \citep{stern2018dissipation,jin2018connected}. 
Because the physics-based models are not the focus of this paper, interested readers can refer to \cite{li2014survey} for a comprehensive survey of stability-based controls. 

\subsection{Differential game based control} 
Assume the HV-dominated environment is deterministic 
and every vehicle interacts among one another in a game-theoretic framework, 
we can formulate a simultaneous differential game between one AV and multiple HVs below: 
\begin{subequations}\label{eq:n-dynamic}
	\begin{align}
	& \dot{\mathbf{s}} = f\left(\mathbf{s},\mathbf{a}^{(AV)},\mathbf{a}_{1}^{(HV)}, \cdots,\mathbf{a}_{M}^{(HV)} \right), \\
	& \mathbf{a}^{(AV)}(t) = \arg\min_{\dot{\mathbf{s}}^{(AV)} = f^{(AV)}\left(\mathbf{s},\mathbf{a}^{AV} \right)} \int_{t}^{t_f} r^{(AV)}\left(\mathbf{s},\mathbf{a}^{(AV)},\mathbf{a}_{1}^{(HV)},\cdots,\mathbf{a}_{M}^{(HV)}\right) dt, \\
	& \mathbf{a}_{m}^{(HV)}(t) = \arg\min_{\dot{\mathbf{s}_m}^{(HV)} = f^{(HV)}\left(\mathbf{s},\mathbf{a}_{m}^{HV} \right)} \int_{t}^{t_f} r_{m}^{(HV)}\left(\mathbf{s},\mathbf{a}^{(AV)},\mathbf{a}_{m}^{(HV)}\right) dt, m=1,\cdots,M.
	\end{align}
\end{subequations}	
where $\mathbf{a}_{m}^{(HV)}$ is the dynamic driving actions or policies for HV $m, m=1,\cdots, M$. 
Other notations carry the same meaning as before. 

\cite{schwarting2019social} develops an autonomous control policy by solving an iterative best-response, with embedded levels of tacit negotiation. 
In a two-agent case, 
an iterative best-response can be written as  $\mathbf{a}^{(AV)}(\mathbf{a}^{(HV)}(\mathbf{a}^{(AV)}(\cdots)))$ 
where one's strategy is solved using Equ~(\ref{eq:one-dynamic}). 
In an MAS, a system of interdependent optimization is reduced to a single-level optimization using KKT conditions. 
The resulting Nash equilibrium not only offers a control law for the AV but also predicted actions for other HVs. 
The innovation of this study is to include a term ``Social Value Orientation (SVO)" into the reward function of HVs, representing HVs' driving aggressiveness. 
One can adjust its SVO value while interacting with another vehicle.  
The control law is validated in highway merging and unprotected left turn. 
Social preference learning can improve the AV's performance by $25\%$. 

\cite{liu2015safe,liu2016enabling} combine multiple HVs as one effective human 
and assume a sequential game in which HVs lead and the AV play reactive strategies.  
By mapping a baseline control law to a set of safe control, 
an online algorithm is developed for the AV controller to incorporate human intentions as safety constraints. 

\subsection{Reinforcement learning based control}

In the HV-dominated traffic, 
a single AV's driving policy selection can be treated as a sequential decision-making process in a partially or fully observable random environment.  
Learning driving policies are needed 
to predict vehicles' acceleration and steering angle using their environmental information as input. 

Reinforcement learning, which enables the intelligent agents to learn optimal policies driven by a reward, 
has made breakthroughs to achieve super-human-level performance in game playing, such as Atari~\cite{Mnih-2015}, Go game~ \citep{Silver-2016}, Poker~\citep{brown_superhuman_2018, brown_superhuman_2019}, Dota 2~\citep{OpenAI_dota}, and StarCraft II~ \citep{vinyals_grandmaster_2019}. 
Its application to autonomous driving has become a promising direction. 
The dynamic motion planning of a single AV is usually modeled using a Markov decision process (MDP) \citep{puterman_markov_1994} or reinforcement learning (RL) 
\citep{sutton_introduction_1998}.  

\begin{figure}[H]
	\centering
	\includegraphics[width=0.45\textwidth, keepaspectratio=true]{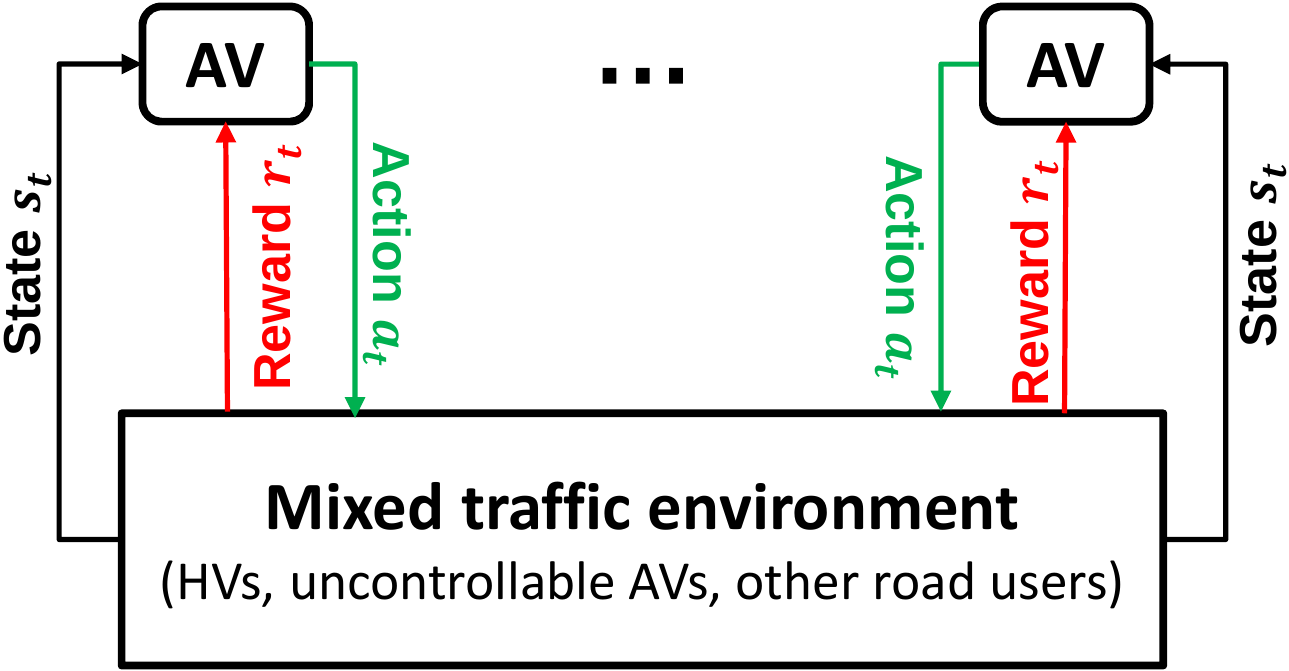}
	\caption{Single- and Multi-Agent Reinforcement Learning Framework}
	\label{fig:rl}
\end{figure}

When RL is used to control AVs in a stochastic environment,  
the basic idea is demonstrated in Fig~(\ref{fig:rl}). 
Let us first discuss the single-agent RL setting. 
One controllable AV (the host vehicle whose controller needs to be designed) perceives the state of the mixed traffic environment, comprised of HVs, other AVs that are not controllable, and road users. 
Based on some predefined reward function, 
it executes an action (such as acceleration or steering angle), which in turn transforms the state of the traffic environment. 
In return, the environment provides a reward to the AV. 
Based on the received reward and the new environment state, the AV further solves an optimal policy based on the reward function and selects an action. This process iterates till the AV finishes its entire control process. 
When there are multiple AVs that are all required to make decisions simultaneously, we need a multi-agent RL (MARL) framework, which will be discussed in Section~\ref{subsec:nAVmHV_noncoop}.

MDP implicitly assumes that the agent can fully observe the state dynamics. 
In other words, after the agent applies an action, he knows 
the probability of the next state the system will move to. 
A majority of transportation studies assume connectivity among vehicles via V2V or V2I. 
Thanks to these communication technologies, every driver obtains full information of other drivers and the system state.    
Control and robotics researchers, on the other hand, make various assumptions on observability \citep{liu2015safe,liu2016enabling,liu2018improving,bouton2017belief,Bouton2018uai}. 
One may observe others' positions, headings, and sometimes longitudinal and lateral velocity, but not accelerations. 
Even a driver observes the entire state of other drivers, she may not know the intention of those drivers.  
Accordingly, the AV has to maintain a belief state space over all possible states based on its observations. 
Therefore a partially observable Markov decision process (POMDP) model is widely used in modeling a single AV's motion planning. 
Assuming each AV follows a (partially observable) Markov decision process, the single-AV control problem in the HV-dominated traffic can be formulated as an (PO)MDP, whose components are described below. 
\begin{itemize}
	
	\item ${\cal S}.$ The state space of the HV-dominated traffic environment. A state $\mathbf{s} \in {\cal S}$ contains the information of the ego car and all surrounding vehicles.  
	For instance, the physical state of the ego car at time $t$ can be represented by $s^e_t = (p^x_t, p^y_t, \theta_t, v_t, a_t)$, where $p^x_t, p^y_t$ are its longitudinal and lateral positions, $\theta_t$ is the axle angle, $v_t, a_t$ are speed and acceleration. 
	The physical state of the $n$ other cars at time $t$ is $s^I_t = (p^x_t, p^y_t, \theta_t, v_t, a_t)$. 
	
	\item ${\cal A}.$ The action space. The action of the AV, denoted as $a \in {\cal A}$, is a two-dimensional vector, including acceleration and steering angle.  
	
	\item ${\cal O}.$ The observation space. The observation of the AV is $o\in {\cal O}$. 
	For instance, the observation space for the $i^{th}$ car at time $t$ is ${\cal O}_t = (p^x_t, p^y_t, \theta_t, v_t)$. The $i^{th}$ car's acceleration $a_t$ is usually not observable.
	
	\item ${\cal G}.$ The observation function. 
	For the AV, $\mathbf{s}$ may not be fully observable. 
	Instead, it draws an observation $o \in {\cal O}$ that is correlated with $\mathbf{s}$ according to an observation function ${\cal G}: S \times {\cal O}\rightarrow [0,1]$, i.e., $o \sim {\cal F}(o|\mathbf{s})$. 
	
%
	
	\item ${\cal P}.$ The state transition function, i.e., ${\cal S}\times {\cal A} \times S \rightarrow [0, 1]$. 
	The action $\mathbf{a}$ triggers a state transition $\mathbf{s} \rightarrow \mathbf{s}'$ according to function  ${\cal P}(\mathbf{s}'|\mathbf{s},\mathbf{a})$. 
	This state transition can rely on a specific form of $P$ 
	or
	provided by a traffic simulator (i.e., model-free).

	\item ${\cal R}.$ The reward space. Along with the state transition, the AV receives an immediate reward, i.e., $r \in {\cal R}:{\cal S}\times {\cal A} \times {\cal S} \rightarrow \mathbb{R}$. 
	The reward $r$ may include traffic safety (e.g., off-road/collision avoidance), efficiency (e.g., fast speed), and emissions. 
	
\end{itemize}


The AV aims to derive an optimal policy $\pi^{*(AV)}$ by maximizing its expected cumulative reward. 


In contrast to theory-driven AV controllers, such as game-theory drive modeling, 
RL-based AV controllers are model-free and End-to-End (End2End), which directly maps sensory inputs to control commands. 
Behavioral cloning (BC) simplifies the AV policy learning as a supervised-learning problem, which usually performs well when driving data are sufficient or the driving task is for limited regions. \cite{Pomerleau-1989}  introduce a multi-layer network learned from simulated road images to control a vehicle to follow real roads. The next milestone of AVs is to employ convolutional neural networks (CNNs) to efficiently process raw camera images, which helps AVs drive through an obstacle-filled road after training on similar scenarios~\citep{Muller-2006}. Later on, CNN-based AV controllers are widely studied, and recently work includes NVIDIA's PilotNet~\citep{Bojarski-2016,Bojarski-2017} to control AVs in real traffic situations, Rausch's deep CNN policy~\citep{Rausch-2017} and DeepPicar~\citep{Bechtel-2018} for steering angle control, Agile driving~\citep{Pan-2018} for steering angle and velocity controlling in aggressive scenarios. Temporal dependencies of the driving data have been considered to improve the performance of AV control, and recently, long-short-term memory (LSTM) and its variants have been leveraged for End2End AV policy learning. \cite{Xu-2017}  proposes FCN-LSTM, a combination of a fully-convolutional network (FCN) and LSTM, which can predict a distribution of future
vehicle ego‐motion data. \cite{Erqi-2017} develops a convolutional LSTM (C-LSTM) for learning both visual and dynamic temporal dependencies of driving. \cite{onboard_hecker2018end}   introduce Drive360, which combines CNN, fully connected layers and LSTM to integrate information from multiple sensors to predict the driving maneuvers. \cite{Bansal-2018} from Waymo  presents  ChauffeurNet, a  mid-to-mid driving policy learning framework, in which inputs are prepossessed before receiving by an RNN to generate low-level controls.
BC has several shortcomings: (1) BC requires the collection of huge amount of expert driving data, which is time-consuming and expensive; (2) It can only learn the driving skills that are covered in the data, and may not generalize to diver real-world driving scenarios; and (3) Since BC is based on supervised learning using human actions as target, it can never exceed the human-level performance of experts.


Different from BC, general deep reinformcement learning (DRL) driving systems are mainly developed in simulation, which provides consequential information, such as reward, for AVs to learn from.
\cite{wu2017emergent}  train an autonomous driving controller using (deep) reinforcement learning  with a designated reward function that avoids crashes into other agents, and applied trust region policy optimization (TRPO) method to train a Gaussian Multilayer perceptron (MLP) policy in SUMO simulator for improving traffic efficiency. \cite{Lill-2015}   apply a deep deterministic policy gradient (DDPG) RL algorithm to control a car in a simulation environment. Their work designed a reward, which provides a positive reward at each step for the velocity of the car projected along the track direction and a penalty of -1 for collisions.
\cite{Sallab-2017}  develop a integrated deep Q-network (DQN), which integrates attention models to make use of glimpse and action networks to direct the CNN kernels for steering command in TORCS simulator, which provides a positive/negative reward for on/off-lane situations. \cite{Perot-2017}  propose an asynchronous advantage Actor-Critic (A3C) method for training a policy network for realistic games, such as World Rally Championship 6 and TORCS simulator. This work has been enhanced by \cite{Jaritz-2018} with an improved convergence and generalization. Both of these studies have designed the reward as a function of the distance to the road center and angle between the road's and car’s heading.

Monte Carlo tree search (MCTS) is one of the most effective methods for solving decision making problems online~\citep{Browne-2012} and has begun to be applied to AVs in recent years. \cite{Paxton-2017} integrate MCTS with hierarchical neural net control policies trained on Linear Temporal Logic (ITL) constraints for motion and path planning in complex road environments. They designed the reward function as a combination of cost terms upon current continuous states (e.g., location or speed), and a bonus terms based on completing immediate goals (e.g., stopping at the sign or existing a region), and a penalty term for constraint violations.
\cite{Sunberg-2017} use  MCTS to infer the internal state of traffic participants for operating safe lane changes on a highway. Their reward design penalized the average time taken for the ego to reach the target lane and the number of hard braking maneuvers that any vehicle undertakes during the time for the ego vehicle to reach the target lane. \cite{Hoel-2020} combine MCTS with DRL to achieve tactical highway driving, and in their work, a deep neural network is trained to guide MCTS to the relevant regions of the search tree, while  MCTS is used to improve the training process of the neural network at the same time. The associated reward is a combination of cost terms concerning the number of lane changes and difference from the desired speeds, and a bonus terms for highway exit (i.e., goal achieved).

There is another direction of DRL-based AV control, using prior knowledge or classical theory-driven controllers to constrain the learning and behaving of neural network-based driving models.
\cite{bouton2017belief} impose a computational safety factor as a penalty in the reward function rather than a hard constraint \citep{bouton2017belief}, and as a result, the driving policy solved from MDP cannot avoid accidents. \cite{Bouton2018uai} add a model checking step to enforce probabilistic guarantees of the trained driving policy on an RL agent, and they used simplified reward function to penalize the number of action steps and award goal accomplishment.
\cite{Tianhao-2016} propose to combine the traditional MPC method with RL in the framework of guided policy search for controlling autonomous aerial vehicles, where a deep neural network policy is trained on data generated by MPC for training robustness and generalizable control. The RL training is based on a cost function that measures the distribution difference between the action generated from the policy model and the data generated from MPC.
\cite{Jianyu-2019} develops a hierarchical control framework, where the higher-level controller employs MDP to solve a reference driving policy and the lower-level controller implements it accounting for safety concerns.

For more examples of using single-agent RL for AV controlling, we refer readers to recent surveys such as~\citep{zhou2019longitudinal,Grigorescu-2020,Kiran-2020-arXiv}.

\section{n AVs: A Driverless World} 
\label{subsec:nAV}

Control of a single AV is far from sufficient to exploit the potentials of AVs in the era of mixed autonomy, when an increasing number of AVs are introduced to public roads. 
Systems with multiple AVs have attracted increasing interest in recent years. 
In this and the next sections, we go against the AV deployment timeline by first discussing 
the pure driverless world, denoted as ``n AVs", followed by the mixed market (of $n$ AVs + $m$ HVs). 
The pure AV market precedes the mixed one,  
because the former can be generalized to the latter by adding a traffic background comprised of HVs. 

 \subsection{Classification of multi-AV control models}


Based on whether all the vehicles solve for a common or an individual objective function, 
the multi-AV control models can be divided into two classes: 
\textbf{\emph{cooperative control}} and \textbf{\emph{non-cooperative game}}. 

The multi-AV control problem can also be categorized based on whether the traffic environment that AVs navigate is deterministic or stochastic. 
The transportation community generally assumes that every CAV has global connectivity with the leader and/or with other vehicles in a platoon via V2V or V2I. 
Accordingly, the traffic environment is known and deterministic.  
These studies are primarily focused on optimization of a centralized or a distributed control,  
with a few exceptions that consider measurement errors or communication delays. 
The robotics community, on the other hand, assumes each AV can only observe local information using its own sensors, such as camera, LiDAR. 
Accordingly, the traffic environment is full of uncertainty, including but not limited to stochasticity induced by other neighboring vehicles' dynamic driving actions and exogenous randomness. 

In this and the next sections, we will categorize the literature using vehicle cooperation as the primary category and the environmental stochasticity as the secondary category. 
In the pure AV market, a majority of studies assume the environment is deterministic because all the AVs are controllable and fully observable, while stochasticity could originate from measurement errors or communication delay. 
In the mixed market, the environment is highly random due to stochasticity of human driving behavior, but there are also studies by the traffic control community that assume a deterministic environment. 

\subsection{Cooperative control}

\subsubsection{A deterministic environment} 

In a deterministic environment, assuming that the central planner knows the state of every vehicle and aims to optimize a total system performance, 
from the system perspective, the multi-AV control problem is formulated as an optimization problem: 


\begin{subequations}
	\begin{align}
	& \min_{\mathbf{a}} 
	J^N (s_1, a_1; \cdots; s_{N}, a_{N}), \\
	\mbox{s.t. } & \dot{s_i}(t) = f\left(s_i, a_i\right), \\
	& s_i \in {\cal S}_i (s_{-i}, a_{-i}), \\
	& a_i \in {\cal A}_i (s_{-i}, a_{-i}), \\
	& \mathbf{a}(t) \in {\cal A}.
	\end{align}
\end{subequations}
where, \\ 
$J^N (\cdot)$: the common objective function shared by a total $N$ vehicles; \\
${\cal S}_i (s_{-i}, a_{-i}), {\cal A}_i (s_{-i}, a_{-i})$:  the vehicle $i$'s state and control constraints.
Vehicle $i$'s state and control are constrained by other vehicles. 
Other notations remain. 

When AVs are programmed to optimize its own objective and not cooperate with other AVs, 
the multi-AV control becomes a non-cooperative game. 
In a non-cooperative system, vehicles select theirs own controls to achieve individual goals, which may likely conflict with others' goals. 
Compared to the cooperative control, the non-cooperative interactions among AVs are relatively understudied. 
A non-cooperative framework for a simultaneous game is formulated as: 

\begin{subequations}
	\begin{align}
	& \min_{a_i} J^N_i(s_i, a_i;s_{-i}, a_{-i}), i=1,\cdots,N \\
	\mbox{s.t. } & \dot{s_i}(t) = f\left(s_i, a_i\right), \\
	& s_i \in {\cal S}_i (s_{-i}, a_{-i}), \\
	& a_i \in {\cal A}_i (s_{-i}, a_{-i}), \\
	& \mathbf{a}(t) \in {\cal A}.
	\end{align}
\end{subequations}
where $J^N_i (\cdot)$ is vehicle $i (i=1,\cdots,N)$'s individual objective function. 

In the next two subsections, we will review the existing models of each category that materialize the above two control schemes.  

 
A majority of research on control of multiple AVs 
falls within the category of cooperative coordination. 
In other words, AVs are assumed to communicate with one another for global traffic information and optimize a common goal of traffic flow improvement. 
Cooperative control has been widely studied in multi-robotic systems. 
Swarm intelligence \citep{bogue2008swarm,venayagamoorthy2004unmanned}, 
formation control \citep{chen2005formation}, 
and consensus control \citep{zegers2017consensus,li2018consensus} have been widely used for a group of robots with a centralized goal to accomplish a task collaboratively, 
so is in multi-AV control~\citep{Cathy-2018,Lazar-2018}. 
 
 
 In a cooperative MVS, 
 the movement of vehicles is coordinated by a central controller or planner to achieve a common goal, such as 
 to collectively stabilize traffic flow and smoothen traffic jam \citep{wang2016cooperative,gong2016constrained,gong2018cooperative}, 
 to optimize driving comfort \citep{wang2014rolling2,zhou2017rolling}, 
 or to improve fuel efficiency \citep{wang2014rolling1,yao2018trajectory}. 
 To achieve coordination, \emph{full observability} and \emph{full controllability} is required, 
 meaning that all vehicles' states and controls are known to the central controller 
 and every vehicle can be controlled in a centralized or distributed manner. 
 The communication topology in a platoon of vehicles determines the degree of cooperation among CAVs \citep{li2014survey}. 
For example, control protocols can be designed for a platoon of vehicles to reach an equilibrium state using a consensus based approach. Accordingly, the car-following coupling among vehicles are modeled as a consensus problem and a distributed nonlinear delay-dependent control algorithm is used to solve a safe velocity \citep{li2018nonlinear}. 
 
 Assuming connectivity between predecessors and followers as well as between platoon leaders and followers, 
 CACC contains two control policies: 
 constant spacing 
 \citep{swaroop1996string,darbha1999intelligent,swaroop2001direct} and
 constant time headway 
 \citep{ioannou1993autonomous,rajamani2001experimental,van2006impact,naus2010string,vanderwerf2001modeling,zhou2017rolling,arefizadeh2018platooning,stern2018dissipation}. 
 AVs longitudinal acceleration control can also be modeled using nonlinear CFMs, which is discussed in Sec.~\ref{subsec:policymap}. 
 All the aforementioned studies aim to develop a string stable car-following controller in order to smoothen traffic flow and prevent stop-and-go waves.  
 But none of them considers control and physical safety constraints \citep{gong2018cooperative}. 
 In other words, interactions among vehicles are not explicitly modeled \citep{li2018nonlinear}. 
 To explicitly model the physical interaction between vehicles, a growing body of literature formulates a platoon of AV longitudinal control as optimal control problems. 
The control policies based on linear spacing policies or non-linear CFMs are special cases of optimal control problems \citep{wang2014rolling2}.

 Define ${\cal C}_{run}, {\cal C}_{ter}$ as running cost and terminal cost of a platoon, respectively, and $t_f$ as the planning horizon. 
 \begin{subequations}
 	\begin{align}
 	& \min_{{\mathbf{a}}} J^N
 	= \min_{{\mathbf{a}}} \int_{0}^{t_f} {\cal C}_{run}\left(\mathbf{s}(\tau),\mathbf{a}(\tau)\right) d\tau
 	+ {\cal C}_{ter}\left(\mathbf{s}(t_f),\mathbf{a}(t_f)\right), \\
 	\mbox{s.t. } & \dot{\mathbf{s}} = f\left(\mathbf{s}(t),\mathbf{a}(t)\right), \\
 	& \mathbf{s}(t) \in {\cal S}, \\
 	& \mathbf{a}(t) \in {\cal A}.
 	\end{align}
 \end{subequations}
 
Centralized control requires the central controller to solve for an optimal control for each car at each time step. 
 It is challenging to solve a centralized control of this type, because: 
 (1) all vehicles' states and controls are coupled through objective functions and constraints;  
 (2) A longer planning horizon requires prediction of future traffic dynamics, which may suffer from both curse of dimensionality and disturbances. 

To resolve the first issue of state coupling, 
a distributed algorithm is usually designed and implemented on each vehicle \citep{wang2016cooperative,gong2016constrained,gong2018cooperative,li2018nonlinear}. 
Consensus based approaches are also employed to design a control protocol for a platoon of vehicles to reach a consensus and a distributed nonlinear delay-dependent control algorithm is designed to solve a safe velocity \citep{li2018nonlinear}. 
To resolve the second issue of prediction horizons, 
the original optimal control problem can be approximated as a one-step MPC and a distributed algorithm is developed. 
The MPC control is close to optimal control strategies if the planning horizon is short, 
but may deviate when the planning horizon is long.  
MPC is also employed as a higher level control model  
to compute reference planning trajectories \citep{wang2014rolling1,wang2014rolling2,gong2016constrained,gong2018cooperative,zhou2017rolling}. 

 \subsubsection{A stochastic environment}

The optimal control framework can be extended in several ways. 
When there are measurement errors or 
when there is only partial observability, 
a measurement equation is introduced into the state-space model  \citep{wang2014rolling1,wang2014rolling2,zhou2017rolling}. 
Considering stochastic communication delay arising from packet drops, 
decomposition is proposed for the stability analysis of a large system of CAVs 
\citep{qin2017scalable,jin2018connected,jin2018experimental}. 

\subsection{Noncooperative control: a deterministic environment}

The noncooperative control of multiple AVs is modeled as N-player game-theoretic models. 
To the best of our knowledge, all the work on multi-AV competitive control assumes a deterministic environment. 

The first group of studies assume there is a small number of AVs to control in specific scenarios such as platooning, in other words, $n$ is a finite number. 
\cite{wang2015game} formulates AVs discrete lane change and continuous acceleration selections as a differential game, where agents' optimal strategies are obtained from solving optimal control problems.
The outcome of a differential game is a dynamic equilibrium. 
Computation of such a dynamic equilibrium involving N players is mathematically intractable when the number of coupled agents becomes large. 
To get around, \cite{wang2015game} decomposes the problem into a finite number of sub-problems and applies MPC to each vehicle. 
\cite{dreves2018generalized} solves a generalized Nash equilibrium by summing up all vehicles objective functions, which is essentially a cooperative control.
Because the game-based control suffer from scalability issues, all the aforementioned studies had to constrain their applications to a limited number of AVs. 
As a growing number of AVs are put on public roads, a scalable and computational efficient algorithm is needed for a large number of AV controllers. 

Another school of reserach assumes a more generic traffic scenario, which is a large number of AVs interacting with one another on a transportation system, in other words, $n$ goes to infinity. 
Mean field game (MFG) has shown to be a scalable model for
the $N$-car differential game, as the AV population grows \citep{huang2019stable,huang2020game,huang2020stable}. 
MFG is a game-theoretic framework to model complex multi-agent dynamics 
arising from the interactions of a large population of rational utility-optimizing agents 
whose dynamical behaviors are characterized by optimal control problems \citep{lasry2007mean,huang2006large}. 
By exploiting the ``smoothing" effect of a large number of interacting individuals, MFG assumes that each agent only responds to and contributes to the density distribution of the whole population. 
It has become increasingly popular in finance \citep{gueant2011mean,lachapelle2010computation},  
engineering \citep{djehiche2016mean}, 
and pedestrian crowds \citep{lachapelle2011mean}.  
%
%
In the longitudinal control of AVs, 
each car solves its optimal velocity backward in time, the aggregate effect of which is formulated by a Hamilton-Jacobi-Bellman (HJB) equation; 
while the mean field approximation derives the evolution of traffic density solved by a transport equation (with many other names like continuity equation, flow conservation equation) forward in time. 
To solve the mean field equilibrium, 
The distributed velocity controller derived from the MFE is shown to be an $\epsilon$-equilibrium of the $N$-car differential game.

\begin{figure}[h]
	\centering	
	\subfloat[\small LWR]{\includegraphics[width=.3\textwidth]{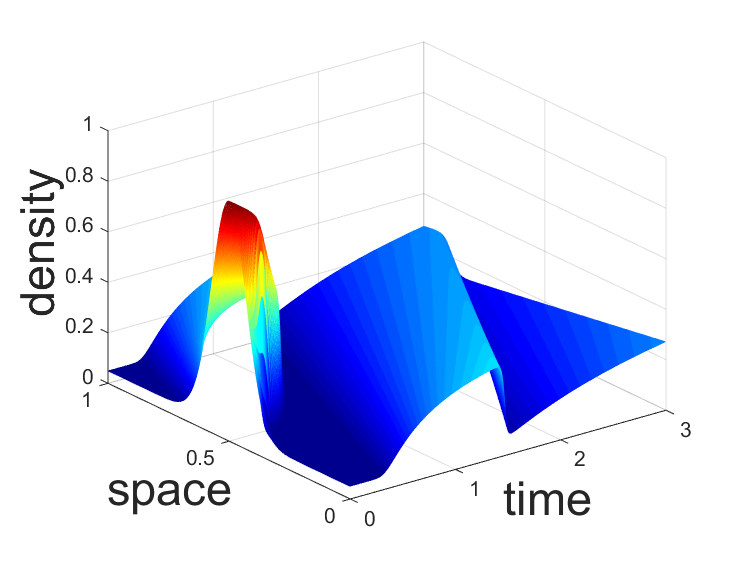}}~
	\subfloat[\small MFG]{\includegraphics[width=.3\textwidth]{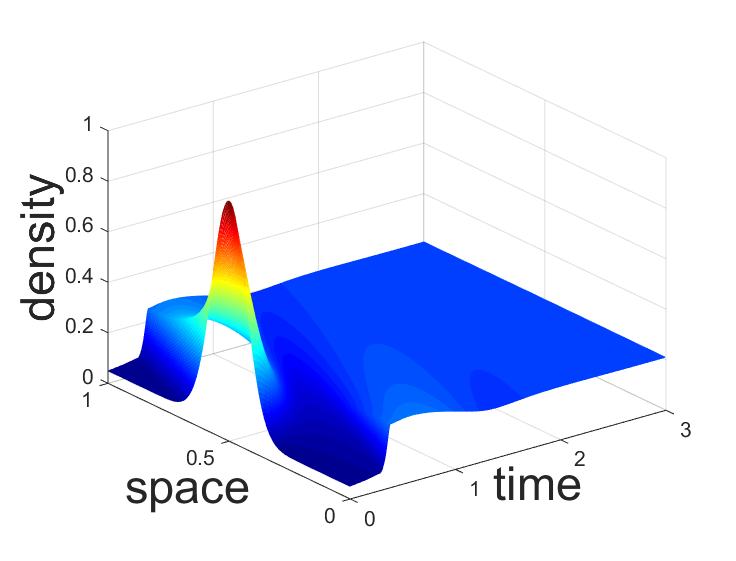}}
	
	\subfloat[\small Velocity]{\includegraphics[width=.3\textwidth]{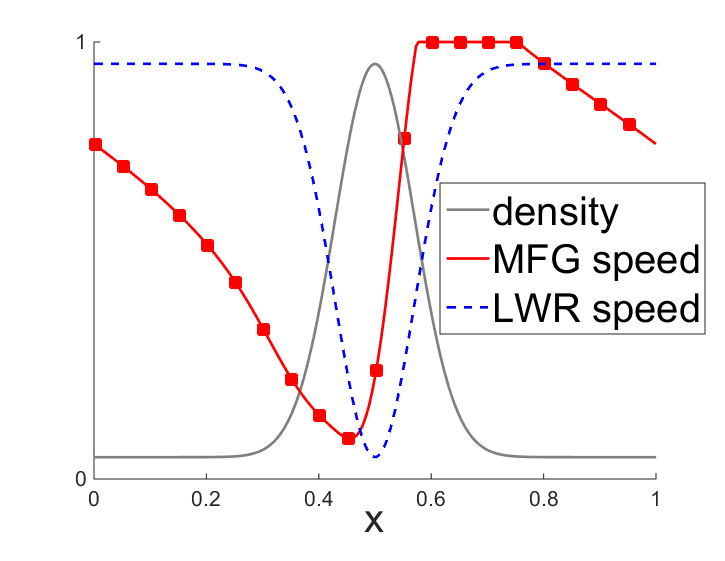}}~ 
	\subfloat[\small Solution gap]{\includegraphics[width=.3\textwidth]{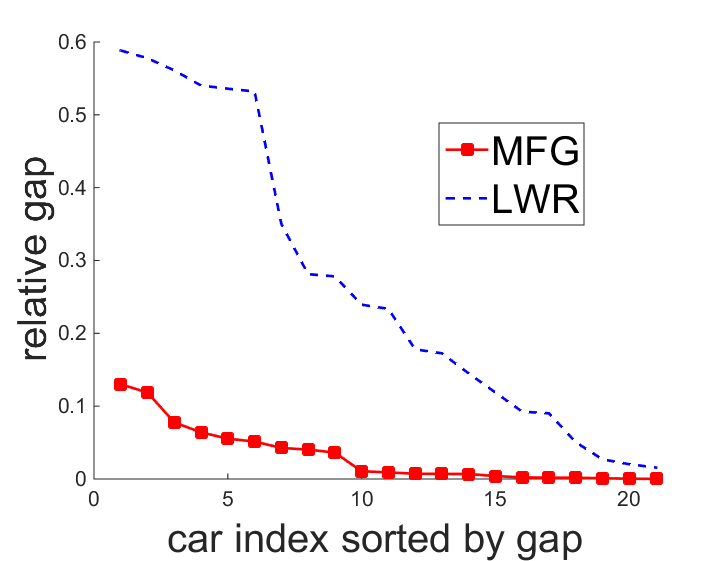}}
	\caption{\small LWR (HV) v.s. MFG (AV)}\label{fig:mfg-lwra}
\end{figure}

\cite{huang2020game} has also established a connection between an MFG-based macroscopic continuum model and the existing traffic flow theory.   
The LWR model,
which implicitly assumes that cars move according to hydrodynamics without modeling driving intent, 
is proved to be 
a myopic MFG with a specially designed objective function.
In conclusion, 
MFG embodies classical traffic flow models with behavioral interpretation, thereby providing a flexible behavioral foundation and a promising direction to accommodate new traffic entities like AVs. 
Under the more intelligent objective function of AVs, 
the LWR velocity does not represent a socially optimal driving strategy as demonstrated by larger deviations from the actual equilibrium in Fig.~\ref{fig:mfg-lwra}(d).  
Fig.~\ref{fig:mfg-lwra}(a-b) illustrate that the MFG mitigates traffic oscillation faster than LWR. 
Fig.~\ref{fig:mfg-lwra}(c) reveals the rationale at one time instant.   
Around a jam area with symmetric traffic density, vehicles driven by MFG controllers tend to slow down farther upstream before joining the jam and immediately speed up after leaving the jam; 
in contrast to those driven by LWR controllers whose speed remains symmetric before and after the jam area. 
This is because 
LWR's velocity is determined only through traffic density at that location, 
while that of the MFG depends on traffic density of the entire horizon.        

\section{n AV + m HV: controllable AVs navigating the HV-dominated traffic} 
\label{subsec:nAVmHV}

As mentioned in the previous section, in this section, we add HVs into the traffic environment where multiple AVs navigate. 
Likewise, we will use vehicle cooperation as the primary category and the environment stochasticity as the secondary category. 


 \subsection{Cooperative control}

When multi-AV control is cooperative, the existing literature covers both a deterministic and a stochastic environment.  

 \subsubsection{A deterministic environment: Mixed Vehicle platooning}

In a mixed traffic platoon comprised of multiple AVs and multiple HVs, how to design a AV controller to stabilize a mixed traffic platoon remains largely unsolved, due to the scalability issue, in other words, the topology of AVs and HVs in a mixed platoon. 

To avoid enumerating various topology of a mixed platoon,  
a majority of studies use a general concept of head-to-tail stability in which the stability of a platoon only depends on the total numbers of AVs and HVs, not their topology \citep{wu2018stabilizing}. 
Using simulations, \cite{talebpour2016influence,yao2019stability} implement CACC on CAVs and investigated the string stability of the mixed traffic system. Different controller parameters and the CAV's penetration rates are tested to illustrate their relations to the stability.

While accounting for the topology of a mixed platoon, 
by decomposing the entire platoon into small subsystems, 
\cite{zhou2020stabilizing} introduce a more practical head-to-tail stability criterion for subsystems and analyzes the mixed traffic system with multiple CAVs and multiple HVs under the new stability criterion. 
\cite{gong2018cooperative} solve a $p$-step MPC instead of a one-step MPC to mitigate the uncertainty of human driver trajectories. 

One alternative approach to address the scalability issues is the PDE approximation \citep{barooah2009mistuning,zheng2016stability}. This approach suggests to study the stability of continuum traffic flow models which are the limits of microscopic models. 
Traffic stability is then defined by whether the deviations on the density and velocity profile from uniform flows are controlled as time increases. \citep{darbha1999intelligent}. 
Building on MFG, \cite{huang2019stable,huang2020stable} analyze traffic stability for mixed traffic, 
assuming that  
HVs are modeled by ARZ 
and AVs are modeled by an MFG. 
Linear stability analysis demonstrates that the MFG traffic flow model behaves differently from traditional traffic flow models.
The impact of AV's penetration rate and controller design on traffic stability are quantified on ring roads. 


 \subsubsection{A stochastic environment}

\cite{wu2017flow,wu2017emergent,wu2017framework,CWu-2018} assume a fully observable system where the goal of multi-AV control is to optimize total system performances, such as velocity, energy consumption. 
A model-free MARL is employed. 
In other words, there is no need to define a state transition matrix explicitly. 
Instead, the state transition is computed from the simulation platform.  
A traffic simulator has been developed in SUMO to simulate HVs and uncontrolled AVs using IDM models. 
Given actions selected by controlled AVs, the simulator updates every car's position based on selected actions of controllable AVs and IDM models of uncontrollable vehicles. 
Then the centralized training and execution with trust region policy optimization (TRPO) policy gradient is implemented to solve for an optimal policy. 
\cite{Kreidieh-2018} trains AVs on a multi-lane ring road and implements transfer learning to execute the AV control on an open multi-lane highway.  



\subsection{Noncooperative control}\label{subsec:nAVmHV_noncoop}

\begin{figure}[H]
	\centering
	\includegraphics[width=0.35\textwidth, keepaspectratio=true]{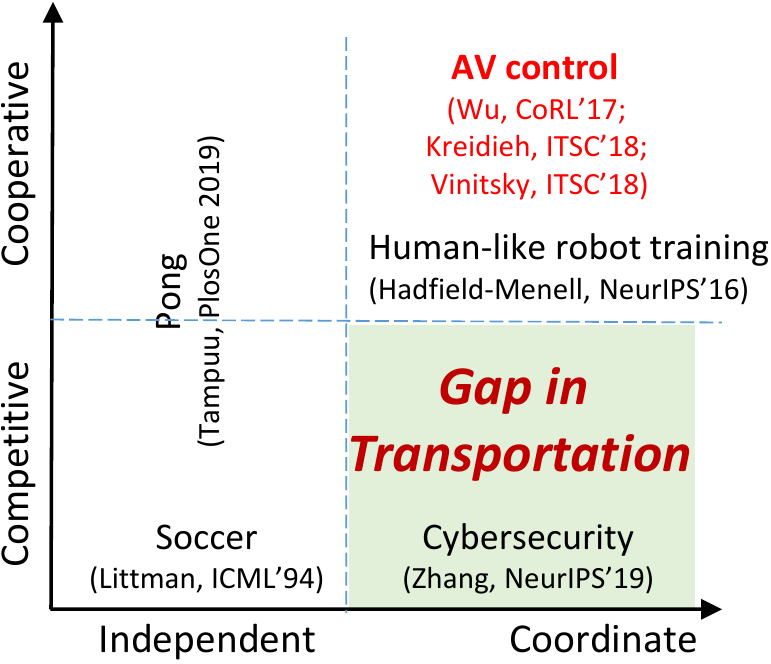}
	\caption{\small Literature on MARL}
	\label{fig:span}
\end{figure}

In a multi-AV system where human drivers exist and dominate the traffic environment, 
uncertainty arises from human driving behavior. 
Controllable AVs have to learn the environment while selecting optimal driving policies with a maximum reward. 
Built upon single-agent RL, multi-agent reinforcement learning (MARL) extends the control of single robot to multiple ones. 
In a multi-agent system with stochasticity and uncertainty, 
MARL becomes a natural tool for control of multiple AVs. 
MARL tasks can be broadly grouped into three categories, namely, fully cooperative, fully competitive, and a mix of the two, depending on different applications \citep{zhang_multi-agent_2019}: (1) In the fully cooperative setting, agents collaborate with each other to optimize a common goal; (2) In the fully competitive setting, agents have competing goals, and the return of agents sums up to zero; (3) The mixed setting is more like a general-sum game where each agent cooperates with some agents while competes with others. For instance, in the video game \emph{Pong}, an agent is expected to be either fully competitive if its goal is to beat its opponent or fully cooperative if its goal is to keep the ball in the game as long as possible \citep{tampuu_multiagent_2017}. A progression from fully competitive to fully cooperative behavior of agents was also presented in \cite{tampuu_multiagent_2017} by simply adjusting the reward. 
Fig.~(\ref{fig:span}) illustrates the classification of MARL based on if agents are collaborative (to optimize a common goal) or competitive (with competing goals) 
and if they share information with others (i.e., being independent or coordinate). 
There exists a void in which multi-AV control in a competitive environment is modeled. 

The future AVs will be manufactured by different companies with different technical specifications. 
It will thus be challenging for AVs to collaborate with a common goal. 
We believe it is reasonable to picture that each AV is an independent and fully decentralized agent with its own goal (e.g., to send its occupant to her destination on a shortest path). 
Below we will lay out a feasible MVS framework where MARL algorithms can potentially be applied to this context. 
Without cooperation and with the presence of human drivers, it is also challenging for AVs to sense and perceive the environment precisely. Therefore we assume a stochastic environment. 
Fig~(\ref{fig:rl}) illustrates the multi-AV control framework. The main different from the single-AV control is the interaction among multiple AVs when they simultaneously explore the mixed traffic environment and select their individual optimal policies. 

The noncooperative multi-AV control problem can be formulated as a 
stochastic game or Markov game \citep{littman_markov_1994}, 
where each agent solves a POMDP. 
A Markov game is 
defined by a tuple $({\cal S}, {\cal O}_1, {\cal O}_2, \cdots, {\cal O}_N$, ${\cal A}_1, {\cal A}_2, \cdots, {\cal A}_N, {\cal P}, {\cal R}_1, {\cal R}_2, \cdots, {\cal R}_N, N, \gamma)$, where $N$ is the number of agents and $S$ is the environment state space. Environment state $\mathbf{s} \in S$ is not fully observable. Instead, agent $i$ draws a private observation $o_i \in {\cal O}_i$ which is correlated with $\mathbf{s}$. ${\cal O}_i$ is the observation space of agent $i$, yielding a joint observation space ${\cal O} = {\cal O}_1 \times {\cal O}_2, \times \cdots \times {\cal O}_N$, ${\cal A}_i$ is the action space of agent $i \in \{1,2,\cdots, N\}$, yielding a joint action space ${\cal A} = {\cal A}_1 \times {\cal A}_2 \times \cdots \times {\cal A}_N$, ${\cal P}:{\cal S}\times {\cal A} \times {\cal S} \rightarrow [0, 1]$ is the state transition probability, ${\cal R}_i:S\times {\cal A} \times {\cal S} \rightarrow \mathbb{R}$ is the reward function for agent $i$, and $\gamma$ is the discount factor. 
The MARL components are:

\begin{itemize}
	
	\item ${\cal S}.$ The state space of the mixed traffic environment. A state $\mathbf{s} \in S$ contains the information of all controllable and uncontrollable AVs and HVs.

	\item ${\cal A}.$ The joint action space, i.e., ${\cal A}_1 \times {\cal A}_2, \times \cdots \times {\cal A}_N$. The action of the $i^{th}$ AV, denoted as $a_i \in A_i$, is a two-dimensional vector, including acceleration and steering angle. The joint action of AVs is $\mathbf{a} = (a_1,...,a_N)$. 
	
	\item ${\cal O}.$ The joint observation space. The joint observation of $N$ AVs, i.e., $\mathbf{o}$, is denoted as $\mathbf{o} = (o_1,..., o_N)$. 
	${\cal O}_i$ is the observation space for the $i^{th}$ AV, yielding a joint observation space ${\cal O} = {\cal O}_1 \times {\cal O}_2, \times \cdots \times {\cal O}_N$. 	
	
	\item ${\cal G}.$ The set of observation functions, i.e., $\{{\cal G}_1,{\cal G}_2,...,{\cal G}_N\}$. 
	For the $i^{th}$ AV, $\mathbf{s}$ may not be fully observable. 
	Instead, it draws an observation $o_i \in {\cal O}_i$ that is correlated with $\mathbf{s}$ according to an observation function ${\cal G}_i: {\cal S} \times {\cal O}_i\rightarrow [0,1]$, i.e., $o_i \sim {\cal G}_i(o_i|\mathbf{s})$. 
	
	%
	
	\item ${\cal P}.$ The state transition function, i.e., ${\cal S}\times {\cal A} \times {\cal S} \rightarrow [0, 1]$. 
	This state transition can be computed from specific form (model-based) or from a mixed traffic simulator (model-free).

	\item ${\cal R}.$ The joint reward space, i.e., ${\cal R}_1 \times {\cal R}_2, \times \cdots \times R_N$. Along with the state transition, car $i$ receives an immediate reward, i.e., $r_i \in {\cal R}_i:{\cal S}\times {\cal A} \times {\cal S} \rightarrow \mathbb{R}$. 
	The reward $r_i$ may include traffic safety (e.g., off-road/collision avoidance), efficiency (e.g., fast speed), and emissions. 
	One aims to maximize its discounted expected cumulative reward by deriving an optimal policy, which is the best response to other AVs' policies. 
	
\end{itemize}

A controllable AV indexed by $i$ aims to derive an optimal policy $\pi^{*(AV)}_i:O_i \times A_i \rightarrow [0,1]$ by maximizing its expected cumulative reward. 
It samples an action from the policy after drawing observation $o_i$. 
After all AVs take actions, the joint action $\mathbf{a}$ triggers a state transition $\mathbf{s} \rightarrow \mathbf{s}'$ based on the state transition probability $P(\mathbf{s}'|\mathbf{s}, \mathbf{a})$. Agent $i$ draws a private observation $o_i'$ corresponding to $\mathbf{s}'$ and receives a reward $r_i(\mathbf{s}, \mathbf{a}, \mathbf{s}')$.  Agent $i$ aims to maximize its discounted expected cumulative reward by deriving an optimal policy $\pi^*_i$ which is the best response to other agents' policies. This process repeats until agents reach their own terminal state. 
Due to the existence of other agents, the Q-value function for agent $i$ 
, i.e., $Q_i$, is now dependent on the environment state $\mathbf{s} \in S$ and the joint action $\mathbf{a} \in A$ of all agents
, i.e, 
$Q_i = Q_i(\mathbf{s}, \mathbf{a})$. 
Similarly, the value function of agent $i$, i.e., $V_i = V_i(\mathbf{s})$, 
is dependent on the environment state $\mathbf{s}$.

\paragraph*{MARL algorithms}

Depending on if AVs can exchange information and learn the environment information, driving policy learning can be categorized into joint or independent learners. 
Local observation leads to independent learners, while information sharing can change AVs' learning behavior to joint learners.

\noindent\textbf{Independent Learners.} 
If AVs only sense neighboring vehicles' information, each AV learns the environment and policies independently. 
From the vehicle perspective, 
vehicle $i$'s objective functional $J^N_i (s_i, u_i, s_{-i}, a_{-i})$ depends on all other vehicles' state and controls. 
The optimal driving strategy for vehicle $i$ ($i=1,\cdots, N$) is computed as \citep{liu2018improving}:
\begin{subequations}
	\begin{align}
	& \min_{a_i} J^N_i(s_i, a_i) \\
	\mbox{s.t. } & \dot{s_i}(t) = f\left(s_i, a_i\right), \\
	& \mathbf{a}(t) \in {\cal A}, \\
	& s_i \in {\cal S}_i (\hat{s}^i_{-i}, \hat{a}^i_{-i}), \forall -i\in{\cal N}_i.
	\end{align}
\end{subequations}
where, \\
${\cal N}_i$: the neighboring vehicles of vehicle $i$;\\
$\hat{s}^i_{-i}(t)$: the state of vehicles other than $i$ estimated by vehicle $i$;\\
$\hat{a}^i_{-i}$: the action of vehicles other than $i$ estimated by vehicle $i$;\\ 
${\cal S}_i (\hat{s}^i_{-i}, \hat{a}^i_{-i})$: the state space of vehicle $i$ given others' states and actions.

Conceptually, most single-agent RL techniques can be directly applied to multi-agent scenarios for independent learners. 
Popular examples include Deep Q Network (DQN)
~\citep{Mnih-2015}, deep deterministic policy gradient 
~\citep{Lill-2015} and soft actor-critic 
~\citep{haaronja-2018}. 
However, difficulties also arise \citep{Omidshafiei-2017,Nguyen-2018}: 
(1) non-stationarity of Q-value estimation due to co-existence of other adaptive AVs; 
(2) invalid theoretical convergence in multi-AV scenarios because the Markovian property may not apply; 
(3) confusing domain stochasticity from both environments and other AVs; 
and more importantly, 
(4) the curse of dimensionality, i.e., the search space in state and action is too large, making the learning intractable. 
	Advanced MARL algorithms are developed to mitigate some of the aforementioned challenges. Decentralized hysteretic deep recurrent Q-networks (Dec-HDRQNs) utilizes different learning rates for different partially-observable domains~\citep{Omidshafiei-2017}. This approach exploits the robustness of hysteresis to non-stationarity and alter-exploration, in addition to the representational power and memory-based decision making of DRQNs. More recently, lenient DQN~\citep{Palmer-2018} is proposed, with which lenient agents map state-action pairs to decaying temperature values that control the amount of leniency applied towards negative policy updates that are sampled from the experience replay. This introduces optimism in the value function update, and can facilitate cooperation in tabular fully-cooperative MARL problems. 

A key challenge arises 
in MARL when independent agents have no knowledge of other agents, that is, the theoretical convergence guarantee is no longer applicable since the environment is no longer Markovian and stationary \citep{matignon_independent_2012}. 
To tackle this issue, one way is to exchange or share information among agents. 

\noindent\textbf{Joint Learners.} 
If AVs receive global information regarding all others' state and action (e.g., via V2V/V2I), they can learn their optimal policies jointly. 
The performance of the agents could be better off through coordination. 

To the best of our knowledge, 
there only exist a small amount of studies \citep{wu2017emergent,wu2017framework,wu2017flow,wu2018stabilizing} on multi-AV control using MARL. 
They assume that all controllable AVs share a common objective, which constitutes a fully observable cooperative MVS taking into account uncontrollable vehicles. The policy network is trained with a TRPO policy gradient method, and transfer learning \citep{kreidieh2018dissipating} is applied to transfer the policy from multi-lane ring roads to highway merging scenarios.

A joint learning framework suffers from the curse of dimensionality, as the agent size grows. 
Thus the centralized learning (i.e., based on global information) and decentralized execution (i.e., based on local observation) paradigm has become an increasingly popular paradigm for independent learners \citep{foerster_learning_2016, lowe_multi-agent_2017, lin_efficient_2018, li_efficient_2019}. 
While training is stabilized conditioning on the information of other agents, 
scalability becomes a critical issue in MARL because the joint state space and joint action space grow exponentially with the number of agents. 
To mitigate the curse of dimensionality, mean field reinforcement learning
~\citep{Yaodong-2018} has become a popular technique, where the interactions within the population of agents are approximated by those between a single agent and the average effect from the overall population or neighboring agents. 
In this way, learning of individual agent's optimal policies depends on the population dynamics, which makes possible a scalable  policy learning for achieving Nash equilibrium in multi-agent environments.
Its potential to multi-AV control could be one direction to explore.

\subsection{n AVs + m HV ($n>>m$): A special case}
\label{subsec:nAV1HV}

There is little research on the AV-dominated world, partly because that it is highly likely that human drivers will adapt their driving behavior when surrounded with AVs. But it remains unclear how such behavior evolves. 
Different hypotheses could drive the evolution of human driving behavior toward opposite directions.  

One hypothesis is that humans may gradually adapt their driving behavior in the presence of AVs and consequently develop moral hazards \citep{pedersen2001game,pedersen2003moral,chatterjee2013evolutionary,chatterjee2016understanding,millard2016pedestrians,chen2019lib}. 
These speculations cannot be validated in the existing market with a too low penetration rate of AVs. 
 	Laboratory driving simulator using driving simulators could serve as a safe and effective alternative \citep{creech2019pedestrian,tilbury2020analysis}.  
 	In spite of the fact that participants could possibly exhibit unrealistic behaviors on a driving simulator, 
 	the value of these simulators should not be ignored for advancing our understanding of people's behavioral adaptation for a future scenario.

\section{Data-driven policy learning}\label{sec:data}

A major challenge in the study of AVs, different from other autonomous systems, is the highly dynamic, uncertain, complex environment in which it navigates. Unlike training robots in a controlled laboratory environment, training intelligent AVs requires them to interact continuously with the traffic environment to learn optimal driving policies. 
Such a traffic environment, primarily comprised of intelligent actors including human drivers and other uncontrollable AVs, 
needs to be learned from real data. 

\subsection{Human driving policy learning}
\label{subsec:HV}

Human movement trajectories are treated as hard safety constraints or boundaries for robots motion planning. 
To this end, accurate and precise models of human behavior are required to ensure safety-critical applications. 
Driving is a complex task.  
It is a sequential decision-making process 
with a complex mapping from the perception of neighboring traffic or the prediction of global traffic environment onto driver actions. 
Human driving behavior has long been studied in the transportation community. 
It has recently gained growing attentions from the control and robotics communities for its importance in designs of AVs that will drive alongside human drivers.

\subsubsection{Dataset}

There are aggregate traffic data and individual trajectory based data. 
Aggregate traffic data are collected from various sensors, 
including 
loop detectors \citep{loop_detector_rakha2010calibration, loop_detector_rakha2002comparison, loop_detector_rakha2003comparison}, 
surveillance cameras \citep{camera_mao2018aic2018,camera_tang2017vehicle,camera_lu2007freeway},  
Bluetooth detection \citep{singer2013travel,onboard_bluetooth_allstrom2014calibration,bluetooth_allstrom2014calibration}, 
roadside radar/LiDAR \citep{lidar_zhang2018background}. 

Emerging traffic sensors, including connected vehicles, smart phones, on-board cameras, and LiDARs, 
are expected to generate terabytes of streaming data daily~\citep{cv_sas}. 
These new datasets would offer new opportunities to understand human driving behavior. 
Collecting real-time vehicle trajectory data, however, is costly and may infringe privacy, 
as it involves placing sensors
inside individual vehicles 
(e.g., naturalistic driving devices continuously collecting vehicle movement information in the real traffic environment 
\citep{onboard_hecker2018end,onboard_hecker2018learning,onboard_hammit2018evaluation,onboard_lidar_flores2018cooperative,onboard_lidar_zhang2018background,onboard_zhu2018modeling}).  
Albeit lower cost, laboratory driving simulators \citep{sadigh2016planning,sadighverifying,abbeel2011inverse,ziebart2008maximum} allow only one driver to test at a time, 
unable to offer realistic experience of interacting with other vehicles on roads. 
To understand the emergent dynamics arising from human drivers requires information of all the vehicles dynamically moving in a traffic stream. 
By far there are only a few such public datasets. 

Next Generation Simulation (NGSIM) is the mostly widely used human driver trajectory dataset. 
It provides all vehicle trajectories across a time span along some multi-lane highways. 
The shortcoming is that no camera images are recorded for each vehicle, which may limit the usage of image features for human driving policy learning. 

Naturalistic data, collected while driving in the real traffic environment, 
provide an non-intrusive approach of personal driving data collection. 
The largest naturalistic dataset has been collected via the
Strategic Highway Research Program (SHRP2) \citep{SHRP2,mclaughlin2015naturalistic,hankey2016description}. 
There were 3,400 participating vehicles instrumented with a data acquisition system recording speed, acceleration, latitude and longitude. Forward radar detects distance and speed relative to other vehicles. Four video views are also available. 
Such a dataset can train a driving policy using camera sensing information.

\subsubsection{Physics-based model parameter calibration}

Human driving behavior includes driving intent identification and prediction of internal states. 
Without communication among one another or via turning on signals, neither the intent nor internal states of neighboring vehicles are unknown and has to be estimated. 
We will first present the estimation of internal states in the car-following behavior, which is extensively studied in the transportation community, and then the prediction of driving intent.

CFMs have been extensively calibrated
using a maximum likelihood approach \citep{maximum_likelihood_hoogendoorn2010generic}, 
Bayesian estimation \citep{bayes_van2009bayesian,kasai2013application,bayes,bayes_and_SPSA,davis2017bayesian}, 
fundamental diagram regression \citep{macro_lsm_qu2015fundamental,macro_stochastic_phegley2014fundamental},
or heuristics \citep{search_SPAP_1,search_GA_1}. 
Most of them are calibrated using a pair of leading and following vehicle trajectories. 
It loses the information of how perturbation in one vehicle may propagate to those far behind in the platoon, thus may not capture instability of traffic. 

With a rising volume of data generated by vehicles and their sensors, the conventional traffic models cannot predict generalizable driving behaviors.
Leveraging big data, 
researchers are able to leverage data-hungry machine learning methods to learn the policies underlying the diverse human driving behaviors. 
In the context of driving, states are observations of a driver's environment and actions are acceleration and steering angle. 
We have seen a growing body of literature characterizing driving behaviors using
(deep) artificial neural networks \citep{NN_sgd_khodayari2012modified,ann,RNN_zhou2017recurrent,huang2018car} 
and
reinforcement learning \citep{NN_DRL_zhu2018human}. 
These models aim to capture various phenomena arising from human drivers, including asymmetric behaviors, traffic oscillations.


Compared to the car-following behavior, lane-change is more challenging to estimate, partly because of intent identification. 
Driving intents, which are intended actions, 
can be represented by discrete categories, including 
driving straight with a constant speed or acceleration or deceleration (or lane-keeping), 
turning (or preparing to change lanes), 
and changing to its left or right lane (or lane-changing). 
The driving intent estimation problem is commonly modeled as a classification problem, which will be discussed in the next subsection. 
However, 
one school of researchers argue that human's unpredictability, randomness, and non-Markovian property makes it infeasible to learn true dynamics \citep{driggs2017integrating}. 
Instead, task-specific Bayesian optimization \citep{bansal2017goal}, 
stochastic reachable set \citep{driggs2018robust}, non-parametric driver model \citep{driggs2017integrating}, 
and probabilistic approaches \citep{bouton2017belief} have been developed.
Humans usually convey intent through motion, which plays a crucial role in social interactions \citep{becchio2012grasping}. 
Built upon such understanding,  
\cite{driggs2016communicating} assume drivers tend to follow some nominal trajectory, given by the spatial empirical distributions on a cost map.  
Accordingly, the lane-change intent can be formulated as an optimal control problem (and can be reduced to an MPC control). 
The parameters of the control objective function are estimated using 10 subjects' 200 lane-change trajectories. 
\cite{bansal2017goal} learns human dynamics via Bayesian optimization. 
The learned dynamic model is the one that achieves the best control performance for the task at hand but could be different from the true dynamic. 
\cite{driggs2017integrating,driggs2018robust} solves a mixed integer linear program
to estimate a stochastic reachable set that encapsulates the likely trajectories of human drivers intent 
and  
this model can generate trajectories that are similar to those performed by humans. 
%
%


Physics-based models simplify the complex decision-making processes of human beings 
and may lack predictive powers due to its open-loop procedure of parameter estimation. 

\subsubsection{AI-based methods}

Estimation of discrete human intent can be essentially formulated as a classification problem. 
Support vector machine (SVM) \citep{aoude2012driver}, 
hidden Markov model (HMM) \citep{li2016lane}, 
dynamic Bayesian Networks \citep{kasper2012object}, 
and Bayesian filtering (BF) \citep{li2016lane}
are commonly used for online classification of human intent. 
Features used for classification include
longitudinal acceleration, deceleration light, turn signal, speed relative to traffic flow \citep{liu2015safe,liu2016enabling},  
steering angle, lateral acceleration, yaw rate \citep{li2016lane}, 
and lane occupancy \citep{kasper2012object}.
%
%

Once human intentions are known, 
the internal state of a vehicle, i.e., its future trajectory, is estimated using
Gaussian mixture models \citep{wiest2012probabilistic}, 
dynamic Bayesian Networks \citep{gindele2010probabilistic}, 
Kalman filter 
with parameter adaptation algorithm 
\citep{liu2015safe,liu2016enabling}.  

Vehicle behavior estimation and prediction is built upon 
vehicle detection and tracking that happens within one's perception system \citep{sivaraman2013looking}.  
Vision-based or feature-based tracking is widely used to detect the presence of moving objects and associate vehicles between frames \citep{darms2008classification}.  
These vehicle tracking techniques provide a foundation for end-to-end (or perception-to-control) training of autonomous driving policies  \citep{amini2020learning}. 



%
%

The mainstream research on human's driving policy learning is imitation learning, 
which will be primarily discussed subsequently.

\paragraph*{Imitation learning}

Imitation learning (IL) approaches learn the policy directly from expert demonstration data in order to behave similarly to an expert. 	
Popular IL approaches include 
BC~\citep{Pomerleau-1989, Bojarski-2016, Syed-2008}, 
inverse reinforcement learning (IRL)~\citep{abbeel-2004, gonzalez-2016,sadigh-2016}, 
and generative adversarial imitation learning (GAIL)~\citep{Ho2-2016}. 

In early attempts to model human driving behavior, BC formulates IL as a supervised learning problem 
and directly learns a mapping from states to actions using available datasets~\citep{Pomerleau-1989}. 
Compared to rule-based models, the advantage of these systems is that no assumptions are made about road conditions or driver behaviors. While BC approaches are conceptually sound~\cite{Syed-2008}, they may fail in practice when there are states and conditions unrepresented in the dataset. As a results, even the post-trained policy model performs well on the observed states, small inaccuracies will compound resulting in cascading errors~\citep{Ross-2010}. In the case of driving behavior, for example, when the vehicle drifts from the center of the lane, a human driver should correct itself and move back to the centre. However, since this condition does not happen very often for human drivers, data on the correcting action is scarce, resulting in the cascading error problem, and the learned policy will continue to deviate from the center and drive off-road. 
dataset aggregation (DAgger) is one popular technique to mitigate propagation errors of BC 
by augmenting original training data with expert demonstration for missing states \citep{ross2011reduction}. 
Assuming human drivers follow hierarchical reasoning decision-making,  
\cite{tian2019game} employs DAgger to establish a mapping from the ego car's state, all others' state, and the ego car's reasoning level $k$ to the ego car's level $k$ action. 
To accommodate heterogeneity in human drivers, different HVs are assumed to follow different reasoning levels.  

Instead of directly learning actions from observed states, 
IRL estimates one's underlying reward function that drives observed actions, thus avoiding the issue of missing states. 
Assuming that the expert follows an optimal policy with respect to an unknown reward function, 
IRL \citep{ng2000algorithms,abbeel2004apprenticeship,abbeel2011inverse} and its variants \citep{ziebart2008maximum} 
have become increasingly popular 
to learn optimal sequential policies from expert demonstration. 
In general, IRL attempts to recover the reward function prior to finding the policy that behaves identically to the expert. Because the recovered reward function extends to unseen states, the corresponding policy can generalize much more efficiently and mitigate the cascading errors from which BC approaches suffer. For example, when driving on the highway, the vehicle knows to return to the centre of the lane when it is close to the side, because the reward function gives a high penalty in this situation. As to BC, due to scarce learning samples of driving at rare situations, such as driving at the side of the road, this would be a problem for BC to handle these situations. IRL has been used for modeling human driving behavior~\cite{gonzalez-2016,sadigh-2016}. 
In particular, the reward function is specified as a linear combination of features (or a DNN) 
\citep{sadighverifying,song2018multi,biyik2018batch,abbeel2011inverse}. 
\cite{sadigh2016planning} employs a continuous-time version of IRL, which is the continuous inverse optimal control with locally optimal examples \citep{levine2012continuous}. 
\cite{schwarting2019social} models heteorgenity in human drivers by introducing a social preference value into one's reward function. 
An online IRL learning algorithm is developed for the AV to learn such value while interacting with HVs. 
Despite the increasing potential in imitation learning, IRL approaches are typically computationally expensive toward recovery of the expert reward function (or cost function).

Instead of learning the expert cost function directly and learning the policy based on it, recent work has attempted to learn the expert behavior through direct policy optimization and skip the step of cost function recovery. These methods have been successfully applied to modeling human driving behavior~\citep{Ho-2016}. 
With the advent of the generative adversarial network (GAN)~\citep{goodfellow-2014} and generative adversarial imitation learning (GAIL)~\citep{Ho2-2016}, new policy learning methods have become available, performing well on certain benchmarking tasks. GANs are based on a two-layer minimax game where one network acts as a discriminator to learn the difference between real and generated samples. The second network, i.e. the generator, is to generate fake samples to fool the discriminator. The goal is to find a Nash-equilibrium of the racing game between the generator and discriminator. More recently, Wasserstain GAN~\citep{martin-2017} is proposed to replace the standard KL divergence objective with Wasserstein distance, which solves the mode collapse issue in standard GAN. GANs can be extended to imitation learning domain by replacing the generator with the policy network, i.e., the action generator given states. The generator generates actions based on a learned policy, which is derived via the the objective of fooling the discriminator. The discriminator distinguishes between the generated actions and expert actions given states. 
GAIL uses the GAN technique in combination with TRPO. TRPO updates the policy within a properly bounded region, and based on which, a monotonic improvement in policy over iterations is guaranteed~\citep{Schulman-2015}. For more stable training, generalized advantage estimation (GAE)~\citep{Schulman2-2015} is  used to adjust variance-bias trade-off and reduce the variance in learning. TRPO combined with GAE is able to learn complicated high-dimensional control tasks~\citep{Schulman2-2015}. 
GAIL combined with recurrent policy learning~\citep{Daan-2010, heess-2015}, in particular, has been used for modeling human driving behavior, achieving advanced results~\citep{Kuefler-2017}. 
Later on, other algorithms combining GAIL with Wasserstein GAN (WGAIL), and gradient penalty (WGAIL-GP)~\citep{Gulrajani-2017} are explored and show improved performances in some conditions compared o standard GAIL~\citep{Diederi-2018}.

GAIL has been used to imitate human driving behaviors~\citep{Kuefler-2017,Mykel2-2018,bhattacharyya2019simulating}. 
It however may not reflect realistic human driving behaviors. 
For instance, humans on a straight road would drive without maneuvering steering wheels, but the trained driving polices could alternate between small left and right wheel-turning actions, indicating instability of policies. 
Moreover, the newest algorithms, such as WGAIL-GP proposed in~\cite{Diederi-2018}, have not been thoroughly evaluated to form a well-built conclusion about the performance for modelling driving behavior, and are open to further practice in future study.

In summary, 
when only a small portion of states are visited in training datasets, BC suffers from cascading errors in prediction over that unseen states. 
IRL mitigates the cascading error issue by learning an expert's unknown reward function, because the inferred reward would provide a feedback to the learned actions generated from unseen states. 
However, IRL approaches are typically computationally expensive.
Instead of learning reward functions, GAIL learns expert behavior through direct policy optimization~\citep{Ho-2016}. 
GAIL can extract a generalizable policy from limited driving scenarios compared to BC, and has a relatively faster learning speed compared to IRL. 
We believe GAIL could be one promising tool for human driving policy learning.

\subsection{Autonomous Driving Models for Uncontrollable AVs}
\label{subsec:AV}

Uncontrollable AVs refer to those AVs that interact with controllable AVs in the traffic environment but cannot be controlled, probably because they are manufactured from different companies and their driving algorithms are unknown to the host AV. 
Their driving behavior also needs to be learned by controllable AVs. 
Unfortunately, researchers' inaccessibility of AV data has greatly hindered such understanding. 
Due to manufacturers' proprietary protection, however, no documentation has revealed how the existing AVs are actually programmed to drive and interact with other road users on public roads. 
In this subsection, we strive to provide some insights into how researchers may leverage some public datasets collected for computer vision to model the driving behavior of existing AV fleets on public roads. 

\subsubsection{Dataset}

Researchers should be very careful when they claim an AV dataset or when they need to seek some AV related data,  
because most public AV datasets are actually collected by HVs. 
We summarize a non-exhaustive list of AV datasets in Table~(\ref{tab:AV-dataset}). 
These data are collected by vehicles equipped with a variety of sensors, such as radar, LiDAR, GPS, cameras, and inertial measurement units (IMU).
These sensor data, if collected from HVs, are solely 
used to train computer vision algorithms for object detection, segmentation, 3D tracking, pedestrian detection, and Simultaneous Localization and Mapping (SLAM). 
Once trained, these computer vision algorithms are mounted to AVs for testing. 
Fortunately, there exist several public datasets collected directly from AVs, that were pre-trained by academic institutes or AV technology companies. It would be a good strategy for academic researchers to make use of these public AV-collected datasets to learn uncontrollable AV models for simulation, which might behave similarly to existing AV fleets.

\begin{table}[H]
	\centering\caption{Public AV related datasets (partly adapted from \cite{zhou2019longitudinal})} 
	\label{tab:AV-dataset}
	\begin{tabular}{p{1 cm}||p{2 cm}|p{2.8 cm}|p{4.5 cm}|p{1.5 cm}|| p{2 cm}}		\hline
		 Data collection vehicle & Dataset & Purpose & Sensor setup & Location & Institute \\  \hline\hline
		 \multirow{6}{*}{\parbox[t]{1 cm}{HV}} & KITTI \citep{Geiger-2012} &  3D object detection tracking & grayscale/ color cameras, a rotating 3D laser scanner, GPS, IMU & Karlsruhe & Karlsruhe Institute of Technology \\  \cline{2-6} 
		 & KAIST \citep{Choi-2018} & object detection, drivable region detection, depth estimation &  2 RGB $\&$ 1 thermal camera, 1 integrated GPS/IMU device & Seoul & Korea Advanced Institute of Science and Technology \\
		 \cline{2-6} 
		 & H3D \citep{Patil-2019} &   3D detection, 3D multi-object tracking &   GPS/IMU device, a LiDAR, 3 cameras & San Francisco & Honda Research Institute  \\
		 \cline{2-6} 
		 & A2D2 \citep{A2D2} &  3D semantic segmentation, object detection   &   5 LiDARs, 5 surround cameras  & Germany  &  Audi AG  \\
		 \cline{2-6} 
		 & ApolloCar3D \citep{Xibin-2019}  &  3D car instance understanding & GPS,  2 laser scanners, 6 video cameras, a combined IMU/GNSS system, LiDARs &  Various cities in China & Baidu \\
		 \cline{2-6} 
		 & nuScenes \citep{Caesar-2020} &  3D detection, tracking &  6 cameras, 5 radars and 1 LiDAR, IMU, GPS & Boston, Singapore & Aptiv Autonomous Mobility (Aptiv) \\
		\hline\hline
		\multirow{6}{*}{\parbox[t]{1 cm}{AV}} & A*3D~\citep{Pham-2019} & 3D object detection & 2 Chameleon3 USB3 cameras, 1 Velodyne 64-beam 3D-LiDAR & Singapore & Agency for Science, Technology And Research (A*STAR) \\ \cline{2-6} 
		& Argoverse \citep{Chang-Ming-Fang-2019} & 3D tracking and motion forecasting & 2 long-range LiDARs, 9 cameras for $360^{\circ}$ coverage, GPS and other localization sensors & Pittsburgh, PA; Miami, PT & Argo AI and Carnegie Mellon Univ. \\  
		\cline{2-6} 
		& Lyft L5 \citep{Lyft-2019} &  perception systems, motion prediction & 2 40-beam and 1 64-beam LiDARs, $360^{\circ}$ cameras built in-house, a long-focal camera points upward & Palo Alto, CA & Lyft level~5 self-driving system  \\ 
		\cline{2-6} 
		& Waymo Open \citep{Waymo-2019} & 2/3D object detection, 2/3D tracking & 1 mid-range LiDAR, 4 short-range LiDARs, 5 cameras (front and sides), IMUs & Various places in USA & Waymo self-driving cars \\   \hline
	\end{tabular}
\end{table}

To train driving behavioral models of AVs (i.e., end-to-end driving policies), we need not only data from sensors mounted for computer vision, but also driving data directly collected from AVs' motion sensors. 
Thus, some of AV-collected datasets in Table~\ref{tab:AV-dataset} might not be ideal for AV policy training in their raw format. 
For example, A*3D and Argoverse did not provide acceleration, which requires inference using other sensor information, such as GPS. Fortunately, Waymo and Lyft datasets provide complete acceleration records, based on which AV policy training can be made.

To the best of our knowledge, Waymo/Lyft data \citep{Waymo-2019,Lyft-2019} are the only two public datasets on how Level-5 automated vehicles drive and interact with other road users on public roads. 
Both sets are composed of sensor data collected from accelerometer (i.e., IUM), camera, and LiDAR. 
They were originally released for the purpose of object detection and tracking algorithm development. 
These datasets also offer valuable insights into the AV driver models and the interaction between AVs and the environment. 
Such datasets are however distinct from conventional traffic data that our transportation community is used to handle and thus novel methods are required. 

\begin{figure}[H]
	\centering 
	\subfloat[Camera and Lida data]{\includegraphics[scale=.6]{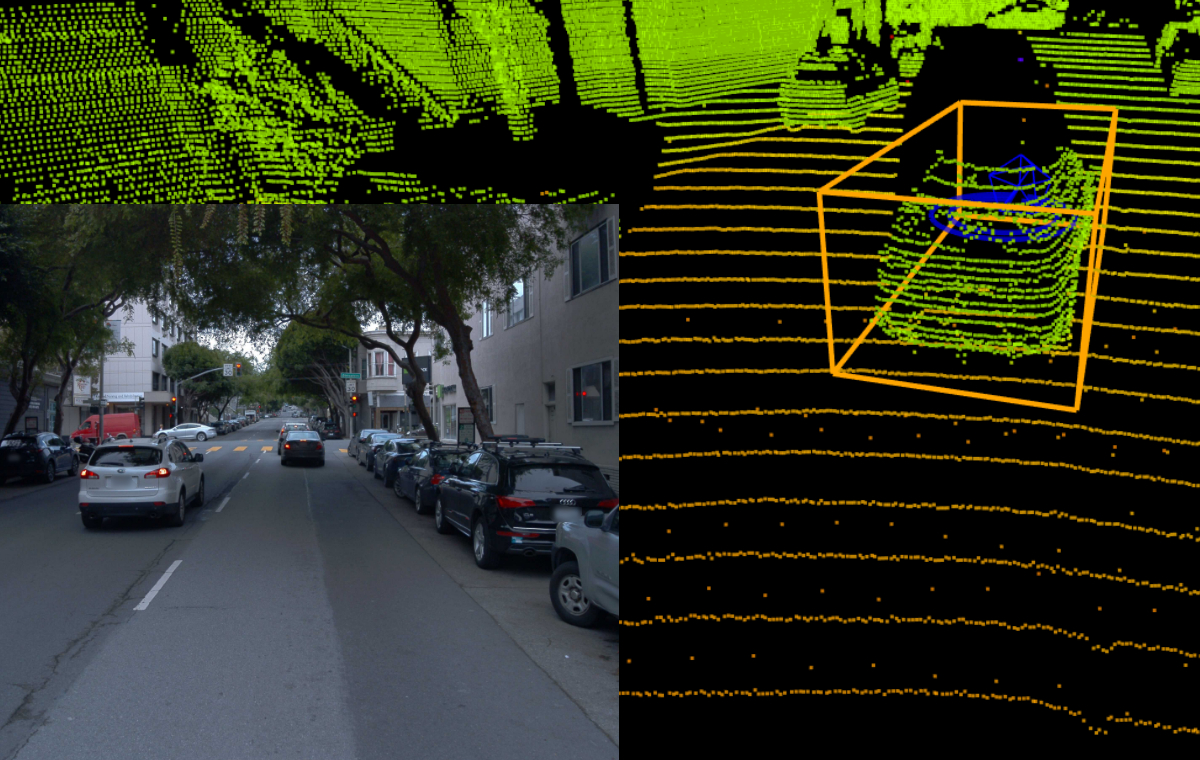}\label{subfig:cam_li}}~
	\subfloat[Data description]
	{\includegraphics[scale=.6]{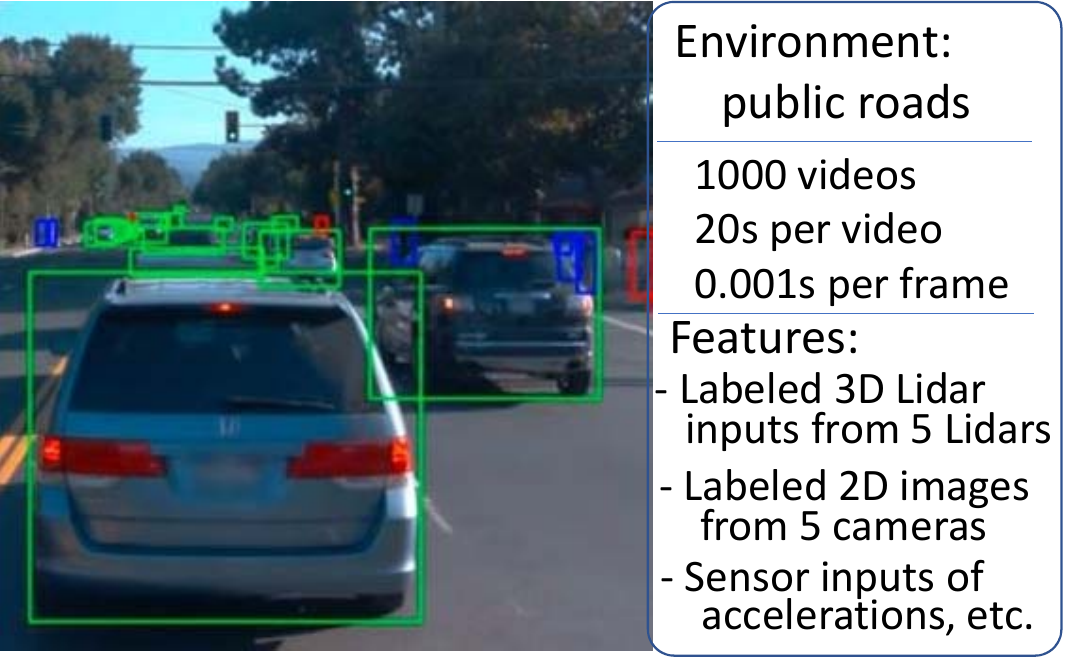}\label{subfig:waymo2}}
	\caption{Waymo self-driving car data \cite{li2017platoon})}
	\label{fig:waymo-fig} 
\end{figure}

We discuss Waymo Open dataset as an example.
As shown in Figure~(\ref{subfig:cam_li}), 
camera images and LiDAR point clouds were collected when a Waymo car was driving on public roads. 
2D/3D bounding box labels were included in camera/LiDAR data in each 
time frame. 
Figure~(\ref{subfig:waymo2}) illustrates the statistics of the dataset.  
It contains 1,000 driving videos, each with a duration of approximately 20 seconds. 
There are a total of $200$ million image frames. 
Built-in accelerometer sensors recorded accelerations taken by a car in each frame, making possible the retrieval of driving policies (that maps surrounding conditions to action of steering and acceleration) programmed in Waymo cars.  


\subsubsection{AV driving model for Waymo cars}

\begin{figure}[H]
	\centering 
	\subfloat[In the first frame of the video, this is a typical "car-following" scenario: the AV follows the truck steadily, and the initial value of the acceleration is around zero.]{\includegraphics[scale=.2]{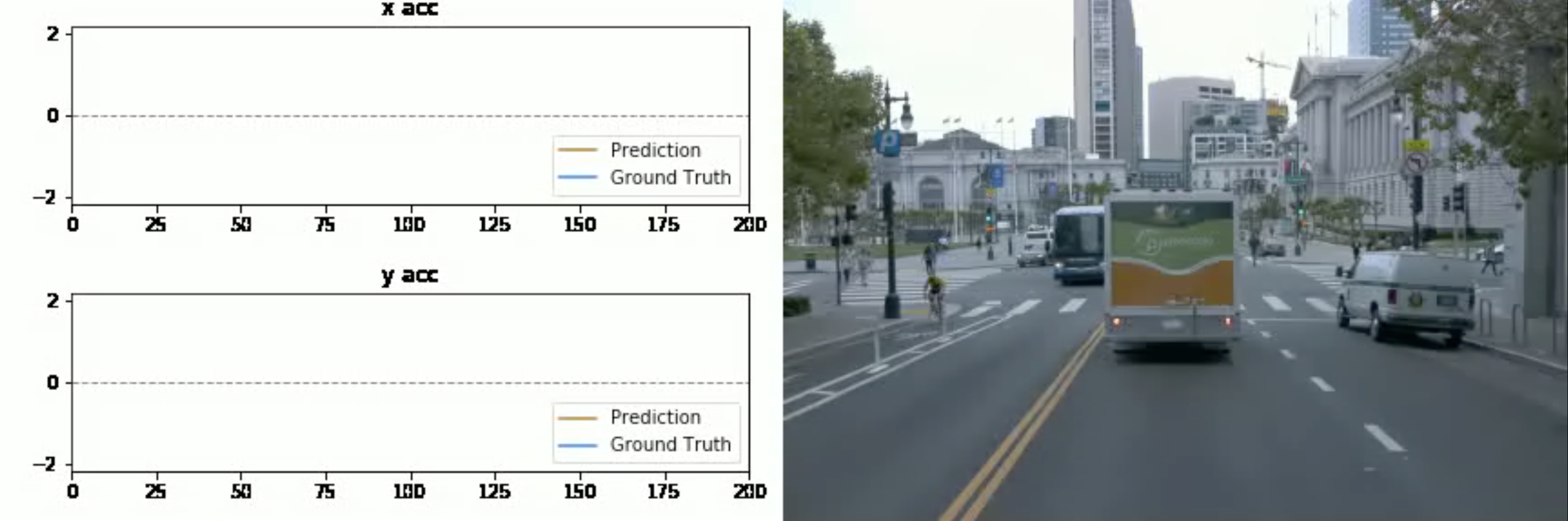}\label{subfig:vis1a}}
	
	\subfloat[In the 108th frame of the video, the truck turns left and at this moment, there is no detected front car and the AV accelerates. The ground-truth and the prediction acceleration curves climb simultaneously.]
	{\includegraphics[scale=.2]{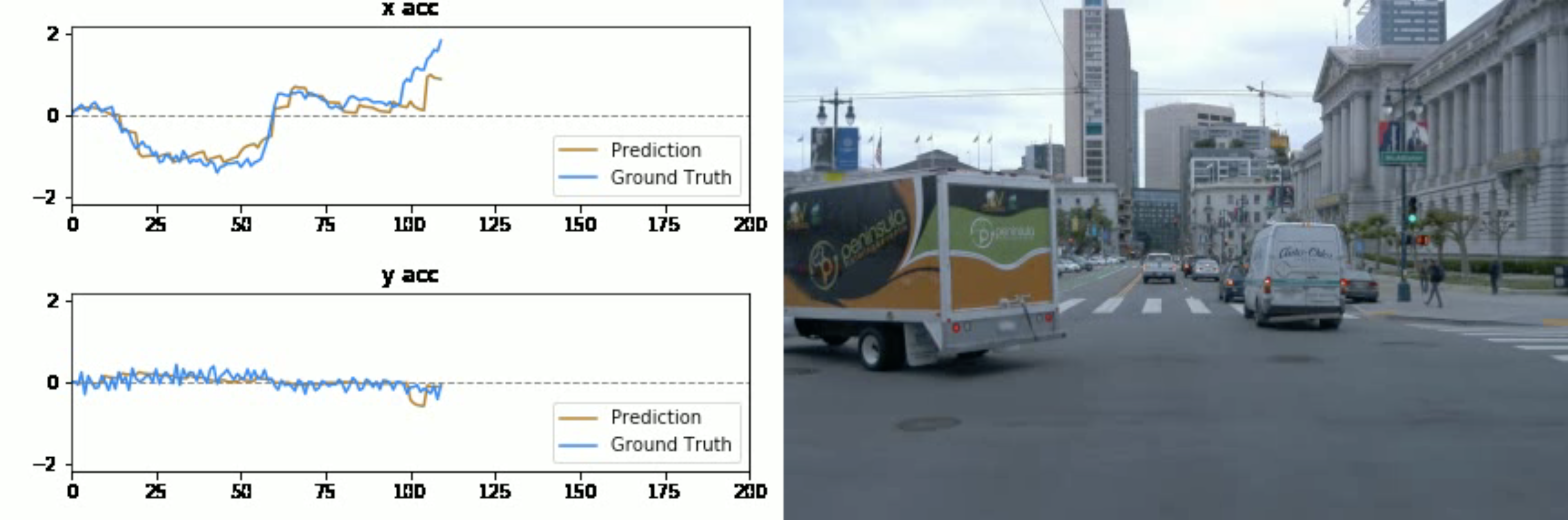}\label{subfig:vis1b}}
	
	\subfloat[In the last frame of the video, the declining acceleration curves shows that the AV brakes accordingly as the front car stops due to the traffic ahead.]
	{\includegraphics[scale=.2]{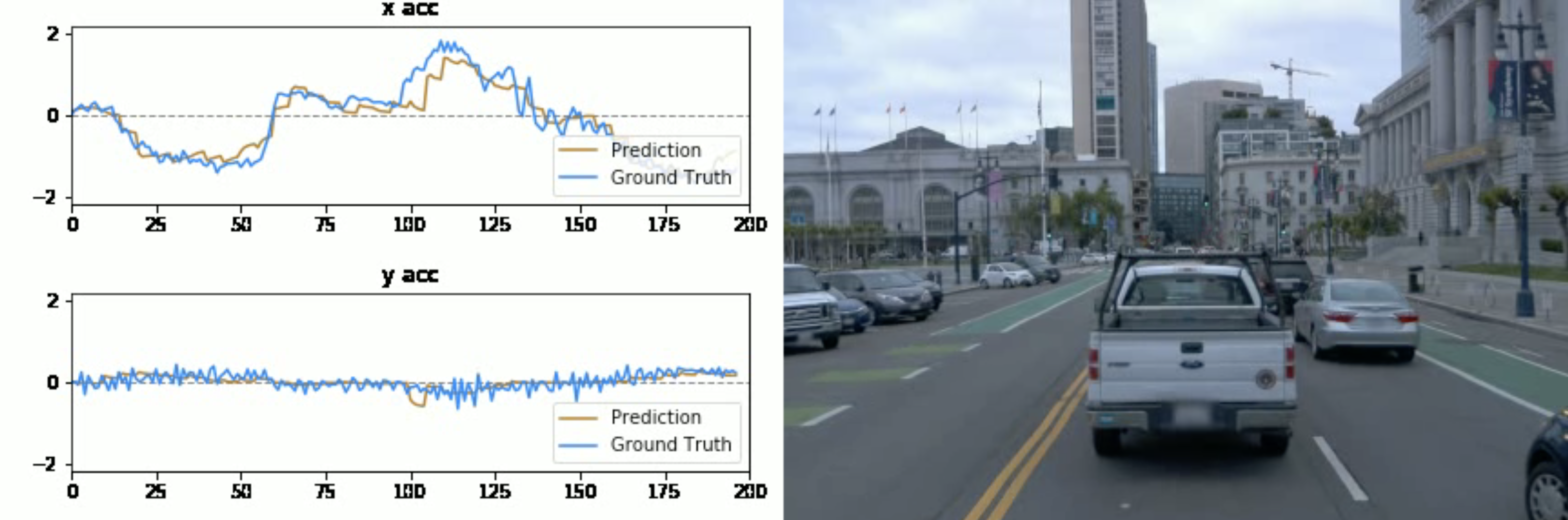}\label{subfig:vis1c}}
	
	\caption{LSTM prediction on longitudinal and lateral accelerations (left) 
		and video snapshot captured from one AV camera (right) 
		(Three key frames are extracted from segment-10289507859301986274 from tar validation\_0001.)}
	\label{fig:vis-images} 
\end{figure}

Leveraging Waymo's sensor data, 
\cite{gu2020lstm} apply BC to learn generalizable autonomous driving polices for two reasons:  
First, AVs are assumed to follow the same driving policies in the same traffic environment, which is different from human drivers who behave highly heterogeneously. 
Second, Waymo datasets cover a wide range of traffic scenarios, including on highways or urban streets, at intersections with traffic lights or stop signs,   
Car-following scenarios were selected from a vast amount of Waymo video data to validate the algorithm performance. 
An LSTM-based learning model is trained, which takes sensor inputs from accelerometer and camera of the past ten frames and predicts acceleration for the next frame. 
Figure~(\ref{fig:vis-images}) illustrates three scenarios in one video: 
the ego car follows a truck, 
the truck leaves, 
and another leading car decelerates. 
This model could be a basis to build a mixed traffic environment that captures the interactions between AVs and their sounding environment. 

 \subsubsection{AV Simulators in Mixed Traffic}
 \label{subsec:val}
 
It is crucial to validate efficiency and safety of designed AV controllers. 
High-risk and high-cost of real AV test urgently requires the development of a mixed traffic simulation environment for virtual testing of AVs. 
We want to stress that there are tons of ``AV simulators" out there but their settings and purposes differ significantly. 
Researchers should be aware of the types of traffic simulators in the market and select the one tailored to their own purposes. 

We divide the existing AV simulators into two primary types based on their purposes: 
one for traffic performance assessment implemented with preprogrammed physics-based CAV driving models, 
and the other for AV driving policy training and evaluation by creating a static or interactive environment. 
The former encode the already calibrated driving models (such as IDM) of each agent to simulate outcomes without further updating these driving models, 
while the latter require continuous interactions of the trained AV system with the environment. 
The transportation community is focused on the first category aiming to evaluate the impact of AVs on traffic congestion or emission, 
while the robotic community has widely used the second category for RL based AV policy training. 



The first category of simulators include VISSIM \citep{ptv} and AIMSUN \citep{aimsun}, 
which have accommodated AV driving modules. 
VISSIM, compatible to vehicle dynamics simulators including CarMaker, is able to simulate the full spectrum of vehicle automation from Level 1 to 5. 
Aimsun Auto can be integrated with sensor testing tools and vehicle dynamics simulation tools, such as Simcenter PreScan,
These simulators are good to answer systematic questions likes the impact of various AV market penetration rates on traffic performances and tipping points, 
which would help operators and policy makers better assess the impact of AVs on safety, mobility, and sustainability. 
But they suffer from two issues: 
First, vehicle motion is simulated using physics-based traffic models, which limits their potential to include AI-based controls trained with high-dimensional features. 
Second, 
these simulators cannot update driving models online. Driving parameters have to be calibrated offline before running online simulation. This may prevent CAVs from adapting to the imminent traffic environment. 

The second category of simulators can be further divided into two types: game-playing and  customizable (see Table~\ref{tab:AV-sim}). Game-playing AV simulators are task-oriented with specified tasks for AV players to accomplish. It is not allowed to change the environment and surrounding vehicles because they are pre-programmed and fixed. TORCS~\citep{wymann} is a racing simulator, which provides real-time observations like speed, position on roads, distance to proceeding car, and image. This simulator has been used for AV training in the lane-keeping and racing scenarios~\citep{chenyi-2015,Sallab-2017,yang2017feature}. Another popular racing simulator currently used for learning a racing AV policy is World Rally Championship 6 (WRC 6)~\citep{Perot-2017, Jaritz-2018}. WRC 6 provide front view image and speed information for players to control steering, brake, and gas. Compared to TORCS, WRC 6 has a more realistic physics engine. In addition to racing simulator, AV communities recently extend their work to action-adventure games, such as Grand Theft Auto V (GTA V), in which multiple vehicle-related missions need to be completed~\citep{Richter-2016,Johnson-Roberson-2017,richter-2017}. 

Customizable AV simulators provide APIs for users to design their environments, surrounding vehicles, and sensor suites at will. Once setting up the initial scenarios, the built-in engine will simulate the involved vehicles and traffic scenarios of interest. Various AV training tasks, such as object detection, lane keeping and collision avoidance, can be made on the platforms provided by customizable simulators. SUMO~\citep{SUMO} is a typical example, with which users can design road networks, surrounding vehicles' policies and sensor systems. SUMO can provide state information, such as speed, location on a road, 2D top-down image of the road, and the state of other designated vehicles in the simulation. CARLA~\citep{Jianyu-2019, Codevilla-2018} has a 3D engine, and can provide more realistic traffic simulation, though the computation is heavier. CARLA also has a flexible setup of sensor suites and signals, including GPS, collision, LiDAR, 3D images and etc. The 3D road traffic can be configured in the simulator. FLOW \citep{Wu-2017} is another customizable simulator, which utilizes SUMO based engine. FLOW incorporates RL libraries, such as rllab and RLlib, and thus, it has convenient interface for RL developers to train and evaluate their AV policies in the simulation provided by FLOW. FLOW has recently been used to train muti-AV policies in a mixed traffic environment~\citep{Wu-2017,Jang-2019}. 

\begin{table}[H]
	\centering\caption{Simulators for training AV policies} 
	\label{tab:AV-sim}
	\begin{tabular}{p{3.1 cm}||p{2.5 cm}|p{4.5 cm}||p{4 cm}}		\hline
		 Type of AV Sim. & Simulator & Features & Reference \\  \hline\hline
		 \multirow{6}{*}{\parbox[t]{3 cm}{Game-playing (vehicle sensor, environment and surrounding vehicles are fixed)}} & TORCS   & speed, position on the road, distance to proceeding car, 3D image  & \cite{wymann} \\ \cline{2-4} 
		 & World Rally Championship 6 (WRC 6)   & 84x84 front view image, speed   & \cite{Perot-2017, Jaritz-2018} \\ 
		 \cline{2-4} 
		 & Grand Theft Auto V  (GTA V)   & images from various viewpoints of vehicle   & \cite{Richter-2016,Johnson-Roberson-2017,richter-2017} \\ 
		\hline\hline
		\multirow{6}{*}{\parbox[t]{3 cm}{Customizable (can configure vehicle sensor systems, environment specification, controlling of surrounding vehicles)}}
		& SUMO & 2D top-down view of road, speed, location on roads, state info. of designated surrounding veh., and etc.  & \cite{SUMO} \\ \cline{2-4} 
		& CARLA & GPS, speed, acceleratin, collision sensor, LiDAR, 3D images, and other sensor suites  & \cite{Jianyu-2019, Codevilla-2018} \\ \cline{2-4}
		& FLOW   & Same to SUMO's features   & \cite{Wu-2017, Jang-2019} \\ \hline
	\end{tabular}
\end{table}

 \subsection{AV-HV Interaction}	\label{subsec:interact}
 
Ideally how one AV interacts with HVs and other AVs should be learned from data. 
But the lack of such data along with an extremely low penetration rate of AVs makes this task infeasible. 
All the existing studies resort to theoretical modeling. 
We will first point out what needs to be learned if we have data 
and then turn our attention to modeling approaches.

 %
 %
 %
 %
 \begin{enumerate}
 	\item \textbf{HV-HV:}  
 	How human drivers interact with one another has been extensively studied in the existing literature. 
 	Its characterization uses both physics-based and AI-based methods (detailed in Section~\ref{subsec:HV}). 	
 	
 	\item \textbf{HV-AV:}  
 	How human drivers react to the presence of AVs has two directions: 
 	(1) Most studies assume that HVs drive the same way as they do in the pure HV traffic environment. 
 	In other words, even when they encounter AVs, they cannot identify AVs and interact as if AVs were HVs. 
 	(2) If HVs have the capability of identifying AVs, AVs are essentially another vehicle class and HVs may likely interact differently. 
 	On one hand, the interaction of heterogeneous vehicle classes can provide comparative studies \citep{ossen2011heterogeneity}. On the other hand, AVs are fundamentally different from other vehicle types propelled by human drivers and may transform humans' car-following behavior significantly.    
 Unfortunately, due to lack of behavioral data for HV-AV scenarios, this arena is understudied. 

A limited number of studies have all pointed out that humans' behavioral reaction to AVs highly depends on their trust to the technology. 
\cite{Farber-2016} studies the importance of communications among road users, such as eye contact, gestures, or anticipatory behavior, on the safety of human drivers. Thus, the challenge arises when humans attempt to communicate but cannot get feedback from AVs. 
As a result, humans may not predict what will happen and behave cautiously. 
\cite{Dekke-2019} raises a concern that lack of local traffic culture, such as when to honk rather than light signals or yield modestly, may cause people's distrust, and increase the risk of conflicts between human drivers and AVs. Unfortunately, most current AV training processes mainly focus on learning a general and culture-blind AV policy instead of learning how to drive like a local driver. 
\cite{zhao2020field} conducts a sequence of field experiments with ten recruited drivers for HV-AV scenarios. 
Using headways, gaps, and speed deviation, 
the participants are grouped into three types: AV-believers, AV-skeptics, and AV- insensitives. 
In other words, one's car-following reaction to AVs highly depends on her trusts on AV technologies.

 	 \item \textbf{AV-HV:} 	
 	For those uncontrollable AVs, how they interact with the HV-dominated traffic depends on the autonomous driving algorithms programmed into these AVs. 
 	The existing AV data, such as Waymo open data, contains rich information of how one AV interacts with its surrounding environment. 
 	Our work~\citep{gu2020lstm} proposes an LSTM model to understand how an AV follows HVs.
 	
 	For those AVs that are controllable, effective mutual interaction would help AVs to effectively communicate and exchange messages with HVs and other road users. 
 	To design AVs' interaction interface is an emerging field in marriage of human factors and human-machine interaction \citep{vinkhuyzen2016developing,muller2016social,Wolf-2016,zhang2017evaluation}. 
 	
 	\item \textbf{AV-AV:} 
 	For those uncontrollable AVs, with a low market penetration rate, it is less likely for an AV to encounter another AV. Thus, at this point it is difficulty to model how two AVs interact in the HV-dominated traffic environment using data-driven approaches. 
 	
 	For those AVs that are controllable, 
 	AVs should be capable of communicating with one another via V2V/V2I or cloud-based technologies.
 \end{enumerate}

In the modeling aspect, 
there does not exist any formal definition of what constitutes ``interactions" between AVs and HVs at a microscopic level. 
Here we provide an abstract definition of vehicular interactions. 
\begin{defn}
	\textbf{Interactions}: 
	how the presence of other vehicles influences the driving strategies of AVs and vice versa. 
	The vehicular interaction can be modeled through 
	joint states, 
	physical constraints,
	coupled rewards or objective functions. 
	It can be categorized into \emph{local} and \emph{global} interactions, replying on information technologies.
\end{defn}

In a platoon of CAVs, 
vehicles interact either 
locally (e.g., the immediate leader and the follower) 
or globally. 
The local pairwise interaction between the immediate leader and the follower is captured in all CFMs,  
where the speed difference and headway with the immediate leader influences one's acceleration \citep{talebpour2016influence,cui2017stabilizing,wu2018stabilizing}. 
When V2V communication links are introduced in a platoon, 
additional interaction terms that reflect the speed and headway influence of far upstream leading vehicles are accounted for in CFMs by adding accelerations of k-ahead-vehicle \citep{jin2014dynamics,qin2017scalable,jin2018connected,jin2018experimental} 
or interaction links \citep{li2017platoon,li2017dynamical,li2018nonlinear}. 
These models do not consider physical collisions. 
To fix it, physical distance constraints must be imposed. 
Accordingly, \cite{wang2014rolling1,wang2014rolling2,gong2016constrained,gong2018cooperative,zhou2017rolling} encode these hard constraints into optimal control problems.

Researchers from control and robotics communities primarily 
focus on local interaction and 
formalize it using different tools. 
\cite{sadigh2016planning} assumes that AVs actions can influence HVs immediately through carefully selected reward functions. 
\cite{lazar2018maximizing} further illustrates that an ``interaction-aware" AV can maximize road capacity leveraging such interaction. 
Furthermore, \cite{sadighverifying} employs the concept of adversarial game for a game-theoretic interaction: 
the HV serves as the AV's adversarial and play an adversarial game with the AV. 
The goal is to find a sequence of human driving actions that could lead to the AV's unsafe behavior. 
In this game, the AV takes actions to maximize its cumulative reward, 
while the HV tries to falsify the AV's action by selecting driving actions that minimize the AV's reward function. 
The driving actions of the AV solved from this game ensures a robust controller design that accommodates measurement or prediction errors of the HV behavior.

Transportation researchers are more interested in the impact of microscopic AV-HV interactions on macroscopic traffic flow patterns \citep{chen2019traffic} 
and its implication for traffic controls in the presence of AVs \citep{levin2016multiclass}. 
A majority of studies use simulations, due to the complex project from micro to macro scales. 
On the macroscopic level, a multi-class traffic modeling approach is commonly adopted.
Among a large amount of studies on the multiclass LWR for the interaction between multiple types of traffic flows,  
\cite{levin2016multiclass,patel2016effects,kockelman2017assessment,melson2018dynamic} have applied it to AV-HV mixed traffic and proposed networked traffic controls. 
To capture the effect of communication and information sharing on traffic flow, 
\cite{ngoduy2009continuum,ngoduy2013instability,ngoduy2013analytical} propose a multiclass non-equilibrium gas-kinetic theory based model to characterize traffic flow dynamics for connected and automated vehicles and analyzed stability of the developed controllers. 
These models, however, may lack detailed interpretations of how two types of vehicles interact on a microscopic level. 
We urgently need a micro-macro analytical framework to offer insights into how microscopic interactions are designed for desirable traffic flow patterns. 

We summarize the above vehicular interaction types in Table~(\ref{tab:interaction}). 
\begin{table}[H]\centering
	\centering\caption{Vehicle interaction types}
	\label{tab:interaction}\vspace{-0.1in}
	\begin{tabular}{|p{0.2 cm}|p{0.2 cm}|p{3.2 cm}||p{8 cm}|p{2.2 cm}|}
		\hline
		\multicolumn{3}{|c||}{Interaction Type} & Reference & Community \\ \hline
		\multirow{6}{*}{\parbox[t]{1mm}{\rotatebox[origin=c]{90}{Micro}}} & \multirow{3}{*}{\parbox[t]{1mm}{\rotatebox[origin=c]{90}{Local}}} & pairwise car-following & \cite{talebpour2016influence,cui2017stabilizing,wu2018stabilizing} & Transportation \\ \cline{3-5}
		& & influence by design &  \cite{sadigh2016planning,lazar2018maximizing} & Robotics \\ \cline{3-5}
		& & adversarial game &  \cite{sadighverifying} & Robotics \\ \cline{2-5}
		& \multirow{3}{*}{\parbox[t]{1mm}{\rotatebox[origin=c]{90}{Global}}} & k-ahead-vehicle term & \cite{jin2014dynamics,qin2017scalable,jin2018connected,jin2018experimental} & Control \\ \cline{3-5}
		& & interaction link &  \cite{li2014survey,li2017platoon,li2017dynamical,li2018nonlinear} & Transportation \& control \\ \cline{3-5}
		& & hard constraints &  \cite{wang2014rolling1,wang2014rolling2,gong2016constrained,gong2018cooperative,zhou2017rolling} & Transportation \\  \hline\hline
        \multirow{2}{*}{\parbox[t]{1mm}{\rotatebox[origin=c]{90}{Macro}}} & \parbox[t]{1mm}{\rotatebox[origin=c]{90}{Equ.}} & multiclass & \cite{levin2016multiclass,patel2016effects,kockelman2017assessment,melson2018dynamic} & Transportation \\ \cline{2-5}
        & {\parbox[t]{1mm}{\rotatebox[origin=c]{90}{Non-equ.}}} & gas-kinetic & \cite{ngoduy2009continuum,ngoduy2013instability,ngoduy2013analytical} & Transportation \\ \hline
	\end{tabular}
\end{table}

\section{Model Summary}\label{sec:sum}

In this section, we summarize all the mixed traffic models based on physics-based and AI-based categories. 

\begin{landscape}
	\setlength\LTcapwidth{\textwidth} 
	\setlength\LTleft{-40pt}            
	\setlength\LTright{-30pt}           
	\small\begin{longtable}{p{1.3 cm}||p{1 cm}|p{1.5 cm}||p{2 cm}|p{2 cm}||p{3 cm}|p{2.5 cm}||p{1 cm}|p{1.5 cm}||p{2 cm}|p{4 cm}}	
		\caption{Physics-based mixed traffic model summary}
		\label{tab:phy} \\ \hline
		Interaction scenario & Coop. or comp. & Model & AV controller & Goal & HV driving model & HV data \& estimation & Traffic scenario & Simulation & Algorithm & Reference \\ \hline\hline
		\multirow{2}{*}{\parbox{1cm}{n AVs}} 
		& Coop. & Linear controller & & string stability & CACC & - & CF & numerical, field experiments & - &  \cite{schakel2010effects,naus2010string,ploeg2011design,milanes2014cooperative,milanes2014modeling,cui2017stabilizing} \\ \cline{2-11} 
		& Coop. & linearly constrained linear quadratic Gaussian, optimal control & General longitudinal vehicle dynamics, (serial distributed) MPC & robust local and string stability (with uncertainties) & - & - & CF & - & Sequential distributed algorithm & \cite{zhou-2017,zhou2019robust,zhou2019distributed} \\ \hline
		\multirow{2}{*}{\parbox{1cm}{n AVs + 1 HV}} & Coop. & CCC & $n$-car-ahead OVM & optimal velocity, close to uniform flow & spacing and speed feedback & sweeping least square  & CF & numerical, field experiments & recursive method for LQ with distributed delay &  \cite{qin2013digital,jin2014dynamics,qin2017scalable,jin2018connected,jin2018experimental} \\ \cline{2-11} 
		& Coop. & Optimal control & fully observable one- or p-step MPC & transient traffic smoothness, asymptotic stability & Newell & Online curve matching NGSIM data & CF & - & Dual-based distributed algorithms &  \cite{gong2016constrained,gong2018cooperative} \\ \hline 
		\multirow{4}{*}{\parbox{1cm}{n AVs + m HV}} & Coop. & Linear quadratic regulator optimization & closed form & Equilibrium spacing, speed difference, acceleration rates & (Chained) asymmetric behavior model & - & CF & A 10-vehicle platoon & - & \cite{chen2019traffic} \\ \cline{2-11}
		& Coop. & \parbox{1.5cm}{distributed frequency-domain-based} & hierarchical control  & String stability for mixed platoons & Linearized general form & CF & NGSIM & Two mixed vehicular platoons led by two HVs sampled from NGSIM in MATLAB & H-infinity control & \cite{zhou2020stabilizing} \\ \hline
 \end{longtable}
 \small{(Abbreviation: coop. -- cooperative, comp. -- competitive. CF -- Car-following; LC -- Lane-change.)}
\end{landscape}

\begin{landscape}
	\setlength\LTcapwidth{\textwidth} 
	\setlength\LTleft{-40pt}            
	\setlength\LTright{-30pt}           
	\small\begin{longtable}{p{1 cm}||p{1 cm}|p{2 cm}||p{2 cm}|p{4 cm}||p{2 cm}|p{1.5 cm}||p{1.5 cm}|p{1.5 cm}||p{2 cm}|p{3 cm}}	
		\caption{AI-based mixed traffic model summary}
		\label{tab:AI} \\ \hline
		\parbox{0.8cm}{Interaction scenario} & \parbox{1cm}{Coop. Comp.} & Model & AV controller & Reward & HV driving model & HV data \& estimation & Traffic scenario & Simulation & Control solution algorithm & Reference \\ \hline\hline 
		\multirow{6}{*}{\parbox{1cm}{1 AV + 1 HV}} & Comp. & two--person non-zero sum game & game & spacing, safety & game & NGSIM & LC, unprotected left-turn & numerical & simulated moments, MLE & \cite{Talebpour2015,yu2018human,zhang2019game,yoo2020game} \\ \cline{2-11}
& Comp. & Two-person game & mixed-motive or Stackelberg game & spacing, safety, LC feasibility & PD controllers  & NHTSA 100-care naturalistic driving safety dataset & CF, LC, merge & multi-lane highway in MATLAB Simulink and dSPACE & Bilevel evolutionary algorithm (BLEAQ) & \cite{yoo2012stackelberg,yoo2013stackelberg,kim2014game,yu2018human,coskun2019receding} \\ \cline{2-11}
		& Coop. & Stackelberg or hierarchical game & MDP & efficiency and safety, the AV's influence on HVs, lane-keeping & Continuous inverse optimal control & Simulated driving trajectories & LC, overtaking, merge & two-lane highway & Feedback Stackelberg dynamic program & \cite{sadigh2016planning,sadighverifying,fisac2019hierarchical} 
		\\ \hline
		\multirow{6}{*}{\parbox{1cm}{1 AV + m HVs}} & Comp. & Reactive game & Stackelberg, decision tree policies & collision, on-road, distance-to-object, safe separation, lane position, speed, effort & DAgger BO (hierarchical reasoning) & NGSIM or simulated & \parbox{2cm}{LC, unsignalized intersection} & multi-lane highways, grid network with roundabouts in TORCS & receding-horizon optimization &  \cite{li2018game,tian2018adaptive,tian2019game} \\ \cline{2-11}
		& - & MDP & End-to-End controller based on BC & Similarity to experts' behavior in data & HV behaviors recorded in data & HV data collected by data collection car driven by a human & Driving on real roads or pre-designed obstacle-filled roads & No simulation involved & CNN, LSTM, and their combinations and variants, such as FCN-LSTM, C-LSTM &  \cite{Pomerleau-1989, Muller-2006, Bojarski-2016,Bojarski-2017,Rausch-2017,Bechtel-2018,Pan-2018,Xu-2017,Erqi-2017,onboard_hecker2018end,Bansal-2018}  \\ \cline{2-11}
		& - & MDP & End-to-end controller based on DRL & cost related to collisions, location on the road, angel between vehicle and road headings, difference from desired speed, et al. & IDM, PD controllers or other pre-programmed HV in the gaming simulator & simulated & CF, LC and racing scenarios & SUMO, TORCS, World Rally Championship 6 (WRC 6) & Deep policy networks, deep Q networks using training methods, such as TRPO, DDPG, advantage Artor-critic, A3C &  \cite{Lill-2015, Tianhao-2016,Sallab-2017,Perot-2017,Jaritz-2018}  \\ \cline{2-11}
		& - & (PO)MDP & MCTS + DRL & collision avoidance, keep on the road, task completion bonus and etc. & IDM, PD controllers or pre-programmed HV in the simulator & simulated & tactical LC, path planning & self-developed simulators, such as POMDPs.jl, SIMULATE function  & MCTS guided by Deep policy network and value estimation & \cite{Paxton-2017,Sunberg-2017,Hoel-2020} \\ \cline{2-11} 
		& - & MDP with probabilistic guarantees & probabilistic specification expressed with linear temporal logic, modified $\epsilon$-greedy exploration policy & make a left turn safely and efficiently & IDM, pedestrians move at a constant speed & - & Unsignalized intersections & - & DQN with prioritized experience replay & \cite{bouton2017belief,Bouton2018uai} \\ \hline
		\multirow{2}{*}{\parbox{2cm}{n AVs}} 
		& Comp., Coop. & differential game & (Multi-anticipative) ACC, rolling horizon control & safety, equilibrium, control, travel efficiency, route, lane preference, lane switch  & IDM, Helly CFM &  &  CF, LC & one-lane highway & Pontryagin’s Minimum Principle & \cite{wang2015game,wang2016cooperative} \\ \hline 
		\multirow{2}{*}{\parbox{1 cm}{n AVs + m HVs}} 
		& Comp. & Multi-population differential game & MFG & efficiency, safety, kinetic energy & ARZ & - & CF & Numerical & Multigrid preconditioned Newton's finite difference & \cite{huang2019stable,huang2020game,huang2020stable} \\ \cline{2-11} 
		& Coop. & Model-free fully observable DRL & GRU NN & close to desirable system-level velocity, collision penalty & IDM & - & CF, LC, merge & FLOW (based on SUMO) & centralized training and execution with TRPO policy gradient in state equivalence class representation, transfer learning & \cite{wu2017emergent,wu2017flow,wu2017framework,wu2018stabilizing,kreidieh2018dissipating,wu2018learning}\\ \hline 
 \end{longtable}
  \small{(Abbreviation: coop. -- cooperative, comp. -- competitive. CF -- Car-following; LC -- Lane-change.)}
\end{landscape}

\section{Conclusions and Open Questions}\label{sec:con}

We will discuss open questions that are unanswered in the existing literature and provide several promising research directions. 

 \subsection{Scalable Multi-AV Controls for Social Optima}


There are few successful applications for MARL in autonomous driving, especially in complex multi-AV driving scenarios. Most previous research focuses on either using centralized but computationally-heavy MARL approaches for cooperative policy to achieve long-term traffic efficiency~\citep{CWu-2018}, or applying decentralized parameter-sharing but non-cooperative techniques for collision-free driving for multiple AVs \citep{Mykel2-2018}.  In addition, MARL is a fast evolving research area, but its application to multi-autonomous driving has lagged behind. Most researchers are still using basic deep RL algorithms such as deep Q network, which is not able to solve some complex problems with more than one AV. As having been discussed above, much more powerful MARL algorithms were developed in recent years but few of them have been applied to multi-AV tasks and traffic domain. In the sense, research is highly in demand on extending existing MARL algorithms or developing brand-new MARL for multi-AV and mixed AV-HV scenarios. 

Nevertheless, AV algorithmic designers tend to program AVs for individual welfare, such as protecting occupants or selecting a fastest route selfishly, with no incentive for improved traffic performance.  
City planners have to regulate the behavior of AVs or their designers for social good. 
Such competing goals pose difficulty in upper level control imposed by planners, which has not been explored in the existing literature. 
A socially optimal control scheme needs to be devised for city planners to guide the autonomous driving technology toward social optima.

\subsection{Human Driving Policy Learning}  

Only when we begin to study AVs, have we learned that as humans, we know little about our own driving behavior. 


 \subsubsection{Physics-Informed AI Models}

Both physics-based and AI-guided methods have limitations: the former highly relies on existing physical traffic models, which may only capture limited dynamics of real-world traffic, resulting in low-quality estimation; 
While the latter requires massive data in order to perform accurate and generalizable estimation. 
Nevertheless, AI-based models may not capture all traffic phenomena observed in the real-world \citep{zhou2019longitudinal}. 
To mitigate the limitations, a physics-informed deep learning (PIDL) framework could potentially predict one's driving behavior with small amounts of observed data more efficiently. 
PIDL contains both model-driven and data-driven components, making possible to leverage the advantages of both scientific models and deep learning techniques while overcoming the shortcomings of either.

The integration of PDE-based physics and deep learning has recently become gradually popular as an effective alternative PDE solver~\citep{Raissi-2018a, Raissi-2018b}. 
Increasing attentions have been paid to the application of PIDL in scientific and engineering areas, to name a few, the discovery of constitutive laws for flow through porous media~\citep{Yang-2019}, the prediction of vortex-induced vibrations~\citep{Maziar-2019}, 3D surface reconstruction~\citep{Fang-2020}, and the inference of hemodynamics in intracranial aneurysm~\citep{Maziar-2020}. 
We have seen a few studies that adopt the similar framework in the transportation community. 
 \cite{Hofleitner-2012} and \cite{Wang-2019} embed a probabilistic models into a dynamic Bayesian network to estimate the evolution of travel time on links. 
\cite{Wu-2018} use neural networks with different structures to capture the general behavior of a car-following model and predict the acceleration of a vehicle using velocity and distance information.   

In conclusion, available human driver data is usually sparse, likely leading to sample bias issues. 
PIDL has shown its predictive robustness with smaller datasets and could become a promising direction in human driving behavior learning. 
Existing traffic models would provide some prior knowledge and help constrain the admissible solutions of AI approaches.  

\subsubsection{Experimental Design}

Little research discusses optimal data or experiment selection for a robust traffic model calibration. 
The interactions of HVs on roads could generate emergent dynamics, i.e., traffic jam in the form of stop-and-go wave or oscillation. 
A driving model calibrated by one dataset may not be capable of predicting the emergent dynamics arising from another dataset. 
In other words, the predictive power of a driving model heavily depends on its training and test datasets. 
Field experiments are, of course, not just costly but highly risky to perform. 
Thus optimal experimental design can help collect representative training and validation datasets. 
What experiments to perform, 
what data to collect, 
and what data to use for behavioral learning and policy training 
are all unresolved questions. 

 \subsubsection{Heterogeneity}

The mixed traffic model need to account for heterogeneity of human drivers (e.g., different capabilities and risk profiles, human driving errors) and AVs (e.g. acceleration and braking capacity, as well as manufacturer's choice of risk tolerance).
In particular, 
humans are highly heterogeneous, due to personal taste and preference, randomness or aggressiveness, and driving experience. 
With the same environment and information, different drivers may maneuver their cars differently. 
A robust model has to be able to accommodate these deviations and predict a distribution of actions that is consistent with real-world observations.

 \subsection{Multi-Scale Human-Machine Ecosystem Modeling}


The overarching goal of researchers is to understand the new traffic pattern comprised of large numbers of AVs and HVs 
and the systematic impact of AVs on traffic safety and efficiency. 
Such an understanding of the macroscopic traffic behavior should be rooted in both the microscopic behavior of AVs \citep{chen2019traffic}   
and evolution of the driving behavior of HVs over time \citep{chen2019lib}. 
This topic can be positioned to a broader context 
which is the 
\emph{collective behavior of hybrid human-machine} \citep{rahwan2019machine}. 
Thus, bridging traffic models on both the micro- and micro-scale using a multi-scale scheme needs to be understood. 

A majority of existing studies on AVs are primarily focused on highways. 
An urban traffic environment consists traffic entities including cars, traffic lights, pedestrians, (motor)cyclists, scooters, and other road users. 
This multimodal mixed traffic environment will further complicate the control of AVs driving alongside various road users.

\subsection{Accountable, Fair, and Ethical AVs}\label{sec:ai}

When automated systems can make life or death decisions for humans, questions arise related to AI-based algorithmic decision-making: 
\begin{enumerate}
	\item Accountability: We need to determine and assign responsibility for damages or injuries caused by AVs.  
	\item Fairness: AVs should make unbiased decisions. 
	\item Ethics: Ethical collection and use of data should be guaranteed for privacy-preserving. AVs should also make ethical decisions.  
\end{enumerate}
There are qualitative studies on above topics, but how to integrate these aspects into engineering decision remains unanswered. 
Interdisciplinary collaboration with legal experts and social scientists is the key to success. 
 
\subsection{A Pathway to Artificial General Intelligence (AGI)}\label{sec:agi}

There is still a long way to reach AGI, which is the ultimate intelligence of machines, 
AVs need to reach human-level AIG, with the capabilities of reasoning, knowledge representation, planning, learning, communicating in natural language, and integration of all these skills towards common goals \citep{hodson2019deepmind}. 
This not only requires bridging gaps with AI tools, 
but also a convergence of engineering, cognitive science, and social science. 
If achieved, it is not only a breakthrough to AV controls, but also to humanity.

In summary, with a rapid growing AV fleet on public roads, 
it is crucial to develop analytical tools for mixed traffic, which will help traffic engineers better understand the impact of AVs on transportation system performances,  
for the AV industry to develop a scalable autonomous driving control algorithm,  
and ultimately, for city planners, policymakers, and lawmakers to manage AVs for social good. 

\section*{Acknowledgments}

The authors would like to thank Data Science Institute from
Columbia University for providing a seed grant for this research. 
This work is also partially supported by the National Science Foundation under award number CMMI-1943998 
and Amazon AWS Machine Learning Research Award Gift (\#US3085926).


\bibliographystyle{elsarticle-harv}{}

\bibliography{survey,survey_MFG,survey_Kuang,survey_Zhaobin,survey_DCL,literatureCV,survey_rongye,survey_Di,survey_mdp}

\begin{thebibliography}{307}
\expandafter\ifx\csname natexlab\endcsname\relax\def\natexlab#1{#1}\fi
\providecommand{\url}[1]{\texttt{#1}}
\providecommand{\href}[2]{#2}
\providecommand{\path}[1]{#1}
\providecommand{\DOIprefix}{doi:}
\providecommand{\ArXivprefix}{arXiv:}
\providecommand{\URLprefix}{URL: }
\providecommand{\Pubmedprefix}{pmid:}
\providecommand{\doi}[1]{\href{http://dx.doi.org/#1}{\path{#1}}}
\providecommand{\Pubmed}[1]{\href{pmid:#1}{\path{#1}}}
\providecommand{\bibinfo}[2]{#2}
\ifx\xfnm\relax \def\xfnm[#1]{\unskip,\space#1}\fi
\bibitem[{Abbeel and Ng(2004a)}]{abbeel-2004}
\bibinfo{author}{Abbeel, P.}, \bibinfo{author}{Ng, A.Y.},
  \bibinfo{year}{2004}a.
\newblock \bibinfo{title}{Apprenticeship learning via inverse reinforcement
  learning}, in: \bibinfo{booktitle}{Proceedings of the twenty-first
  international conference on Machine learning (ICML)}, p.~\bibinfo{pages}{1}.
\bibitem[{Abbeel and Ng(2004b)}]{abbeel2004apprenticeship}
\bibinfo{author}{Abbeel, P.}, \bibinfo{author}{Ng, A.Y.},
  \bibinfo{year}{2004}b.
\newblock \bibinfo{title}{Apprenticeship learning via inverse reinforcement
  learning}, in: \bibinfo{booktitle}{Proceedings of the twenty-first
  international conference on Machine learning}, \bibinfo{organization}{ACM}.
  p.~\bibinfo{pages}{1}.
\bibitem[{Abbeel and Ng(2011)}]{abbeel2011inverse}
\bibinfo{author}{Abbeel, P.}, \bibinfo{author}{Ng, A.Y.}, \bibinfo{year}{2011}.
\newblock \bibinfo{title}{Inverse reinforcement learning}, in:
  \bibinfo{booktitle}{Encyclopedia of machine learning}.
  \bibinfo{publisher}{Springer}, pp. \bibinfo{pages}{554--558}.
\bibitem[{A{I}{M}{S}{U}{N}()}]{aimsun}
\bibinfo{author}{A{I}{M}{S}{U}{N}}, .
\newblock \bibinfo{title}{Simulate a driverless future}.
\newblock \bibinfo{howpublished}{\url{https://www.aimsun.com/aimsun-auto/}}.
\newblock \bibinfo{note}{[Online; accessed 07.01.2020]}.
\bibitem[{Allstr{\"o}m et~al.(2014a)Allstr{\"o}m, Bayen, Fransson,
  Gundleg{\aa}rd, Patire, Rydergren and
  Sandin}]{bluetooth_allstrom2014calibration}
\bibinfo{author}{Allstr{\"o}m, A.}, \bibinfo{author}{Bayen, A.M.},
  \bibinfo{author}{Fransson, M.}, \bibinfo{author}{Gundleg{\aa}rd, D.},
  \bibinfo{author}{Patire, A.D.}, \bibinfo{author}{Rydergren, C.},
  \bibinfo{author}{Sandin, M.}, \bibinfo{year}{2014}a.
\newblock \bibinfo{title}{Calibration framework based on bluetooth sensors for
  traffic state estimation using a velocity based cell transmission model}.
\newblock \bibinfo{journal}{Transportation Research Procedia}
  \bibinfo{volume}{3}, \bibinfo{pages}{972--981}.
\bibitem[{Allstr{\"o}m et~al.(2014b)Allstr{\"o}m, Bayen, Fransson,
  Gundleg{\aa}rd, Patire, Rydergren and
  Sandin}]{onboard_bluetooth_allstrom2014calibration}
\bibinfo{author}{Allstr{\"o}m, A.}, \bibinfo{author}{Bayen, A.M.},
  \bibinfo{author}{Fransson, M.}, \bibinfo{author}{Gundleg{\aa}rd, D.},
  \bibinfo{author}{Patire, A.D.}, \bibinfo{author}{Rydergren, C.},
  \bibinfo{author}{Sandin, M.}, \bibinfo{year}{2014}b.
\newblock \bibinfo{title}{Calibration framework based on bluetooth sensors for
  traffic state estimation using a velocity based cell transmission model}.
\newblock \bibinfo{journal}{Transportation Research Procedia}
  \bibinfo{volume}{3}, \bibinfo{pages}{972--981}.
\bibitem[{Altan et~al.(2017)Altan, Wu, Barth, Boriboonsomsin and
  Stark}]{altan2017glidepath}
\bibinfo{author}{Altan, O.D.}, \bibinfo{author}{Wu, G.},
  \bibinfo{author}{Barth, M.J.}, \bibinfo{author}{Boriboonsomsin, K.},
  \bibinfo{author}{Stark, J.A.}, \bibinfo{year}{2017}.
\newblock \bibinfo{title}{Glidepath: Eco-friendly automated approach and
  departure at signalized intersections}.
\newblock \bibinfo{journal}{IEEE Transactions on Intelligent Vehicles}
  \bibinfo{volume}{2}, \bibinfo{pages}{266--277}.
\bibitem[{Amini et~al.(2020)Amini, Gilitschenski, Phillips, Moseyko, Banerjee,
  Karaman and Rus}]{amini2020learning}
\bibinfo{author}{Amini, A.}, \bibinfo{author}{Gilitschenski, I.},
  \bibinfo{author}{Phillips, J.}, \bibinfo{author}{Moseyko, J.},
  \bibinfo{author}{Banerjee, R.}, \bibinfo{author}{Karaman, S.},
  \bibinfo{author}{Rus, D.}, \bibinfo{year}{2020}.
\newblock \bibinfo{title}{Learning robust control policies for end-to-end
  autonomous driving from data-driven simulation}.
\newblock \bibinfo{journal}{IEEE Robotics and Automation Letters}
  \bibinfo{volume}{5}, \bibinfo{pages}{1143--1150}.
\bibitem[{Aoude et~al.(2012)Aoude, Desaraju, Stephens and
  How}]{aoude2012driver}
\bibinfo{author}{Aoude, G.S.}, \bibinfo{author}{Desaraju, V.R.},
  \bibinfo{author}{Stephens, L.H.}, \bibinfo{author}{How, J.P.},
  \bibinfo{year}{2012}.
\newblock \bibinfo{title}{Driver behavior classification at intersections and
  validation on large naturalistic data set}.
\newblock \bibinfo{journal}{IEEE Transactions on Intelligent Transportation
  Systems} \bibinfo{volume}{13}, \bibinfo{pages}{724--736}.
\bibitem[{Arefizadeh and Talebpour(2018)}]{arefizadeh2018platooning}
\bibinfo{author}{Arefizadeh, S.}, \bibinfo{author}{Talebpour, A.},
  \bibinfo{year}{2018}.
\newblock \bibinfo{title}{A platooning strategy for automated vehicles in the
  presence of speed limit fluctuations}.
\newblock \bibinfo{journal}{Transportation Research Record}
  \bibinfo{volume}{2672}, \bibinfo{pages}{154--161}.
\bibitem[{Arjovsky et~al.(2017)Arjovsky, Chintala and Bottou}]{martin-2017}
\bibinfo{author}{Arjovsky, M.}, \bibinfo{author}{Chintala, S.},
  \bibinfo{author}{Bottou, L.}, \bibinfo{year}{2017}.
\newblock \bibinfo{title}{Wasserstein generative adversarial networks}, in:
  \bibinfo{booktitle}{International Conference on Machine Learning (ICML)}, pp.
  \bibinfo{pages}{214--223}.
\bibitem[{Aw and Rascle(2000)}]{aw2000resurrection}
\bibinfo{author}{Aw, A.}, \bibinfo{author}{Rascle, M.}, \bibinfo{year}{2000}.
\newblock \bibinfo{title}{Resurrection of" second order" models of traffic
  flow}.
\newblock \bibinfo{journal}{SIAM journal on applied mathematics}
  \bibinfo{volume}{60}, \bibinfo{pages}{916--938}.
\bibitem[{Bansal et~al.(2019)Bansal, Krizhevsky and Ogale}]{Bansal-2018}
\bibinfo{author}{Bansal, M.}, \bibinfo{author}{Krizhevsky, A.},
  \bibinfo{author}{Ogale, A.}, \bibinfo{year}{2019}.
\newblock \bibinfo{title}{Chauffeurnet: Learning to drive by imitating the best
  and synthesizing the worst}, in: \bibinfo{booktitle}{Robotics: Science and
  Systems}.
\bibitem[{Bansal et~al.(2017)Bansal, Calandra, Xiao, Levine and
  Tomlin}]{bansal2017goal}
\bibinfo{author}{Bansal, S.}, \bibinfo{author}{Calandra, R.},
  \bibinfo{author}{Xiao, T.}, \bibinfo{author}{Levine, S.},
  \bibinfo{author}{Tomlin, C.J.}, \bibinfo{year}{2017}.
\newblock \bibinfo{title}{Goal-driven dynamics learning via bayesian
  optimization}.
\newblock \bibinfo{journal}{arXiv preprint arXiv:1703.09260} .
\bibitem[{Barooah et~al.(2009)Barooah, Mehta and
  Hespanha}]{barooah2009mistuning}
\bibinfo{author}{Barooah, P.}, \bibinfo{author}{Mehta, P.G.},
  \bibinfo{author}{Hespanha, J.P.}, \bibinfo{year}{2009}.
\newblock \bibinfo{title}{Mistuning-based control design to improve closed-loop
  stability margin of vehicular platoons}.
\newblock \bibinfo{journal}{IEEE Transactions on Automatic Control}
  \bibinfo{volume}{54}, \bibinfo{pages}{2100--2113}.
\bibitem[{Becchio et~al.(2012)Becchio, Manera, Sartori, Cavallo and
  Castiello}]{becchio2012grasping}
\bibinfo{author}{Becchio, C.}, \bibinfo{author}{Manera, V.},
  \bibinfo{author}{Sartori, L.}, \bibinfo{author}{Cavallo, A.},
  \bibinfo{author}{Castiello, U.}, \bibinfo{year}{2012}.
\newblock \bibinfo{title}{Grasping intentions: from thought experiments to
  empirical evidence}.
\newblock \bibinfo{journal}{Frontiers in human neuroscience}
  \bibinfo{volume}{6}, \bibinfo{pages}{117}.
\bibitem[{Bechtel et~al.(2018)Bechtel, McEllhiney, Kim and Yun}]{Bechtel-2018}
\bibinfo{author}{Bechtel, M.G.}, \bibinfo{author}{McEllhiney, E.},
  \bibinfo{author}{Kim, M.}, \bibinfo{author}{Yun, H.}, \bibinfo{year}{2018}.
\newblock \bibinfo{title}{Deep{P}icar: A low-cost deep neural network-based
  autonomous car}, in: \bibinfo{booktitle}{IEEE 24th International Conference
  on Embedded and Real-Time Computing Systems and Applications (RTCSA)}, pp.
  \bibinfo{pages}{11--21}.
\bibitem[{Bevly et~al.(2016)Bevly, Cao, Gordon, Ozbilgin, Kari, Nelson,
  Woodruff, Barth, Murray, Kurt et~al.}]{bevly2016lane}
\bibinfo{author}{Bevly, D.}, \bibinfo{author}{Cao, X.},
  \bibinfo{author}{Gordon, M.}, \bibinfo{author}{Ozbilgin, G.},
  \bibinfo{author}{Kari, D.}, \bibinfo{author}{Nelson, B.},
  \bibinfo{author}{Woodruff, J.}, \bibinfo{author}{Barth, M.},
  \bibinfo{author}{Murray, C.}, \bibinfo{author}{Kurt, A.}, et~al.,
  \bibinfo{year}{2016}.
\newblock \bibinfo{title}{Lane change and merge maneuvers for connected and
  automated vehicles: A survey}.
\newblock \bibinfo{journal}{IEEE Transactions on Intelligent Vehicles}
  \bibinfo{volume}{1}, \bibinfo{pages}{105--120}.
\bibitem[{Bhattacharyya et~al.(2018a)Bhattacharyya, Phillips, Wulfe, Morton,
  Kuefler and Kochenderfer}]{Bhattacharyya2018}
\bibinfo{author}{Bhattacharyya, R.}, \bibinfo{author}{Phillips, D.P.},
  \bibinfo{author}{Wulfe, B.}, \bibinfo{author}{Morton, J.},
  \bibinfo{author}{Kuefler, A.}, \bibinfo{author}{Kochenderfer, M.J.},
  \bibinfo{year}{2018}a.
\newblock \bibinfo{title}{Multi-agent imitation learning for driving
  simulation}, in: \bibinfo{booktitle}{IROS}.
\newblock \URLprefix \url{https://arxiv.org/abs/1803.01044}.
\bibitem[{Bhattacharyya et~al.(2019)Bhattacharyya, Phillips, Liu, Gupta,
  Driggs-Campbell and Kochenderfer}]{bhattacharyya2019simulating}
\bibinfo{author}{Bhattacharyya, R.P.}, \bibinfo{author}{Phillips, D.J.},
  \bibinfo{author}{Liu, C.}, \bibinfo{author}{Gupta, J.K.},
  \bibinfo{author}{Driggs-Campbell, K.}, \bibinfo{author}{Kochenderfer, M.J.},
  \bibinfo{year}{2019}.
\newblock \bibinfo{title}{Simulating emergent properties of human driving
  behavior using multi-agent reward augmented imitation learning}, in:
  \bibinfo{booktitle}{2019 International Conference on Robotics and Automation
  (ICRA)}, \bibinfo{organization}{IEEE}. pp. \bibinfo{pages}{789--795}.
\bibitem[{Bhattacharyya et~al.(2018b)Bhattacharyya, Phillips, Wulfe, Morton,
  Kuefler and Kochenderfer}]{Mykel2-2018}
\bibinfo{author}{Bhattacharyya, R.P.}, \bibinfo{author}{Phillips, D.J.},
  \bibinfo{author}{Wulfe, B.}, \bibinfo{author}{Morton, J.},
  \bibinfo{author}{Kuefler, A.}, \bibinfo{author}{Kochenderfer, M.J.},
  \bibinfo{year}{2018}b.
\newblock \bibinfo{title}{Multi-agent imitation learning for driving
  simulation}, in: \bibinfo{booktitle}{2018 IEEE/RSJ International Conference
  on Intelligent Robots and Systems (IROS)}, \bibinfo{organization}{IEEE}. pp.
  \bibinfo{pages}{1534--1539}.
\bibitem[{B{\i}y{\i}k and Sadigh(2018)}]{biyik2018batch}
\bibinfo{author}{B{\i}y{\i}k, E.}, \bibinfo{author}{Sadigh, D.},
  \bibinfo{year}{2018}.
\newblock \bibinfo{title}{Batch active preference-based learning of reward
  functions}.
\newblock \bibinfo{journal}{arXiv preprint arXiv:1810.04303} .
\bibitem[{Bogue(2008)}]{bogue2008swarm}
\bibinfo{author}{Bogue, R.}, \bibinfo{year}{2008}.
\newblock \bibinfo{title}{Swarm intelligence and robotics}.
\newblock \bibinfo{journal}{Industrial Robot: An International Journal}
  \bibinfo{volume}{35}, \bibinfo{pages}{488--495}.
\bibitem[{Bojarski et~al.(2016)Bojarski, Del~Testa, Dworakowski, Firner, Flepp,
  Goyal, Jackel, Monfort, Muller, Zhang, Zhang and Zhao}]{Bojarski-2016}
\bibinfo{author}{Bojarski, M.}, \bibinfo{author}{Del~Testa, D.},
  \bibinfo{author}{Dworakowski, D.}, \bibinfo{author}{Firner, B.},
  \bibinfo{author}{Flepp, B.}, \bibinfo{author}{Goyal, P.},
  \bibinfo{author}{Jackel, L.D.}, \bibinfo{author}{Monfort, M.},
  \bibinfo{author}{Muller, U.}, \bibinfo{author}{Zhang, J.},
  \bibinfo{author}{Zhang, X.}, \bibinfo{author}{Zhao, J.},
  \bibinfo{year}{2016}.
\newblock \bibinfo{title}{End to end learning for self-driving cars}.
\newblock \bibinfo{journal}{arXiv preprint arXiv:1604.07316} .
\bibitem[{Bojarski et~al.(2017)Bojarski, Yeres, Choromanska, Choromanski,
  Firner, Jackel and Muller}]{Bojarski-2017}
\bibinfo{author}{Bojarski, M.}, \bibinfo{author}{Yeres, P.},
  \bibinfo{author}{Choromanska, A.}, \bibinfo{author}{Choromanski, K.},
  \bibinfo{author}{Firner, B.}, \bibinfo{author}{Jackel, L.},
  \bibinfo{author}{Muller, U.}, \bibinfo{year}{2017}.
\newblock \bibinfo{title}{Explaining how a deep neural network trained with
  end-to-end learning steers a car}.
\newblock \bibinfo{journal}{arXiv preprint arXiv:1704.07911} .
\bibitem[{Bouton et~al.(2017)Bouton, Cosgun and
  Kochenderfer}]{bouton2017belief}
\bibinfo{author}{Bouton, M.}, \bibinfo{author}{Cosgun, A.},
  \bibinfo{author}{Kochenderfer, M.J.}, \bibinfo{year}{2017}.
\newblock \bibinfo{title}{Belief state planning for autonomously navigating
  urban intersections}, in: \bibinfo{booktitle}{Intelligent Vehicles Symposium
  (IV), 2017 IEEE}, \bibinfo{organization}{IEEE}. pp.
  \bibinfo{pages}{825--830}.
\bibitem[{Bouton et~al.(2018)Bouton, Karlsson, Nakhaei, Fujimura, Kochenderfer
  and Tumova}]{Bouton2018uai}
\bibinfo{author}{Bouton, M.}, \bibinfo{author}{Karlsson, J.},
  \bibinfo{author}{Nakhaei, A.}, \bibinfo{author}{Fujimura, K.},
  \bibinfo{author}{Kochenderfer, M.J.}, \bibinfo{author}{Tumova, J.},
  \bibinfo{year}{2018}.
\newblock \bibinfo{title}{Reinforcement learning with probabilistic guarantees
  for autonomous driving}, in: \bibinfo{booktitle}{Workshop on Safety Risk and
  Uncertainty in Reinforcement Learning,}.
\newblock \URLprefix
  \url{https://drive.google.com/open?id=1d2tl4f6GQgH1SERveTmPAR42bMVwHZAZ}.
\bibitem[{Brown and Sandholm(2018)}]{brown_superhuman_2018}
\bibinfo{author}{Brown, N.}, \bibinfo{author}{Sandholm, T.},
  \bibinfo{year}{2018}.
\newblock \bibinfo{title}{Superhuman {AI} for heads-up no-limit poker:
  {Libratus} beats top professionals}.
\newblock \bibinfo{journal}{Science} \bibinfo{volume}{359},
  \bibinfo{pages}{418--424}.
\bibitem[{Brown and Sandholm(2019)}]{brown_superhuman_2019}
\bibinfo{author}{Brown, N.}, \bibinfo{author}{Sandholm, T.},
  \bibinfo{year}{2019}.
\newblock \bibinfo{title}{Superhuman {AI} for multiplayer poker}.
\newblock \bibinfo{journal}{Science} \bibinfo{volume}{365},
  \bibinfo{pages}{885--890}.
\bibitem[{Browne et~al.(2012)Browne, Powley, Whitehouse, Lucas, Cowling,
  Rohlfshagen, Tavener, Perez, Samothrakis and Colton}]{Browne-2012}
\bibinfo{author}{Browne, C.B.}, \bibinfo{author}{Powley, E.},
  \bibinfo{author}{Whitehouse, D.}, \bibinfo{author}{Lucas, S.M.},
  \bibinfo{author}{Cowling, P.I.}, \bibinfo{author}{Rohlfshagen, P.},
  \bibinfo{author}{Tavener, S.}, \bibinfo{author}{Perez, D.},
  \bibinfo{author}{Samothrakis, S.}, \bibinfo{author}{Colton, S.},
  \bibinfo{year}{2012}.
\newblock \bibinfo{title}{A survey of monte carlo tree search methods}.
\newblock \bibinfo{journal}{IEEE Transactions on Computational Intelligence and
  AI in games} \bibinfo{volume}{4}, \bibinfo{pages}{1--43}.
\bibitem[{Caesar et~al.(2020)Caesar, Bankiti, Lang, Vora, Liong, Xu, Krishnan,
  Pan, Baldan and Beijbom}]{Caesar-2020}
\bibinfo{author}{Caesar, H.}, \bibinfo{author}{Bankiti, V.},
  \bibinfo{author}{Lang, A.H.}, \bibinfo{author}{Vora, S.},
  \bibinfo{author}{Liong, V.E.}, \bibinfo{author}{Xu, Q.},
  \bibinfo{author}{Krishnan, A.}, \bibinfo{author}{Pan, Y.},
  \bibinfo{author}{Baldan, G.}, \bibinfo{author}{Beijbom, O.},
  \bibinfo{year}{2020}.
\newblock \bibinfo{title}{{nuScenes}: A multimodal dataset for autonomous
  driving}, in: \bibinfo{booktitle}{the IEEE Conference on Computer Vision and
  Pattern Recognition (CVPR)}, pp. \bibinfo{pages}{11621--11631}.
\bibitem[{Chang et~al.(2019)Chang, Lambert, Sangkloy, Singh, Bak, Hartnett,
  Wang, Carr, Lucey, Ramanan and Hays}]{Chang-Ming-Fang-2019}
\bibinfo{author}{Chang, M.F.}, \bibinfo{author}{Lambert, J.},
  \bibinfo{author}{Sangkloy, P.}, \bibinfo{author}{Singh, J.},
  \bibinfo{author}{Bak, S.}, \bibinfo{author}{Hartnett, A.},
  \bibinfo{author}{Wang, D.}, \bibinfo{author}{Carr, P.},
  \bibinfo{author}{Lucey, S.}, \bibinfo{author}{Ramanan, D.},
  \bibinfo{author}{Hays, J.}, \bibinfo{year}{2019}.
\newblock \bibinfo{title}{Argoverse: 3{D} tracking and forecasting with rich
  maps}, in: \bibinfo{booktitle}{Proceedings of the IEEE Conference on Computer
  Vision and Pattern Recognition (CVPR)}, pp. \bibinfo{pages}{8748--8757}.
\bibitem[{Chatterjee(2016)}]{chatterjee2016understanding}
\bibinfo{author}{Chatterjee, I.}, \bibinfo{year}{2016}.
\newblock \bibinfo{title}{Understanding Driver Contributions to Rear-End
  Crashes on Congested Freeways and their Implications for Future Safety
  Measures}.
\newblock Ph.D. thesis. University of Minnesota.
\bibitem[{Chatterjee and Davis(2013)}]{chatterjee2013evolutionary}
\bibinfo{author}{Chatterjee, I.}, \bibinfo{author}{Davis, G.},
  \bibinfo{year}{2013}.
\newblock \bibinfo{title}{Evolutionary game theoretic approach to rear-end
  events on congested freeway}.
\newblock \bibinfo{journal}{Transportation Research Record: Journal of the
  Transportation Research Board} , \bibinfo{pages}{121--127}.
\bibitem[{Chen et~al.(2015)Chen, Seff, Kornhauser and Xiao}]{chenyi-2015}
\bibinfo{author}{Chen, C.}, \bibinfo{author}{Seff, A.},
  \bibinfo{author}{Kornhauser, A.}, \bibinfo{author}{Xiao, J.},
  \bibinfo{year}{2015}.
\newblock \bibinfo{title}{Deepdriving: Learning affordance for direct
  perception in autonomous driving}, in: \bibinfo{booktitle}{IEEE International
  Conference on Computer Vision}, pp. \bibinfo{pages}{2722--2730}.
\bibitem[{Chen et~al.(2019a)Chen, Srivastava, Ahn and Li}]{chen2019traffic}
\bibinfo{author}{Chen, D.}, \bibinfo{author}{Srivastava, A.},
  \bibinfo{author}{Ahn, S.}, \bibinfo{author}{Li, T.}, \bibinfo{year}{2019}a.
\newblock \bibinfo{title}{Traffic dynamics under speed disturbance in mixed
  traffic with automated and non-automated vehicles}.
\newblock \bibinfo{journal}{Transportation Research Part C: Emerging
  Technologies} .
\bibitem[{Chen et~al.(2019b)Chen, Yuan and Tomizuka}]{Jianyu-2019}
\bibinfo{author}{Chen, J.}, \bibinfo{author}{Yuan, B.},
  \bibinfo{author}{Tomizuka, M.}, \bibinfo{year}{2019}b.
\newblock \bibinfo{title}{Deep imitation learning for autonomous driving in
  generic urban scenarios with enhanced safety}, in: \bibinfo{booktitle}{arXiv
  preprint arXiv:1903.00640}.
\bibitem[{Chen et~al.(2019c)Chen, Yuan and Tomizuka}]{chen2019model}
\bibinfo{author}{Chen, J.}, \bibinfo{author}{Yuan, B.},
  \bibinfo{author}{Tomizuka, M.}, \bibinfo{year}{2019}c.
\newblock \bibinfo{title}{Model-free deep reinforcement learning for urban
  autonomous driving}, in: \bibinfo{booktitle}{2019 IEEE Intelligent
  Transportation Systems Conference (ITSC)}, \bibinfo{organization}{IEEE}. pp.
  \bibinfo{pages}{2765--2771}.
\bibitem[{Chen and Wang(2005)}]{chen2005formation}
\bibinfo{author}{Chen, Y.Q.}, \bibinfo{author}{Wang, Z.}, \bibinfo{year}{2005}.
\newblock \bibinfo{title}{Formation control: a review and a new consideration},
  in: \bibinfo{booktitle}{2005 IEEE/RSJ International conference on intelligent
  robots and systems}, \bibinfo{organization}{IEEE}. pp.
  \bibinfo{pages}{3181--3186}.
\bibitem[{Chen et~al.(2017)Chen, He, Yin and Du}]{chen2017optimal}
\bibinfo{author}{Chen, Z.}, \bibinfo{author}{He, F.}, \bibinfo{author}{Yin,
  Y.}, \bibinfo{author}{Du, Y.}, \bibinfo{year}{2017}.
\newblock \bibinfo{title}{Optimal design of autonomous vehicle zones in
  transportation networks}.
\newblock \bibinfo{journal}{Transportation Research Part B: Methodological}
  \bibinfo{volume}{99}, \bibinfo{pages}{44--61}.
\bibitem[{Chen et~al.(2016)Chen, He, Zhang and Yin}]{chen2016optimal}
\bibinfo{author}{Chen, Z.}, \bibinfo{author}{He, F.}, \bibinfo{author}{Zhang,
  L.}, \bibinfo{author}{Yin, Y.}, \bibinfo{year}{2016}.
\newblock \bibinfo{title}{Optimal deployment of autonomous vehicle lanes with
  endogenous market penetration}.
\newblock \bibinfo{journal}{Transportation Research Part C: Emerging
  Technologies} \bibinfo{volume}{72}, \bibinfo{pages}{143--156}.
\bibitem[{Choi et~al.(2018)Choi, Kim, Hwang, Park, Yoon, An and
  Kweon}]{Choi-2018}
\bibinfo{author}{Choi, Y.}, \bibinfo{author}{Kim, N.}, \bibinfo{author}{Hwang,
  S.}, \bibinfo{author}{Park, K.}, \bibinfo{author}{Yoon, J.S.},
  \bibinfo{author}{An, K.}, \bibinfo{author}{Kweon, I.S.},
  \bibinfo{year}{2018}.
\newblock \bibinfo{title}{{KAIST} multi-spectral day/night data set for
  autonomous and assisted driving}.
\newblock \bibinfo{journal}{IEEE Transactions on Intelligent Transportation
  Systems} \bibinfo{volume}{19}, \bibinfo{pages}{934--948}.
\bibitem[{Codevilla et~al.(2018)Codevilla, Miiller, L{\'o}pez, Koltun and
  Dosovitskiy}]{Codevilla-2018}
\bibinfo{author}{Codevilla, F.}, \bibinfo{author}{Miiller, M.},
  \bibinfo{author}{L{\'o}pez, A.}, \bibinfo{author}{Koltun, V.},
  \bibinfo{author}{Dosovitskiy, A.}, \bibinfo{year}{2018}.
\newblock \bibinfo{title}{End-to-end driving via conditional imitation
  learning}, in: \bibinfo{booktitle}{2018 IEEE International Conference on
  Robotics and Automation (ICRA)}, \bibinfo{organization}{IEEE}. pp.
  \bibinfo{pages}{1--9}.
\bibitem[{Colombaroni and Fusco(2014)}]{traj_colombaroni2014artificial}
\bibinfo{author}{Colombaroni, C.}, \bibinfo{author}{Fusco, G.},
  \bibinfo{year}{2014}.
\newblock \bibinfo{title}{Artificial neural network models for car following:
  experimental analysis and calibration issues}.
\newblock \bibinfo{journal}{Journal of Intelligent Transportation Systems}
  \bibinfo{volume}{18}, \bibinfo{pages}{5--16}.
\bibitem[{Coskun et~al.(2019)Coskun, Zhang and Langari}]{coskun2019receding}
\bibinfo{author}{Coskun, S.}, \bibinfo{author}{Zhang, Q.},
  \bibinfo{author}{Langari, R.}, \bibinfo{year}{2019}.
\newblock \bibinfo{title}{Receding horizon markov game autonomous driving
  strategy}, in: \bibinfo{booktitle}{2019 American Control Conference (ACC)},
  \bibinfo{organization}{IEEE}. pp. \bibinfo{pages}{1367--1374}.
\bibitem[{Creech et~al.(2019)Creech, Tilbury, Yang, Pradhan, Tsui, Robert
  et~al.}]{creech2019pedestrian}
\bibinfo{author}{Creech, C.}, \bibinfo{author}{Tilbury, D.},
  \bibinfo{author}{Yang, J.}, \bibinfo{author}{Pradhan, A.},
  \bibinfo{author}{Tsui, K.}, \bibinfo{author}{Robert, L.}, et~al.,
  \bibinfo{year}{2019}.
\newblock \bibinfo{title}{Pedestrian trust in automated vehicles: Role of
  traffic signal and av driving behavior}.
\newblock \bibinfo{journal}{Frontiers in Robotics and AI, Forthcoming} .
\bibitem[{Cui et~al.(2017)Cui, Seibold, Stern and Work}]{cui2017stabilizing}
\bibinfo{author}{Cui, S.}, \bibinfo{author}{Seibold, B.},
  \bibinfo{author}{Stern, R.}, \bibinfo{author}{Work, D.B.},
  \bibinfo{year}{2017}.
\newblock \bibinfo{title}{Stabilizing traffic flow via a single autonomous
  vehicle: Possibilities and limitations}, in: \bibinfo{booktitle}{Intelligent
  Vehicles Symposium (IV), 2017 IEEE}, \bibinfo{organization}{IEEE}. pp.
  \bibinfo{pages}{1336--1341}.
\bibitem[{Darbha and Rajagopal(1999)}]{darbha1999intelligent}
\bibinfo{author}{Darbha, S.}, \bibinfo{author}{Rajagopal, K.},
  \bibinfo{year}{1999}.
\newblock \bibinfo{title}{Intelligent cruise control systems and traffic flow
  stability}.
\newblock \bibinfo{journal}{Transportation Research Part C: Emerging
  Technologies} \bibinfo{volume}{7}, \bibinfo{pages}{329--352}.
\bibitem[{Darms et~al.(2008)Darms, Rybski and Urmson}]{darms2008classification}
\bibinfo{author}{Darms, M.}, \bibinfo{author}{Rybski, P.},
  \bibinfo{author}{Urmson, C.}, \bibinfo{year}{2008}.
\newblock \bibinfo{title}{Classification and tracking of dynamic objects with
  multiple sensors for autonomous driving in urban environments}, in:
  \bibinfo{booktitle}{2008 IEEE Intelligent Vehicles Symposium},
  \bibinfo{organization}{IEEE}. pp. \bibinfo{pages}{1197--1202}.
\bibitem[{Davis(2017)}]{davis2017bayesian}
\bibinfo{author}{Davis, G.A.}, \bibinfo{year}{2017}.
\newblock \bibinfo{title}{Bayesian Estimation of Drivers’ Gap Selections and
  Reaction Times in Left-Turning Crashes from Event Data Recorder Pre-Crash
  Data}.
\newblock \bibinfo{type}{Technical Report}. SAE Technical Paper.
\bibitem[{Dekker(2019)}]{Dekke-2019}
\bibinfo{author}{Dekker, J.}, \bibinfo{year}{2019}.
\newblock \bibinfo{title}{The problem with autonomous cars that no ones talking
  about}.
\newblock
  \bibinfo{howpublished}{https://www.fastcompany.com/90392131/the-problem-with-autonomous-cars-that-no-ones-talking-about}.
\bibitem[{Delis et~al.(2016)Delis, Nikolos and
  Papageorgiou}]{delis2016simulation}
\bibinfo{author}{Delis, A.I.}, \bibinfo{author}{Nikolos, I.K.},
  \bibinfo{author}{Papageorgiou, M.}, \bibinfo{year}{2016}.
\newblock \bibinfo{title}{Simulation of the penetration rate effects of acc and
  cacc on macroscopic traffic dynamics}, in: \bibinfo{booktitle}{Intelligent
  Transportation Systems (ITSC), 2016 IEEE 19th International Conference on},
  \bibinfo{organization}{IEEE}. pp. \bibinfo{pages}{336--341}.
\bibitem[{Di et~al.(2020)Di, Chen and Talley}]{chen2019lib}
\bibinfo{author}{Di, X.}, \bibinfo{author}{Chen, X.}, \bibinfo{author}{Talley,
  E.}, \bibinfo{year}{2020}.
\newblock \bibinfo{title}{Liability design for autonomous vehicles and
  human-driven vehicles: A hierarchical game-theoretic approach}.
\newblock \bibinfo{journal}{Transportation Research Part C} .
\bibitem[{Djehiche et~al.(2016)Djehiche, Tcheukam and
  Tembine}]{djehiche2016mean}
\bibinfo{author}{Djehiche, B.}, \bibinfo{author}{Tcheukam, A.},
  \bibinfo{author}{Tembine, H.}, \bibinfo{year}{2016}.
\newblock \bibinfo{title}{Mean-field-type games in engineering}.
\newblock \bibinfo{journal}{arXiv preprint arXiv:1605.03281} .
\bibitem[{Dresner and Stone(2007)}]{dresner2007sharing}
\bibinfo{author}{Dresner, K.M.}, \bibinfo{author}{Stone, P.},
  \bibinfo{year}{2007}.
\newblock \bibinfo{title}{Sharing the road: Autonomous vehicles meet human
  drivers.}, in: \bibinfo{booktitle}{IJCAI}, pp. \bibinfo{pages}{1263--1268}.
\bibitem[{Dreves and Gerdts(2018)}]{dreves2018generalized}
\bibinfo{author}{Dreves, A.}, \bibinfo{author}{Gerdts, M.},
  \bibinfo{year}{2018}.
\newblock \bibinfo{title}{A generalized {N}ash equilibrium approach for optimal
  control problems of autonomous cars}.
\newblock \bibinfo{journal}{Optimal Control Applications and Methods}
  \bibinfo{volume}{39}, \bibinfo{pages}{326--342}.
\bibitem[{Driggs-Campbell and Bajcsy(2016)}]{driggs2016communicating}
\bibinfo{author}{Driggs-Campbell, K.}, \bibinfo{author}{Bajcsy, R.},
  \bibinfo{year}{2016}.
\newblock \bibinfo{title}{Communicating intent on the road through
  human-inspired control schemes}, in: \bibinfo{booktitle}{2016 IEEE/RSJ
  International Conference on Intelligent Robots and Systems (IROS)},
  \bibinfo{organization}{IEEE}. pp. \bibinfo{pages}{3042--3047}.
\bibitem[{Driggs-Campbell et~al.(2018)Driggs-Campbell, Dong and
  Bajcsy}]{driggs2018robust}
\bibinfo{author}{Driggs-Campbell, K.}, \bibinfo{author}{Dong, R.},
  \bibinfo{author}{Bajcsy, R.}, \bibinfo{year}{2018}.
\newblock \bibinfo{title}{Robust, informative human-in-the-loop predictions via
  empirical reachable sets}.
\newblock \bibinfo{journal}{IEEE Transactions on Intelligent Vehicles} .
\bibitem[{Driggs-Campbell et~al.(2017)Driggs-Campbell, Govindarajan and
  Bajcsy}]{driggs2017integrating}
\bibinfo{author}{Driggs-Campbell, K.}, \bibinfo{author}{Govindarajan, V.},
  \bibinfo{author}{Bajcsy, R.}, \bibinfo{year}{2017}.
\newblock \bibinfo{title}{Integrating intuitive driver models in autonomous
  planning for interactive maneuvers}.
\newblock \bibinfo{journal}{IEEE Transactions on Intelligent Transportation
  Systems} \bibinfo{volume}{18}, \bibinfo{pages}{3461--3472}.
\bibitem[{Eraqi et~al.(2017)Eraqi, Moustafa and Honer.}]{Erqi-2017}
\bibinfo{author}{Eraqi, H.M.}, \bibinfo{author}{Moustafa, M.N.},
  \bibinfo{author}{Honer., J.}, \bibinfo{year}{2017}.
\newblock \bibinfo{title}{End‐to‐end deep learning for steering autonomous
  vehicles considering temporal dependencies}, in: \bibinfo{booktitle}{Machine
  Learning for Intelligent Transportation Systems Workshop in the 31st
  Conference on Neural Information Processing Systems}.
\bibitem[{Fang and Zhan(2020)}]{Fang-2020}
\bibinfo{author}{Fang, Z.}, \bibinfo{author}{Zhan, J.}, \bibinfo{year}{2020}.
\newblock \bibinfo{title}{Physics-informed neural network framework for partial
  differential equations on 3{D} surfaces: Time independent problems}.
\newblock \bibinfo{journal}{IEEE Access} .
\bibitem[{Fernandez~Fisac(2019)}]{fernandez2019game}
\bibinfo{author}{Fernandez~Fisac, J.}, \bibinfo{year}{2019}.
\newblock \bibinfo{title}{Game-Theoretic Safety Assurance for Human-Centered
  Robotic Systems}.
\newblock Ph.D. thesis. UC Berkeley.
\bibitem[{Fisac et~al.(2019)Fisac, Bronstein, Stefansson, Sadigh, Sastry and
  Dragan}]{fisac2019hierarchical}
\bibinfo{author}{Fisac, J.F.}, \bibinfo{author}{Bronstein, E.},
  \bibinfo{author}{Stefansson, E.}, \bibinfo{author}{Sadigh, D.},
  \bibinfo{author}{Sastry, S.S.}, \bibinfo{author}{Dragan, A.D.},
  \bibinfo{year}{2019}.
\newblock \bibinfo{title}{Hierarchical game-theoretic planning for autonomous
  vehicles}, in: \bibinfo{booktitle}{2019 International Conference on Robotics
  and Automation (ICRA)}, \bibinfo{organization}{IEEE}. pp.
  \bibinfo{pages}{9590--9596}.
\bibitem[{Flores et~al.(2018)Flores, Merdrignac, de~Charette, Navas,
  Milan{\'e}s and Nashashibi}]{onboard_lidar_flores2018cooperative}
\bibinfo{author}{Flores, C.}, \bibinfo{author}{Merdrignac, P.},
  \bibinfo{author}{de~Charette, R.}, \bibinfo{author}{Navas, F.},
  \bibinfo{author}{Milan{\'e}s, V.}, \bibinfo{author}{Nashashibi, F.},
  \bibinfo{year}{2018}.
\newblock \bibinfo{title}{A cooperative car-following/emergency braking system
  with prediction-based pedestrian avoidance capabilities}.
\newblock \bibinfo{journal}{IEEE Transactions on Intelligent Transportation
  Systems} , \bibinfo{pages}{1--10}.
\bibitem[{Foerster et~al.(2016)Foerster, Assael, de~Freitas and
  Whiteson}]{foerster_learning_2016}
\bibinfo{author}{Foerster, J.N.}, \bibinfo{author}{Assael, Y.M.},
  \bibinfo{author}{de~Freitas, N.}, \bibinfo{author}{Whiteson, S.},
  \bibinfo{year}{2016}.
\newblock \bibinfo{title}{Learning to {Communicate} with {Deep} {Multi}-agent
  {Reinforcement} {Learning}}, in: \bibinfo{booktitle}{Proceedings of the 30th
  {International} {Conference} on {Neural} {Information} {Processing}
  {Systems}}, \bibinfo{publisher}{Curran Associates Inc.},
  \bibinfo{address}{USA}. pp. \bibinfo{pages}{2145--2153}.
\newblock \bibinfo{note}{Event-place: Barcelona, Spain}.
\bibitem[{F{ä}rber(2016)}]{Farber-2016}
\bibinfo{author}{F{ä}rber, B.}, \bibinfo{year}{2016}.
\newblock \bibinfo{title}{Communication and communication problems between
  autonomous vehicles and human drivers}, in: \bibinfo{booktitle}{Autonomous
  Driving}. \bibinfo{publisher}{Springer}, pp. \bibinfo{pages}{125--144}.
\bibitem[{Geiger et~al.(2012)Geiger, Lenz and Urtasun}]{Geiger-2012}
\bibinfo{author}{Geiger, A.}, \bibinfo{author}{Lenz, P.},
  \bibinfo{author}{Urtasun, R.}, \bibinfo{year}{2012}.
\newblock \bibinfo{title}{Are we ready for autonomous driving? the {KITTI}
  vision benchmark suite}, in: \bibinfo{booktitle}{2012 IEEE Conference on
  Computer Vision and Pattern Recognition (CVPR)}, pp.
  \bibinfo{pages}{3354--3361}.
\bibitem[{Geyer et~al.(2019)Geyer, Kassahun, Mahmudi, Ricou, Durgesh, Chung,
  Hauswald, Pham, Muhlegg, Dorn and Fernandez}]{A2D2}
\bibinfo{author}{Geyer, J.}, \bibinfo{author}{Kassahun, Y.},
  \bibinfo{author}{Mahmudi, M.}, \bibinfo{author}{Ricou, X.},
  \bibinfo{author}{Durgesh, R.}, \bibinfo{author}{Chung, A.},
  \bibinfo{author}{Hauswald, L.}, \bibinfo{author}{Pham, V.},
  \bibinfo{author}{Muhlegg, M.}, \bibinfo{author}{Dorn, S.},
  \bibinfo{author}{Fernandez, T.}, \bibinfo{year}{2019}.
\newblock \bibinfo{title}{{A2D2}: Aev autonomous driving dataset}.
\newblock \bibinfo{howpublished}{\url{https://www.a2d2.audi}}.
\bibitem[{Gindele et~al.(2010)Gindele, Brechtel and
  Dillmann}]{gindele2010probabilistic}
\bibinfo{author}{Gindele, T.}, \bibinfo{author}{Brechtel, S.},
  \bibinfo{author}{Dillmann, R.}, \bibinfo{year}{2010}.
\newblock \bibinfo{title}{A probabilistic model for estimating driver behaviors
  and vehicle trajectories in traffic environments}, in:
  \bibinfo{booktitle}{13th International IEEE Conference on Intelligent
  Transportation Systems}, \bibinfo{organization}{IEEE}. pp.
  \bibinfo{pages}{1625--1631}.
\bibitem[{Gipps(1981)}]{gipps1981behavioural}
\bibinfo{author}{Gipps, P.G.}, \bibinfo{year}{1981}.
\newblock \bibinfo{title}{A behavioural car-following model for computer
  simulation}.
\newblock \bibinfo{journal}{Transportation Research Part B: Methodological}
  \bibinfo{volume}{15}, \bibinfo{pages}{105--111}.
\bibitem[{Gong and Du(2018)}]{gong2018cooperative}
\bibinfo{author}{Gong, S.}, \bibinfo{author}{Du, L.}, \bibinfo{year}{2018}.
\newblock \bibinfo{title}{Cooperative platoon control for a mixed traffic flow
  including human drive vehicles and connected and autonomous vehicles}.
\newblock \bibinfo{journal}{Transportation research part B: methodological}
  \bibinfo{volume}{116}, \bibinfo{pages}{25--61}.
\bibitem[{Gong et~al.(2016)Gong, Shen and Du}]{gong2016constrained}
\bibinfo{author}{Gong, S.}, \bibinfo{author}{Shen, J.}, \bibinfo{author}{Du,
  L.}, \bibinfo{year}{2016}.
\newblock \bibinfo{title}{Constrained optimization and distributed computation
  based car following control of a connected and autonomous vehicle platoon}.
\newblock \bibinfo{journal}{Transportation Research Part B: Methodological}
  \bibinfo{volume}{94}, \bibinfo{pages}{314--334}.
\bibitem[{Gonzalez et~al.(2016)Gonzalez, Dibangoye and Laugier}]{gonzalez-2016}
\bibinfo{author}{Gonzalez, D.S.}, \bibinfo{author}{Dibangoye, J.S.},
  \bibinfo{author}{Laugier, C.}, \bibinfo{year}{2016}.
\newblock \bibinfo{title}{High-speed highway scene prediction based on driver
  models learned from demonstrations}, in: \bibinfo{booktitle}{Proceedings of
  the 19th IEEE International Conference on Intelligent Transportation Systems
  (ITSC)}, pp. \bibinfo{pages}{149--155}.
\bibitem[{Goodfellow et~al.(2014)Goodfellow, Pouget-Abadie, Mirza, Xu,
  Warde-Farley, Ozair, Courville and Bengio}]{goodfellow-2014}
\bibinfo{author}{Goodfellow, I.}, \bibinfo{author}{Pouget-Abadie, J.},
  \bibinfo{author}{Mirza, M.}, \bibinfo{author}{Xu, B.},
  \bibinfo{author}{Warde-Farley, D.}, \bibinfo{author}{Ozair, S.},
  \bibinfo{author}{Courville, A.}, \bibinfo{author}{Bengio, Y.},
  \bibinfo{year}{2014}.
\newblock \bibinfo{title}{Generative adversarial nets}, in:
  \bibinfo{booktitle}{Advances in neural information processing systems}, pp.
  \bibinfo{pages}{2672--2680}.
\bibitem[{Greveling(2018)}]{Diederi-2018}
\bibinfo{author}{Greveling, D.P.}, \bibinfo{year}{2018}.
\newblock \bibinfo{title}{Modelling human driving behaviour using Generative
  Adversarial Networks}.
\newblock \bibinfo{publisher}{MS Thesis, University of Groningen}.
\bibitem[{Grigorescu et~al.(2020)Grigorescu, Trasnea, Cocias and
  Macesanu}]{Grigorescu-2020}
\bibinfo{author}{Grigorescu, S.}, \bibinfo{author}{Trasnea, B.},
  \bibinfo{author}{Cocias, T.}, \bibinfo{author}{Macesanu, G.},
  \bibinfo{year}{2020}.
\newblock \bibinfo{title}{A survey of deep learning techniques for autonomous
  driving}.
\newblock \bibinfo{journal}{Journal of Field Robotics} \bibinfo{volume}{37},
  \bibinfo{pages}{362--386}.
\bibitem[{{G}roup()}]{ptv}
\bibinfo{author}{{G}roup, P.}, .
\newblock \bibinfo{title}{Virtual testing of autonomous vehicles with ptv
  vissim}.
\newblock
  \bibinfo{howpublished}{\url{https://www.ptvgroup.com/en/solutions/products/ptv-vissim/areas-of-application/autonomous-vehicles-and-new-mobility/}}.
\newblock \bibinfo{note}{[Online; accessed 07.01.2020]}.
\bibitem[{Gu et~al.(2020)Gu, Li, Di and Shi}]{gu2020lstm}
\bibinfo{author}{Gu, Z.}, \bibinfo{author}{Li, Z.}, \bibinfo{author}{Di, X.},
  \bibinfo{author}{Shi, R.}, \bibinfo{year}{2020}.
\newblock \bibinfo{title}{An lstm-based autonomous driving model using a waymo
  open dataset}.
\newblock \bibinfo{journal}{Applied Sciences} \bibinfo{volume}{10},
  \bibinfo{pages}{2046}.
\bibitem[{Gu{\'e}ant et~al.(2011)Gu{\'e}ant, Lasry and Lions}]{gueant2011mean}
\bibinfo{author}{Gu{\'e}ant, O.}, \bibinfo{author}{Lasry, J.M.},
  \bibinfo{author}{Lions, P.L.}, \bibinfo{year}{2011}.
\newblock \bibinfo{title}{Mean field games and applications}, in:
  \bibinfo{booktitle}{Paris-Princeton lectures on mathematical finance 2010}.
  \bibinfo{publisher}{Springer}, pp. \bibinfo{pages}{205--266}.
\bibitem[{Gulrajani et~al.(2017)Gulrajani, Ahmed, Arjovsky, Dumoulin and
  Courvill}]{Gulrajani-2017}
\bibinfo{author}{Gulrajani, I.}, \bibinfo{author}{Ahmed, F.},
  \bibinfo{author}{Arjovsky, M.}, \bibinfo{author}{Dumoulin, V.},
  \bibinfo{author}{Courvill, A.C.}, \bibinfo{year}{2017}.
\newblock \bibinfo{title}{Improved training of {W}asserstein {GANs}}, in:
  \bibinfo{booktitle}{Advances in Neural Information Processing Systems
  (NIPS)}, pp. \bibinfo{pages}{5767--5777}.
\bibitem[{Haarnoja et~al.(2018)Haarnoja, Zhou, Abbeel and
  Levine}]{haaronja-2018}
\bibinfo{author}{Haarnoja, T.}, \bibinfo{author}{Zhou, A.},
  \bibinfo{author}{Abbeel, P.}, \bibinfo{author}{Levine, S.},
  \bibinfo{year}{2018}.
\newblock \bibinfo{title}{{Soft Actor-Critic: Off-Policy Maximum Entropy Deep
  Reinforcement Learning with a Stochastic Actor}}, in:
  \bibinfo{booktitle}{International Conference on Machine Learning}, pp.
  \bibinfo{pages}{1856--1865}.
\bibitem[{Hammit et~al.(2018)Hammit, Ghasemzadeh, James, Ahmed and
  Young}]{onboard_hammit2018evaluation}
\bibinfo{author}{Hammit, B.E.}, \bibinfo{author}{Ghasemzadeh, A.},
  \bibinfo{author}{James, R.M.}, \bibinfo{author}{Ahmed, M.M.},
  \bibinfo{author}{Young, R.K.}, \bibinfo{year}{2018}.
\newblock \bibinfo{title}{Evaluation of weather-related freeway car-following
  behavior using the shrp2 naturalistic driving study database}.
\newblock \bibinfo{journal}{Transportation research part F: traffic psychology
  and behaviour} \bibinfo{volume}{59}, \bibinfo{pages}{244--259}.
\bibitem[{Hankey et~al.(2016)Hankey, Perez and
  McClafferty}]{hankey2016description}
\bibinfo{author}{Hankey, J.M.}, \bibinfo{author}{Perez, M.A.},
  \bibinfo{author}{McClafferty, J.A.}, \bibinfo{year}{2016}.
\newblock \bibinfo{title}{Description of the {SHRP} 2 naturalistic database and
  the crash, near-crash, and baseline data sets}.
\newblock \bibinfo{type}{Technical Report}. Virginia Tech Transportation
  Institute.
\bibitem[{Hao et~al.(2018)Hao, Wu, Boriboonsomsin and Barth}]{hao2018eco}
\bibinfo{author}{Hao, P.}, \bibinfo{author}{Wu, G.},
  \bibinfo{author}{Boriboonsomsin, K.}, \bibinfo{author}{Barth, M.J.},
  \bibinfo{year}{2018}.
\newblock \bibinfo{title}{Eco-approach and departure (ead) application for
  actuated signals in real-world traffic}.
\newblock \bibinfo{journal}{IEEE Transactions on Intelligent Transportation
  Systems} \bibinfo{volume}{20}, \bibinfo{pages}{30--40}.
\bibitem[{Hecker et~al.(2018a)Hecker, Dai and Van~Gool}]{onboard_hecker2018end}
\bibinfo{author}{Hecker, S.}, \bibinfo{author}{Dai, D.},
  \bibinfo{author}{Van~Gool, L.}, \bibinfo{year}{2018}a.
\newblock \bibinfo{title}{End-to-end learning of driving models with
  surround-view cameras and route planners}, in:
  \bibinfo{booktitle}{Proceedings of the European Conference on Computer Vision
  (ECCV)}, pp. \bibinfo{pages}{435--453}.
\bibitem[{Hecker et~al.(2018b)Hecker, Dai and
  Van~Gool}]{onboard_hecker2018learning}
\bibinfo{author}{Hecker, S.}, \bibinfo{author}{Dai, D.},
  \bibinfo{author}{Van~Gool, L.}, \bibinfo{year}{2018}b.
\newblock \bibinfo{title}{Learning driving models with a surround-view camera
  system and a route planner}.
\newblock \bibinfo{journal}{ArXiv e-prints} .
\bibitem[{Heess et~al.(2015)Heess, Hunt, Lillicrap and Silver}]{heess-2015}
\bibinfo{author}{Heess, N.}, \bibinfo{author}{Hunt, J.J.},
  \bibinfo{author}{Lillicrap, T.P.}, \bibinfo{author}{Silver, D.},
  \bibinfo{year}{2015}.
\newblock \bibinfo{title}{Memory-based control with recurrent neural networks},
  in: \bibinfo{booktitle}{arXiv preprint arXiv:1512.04455}.
\bibitem[{Herrera et~al.(2010)Herrera, Work, Herring, Ban, Jacobson and
  Bayen}]{herrera2010evaluation}
\bibinfo{author}{Herrera, J.C.}, \bibinfo{author}{Work, D.B.},
  \bibinfo{author}{Herring, R.}, \bibinfo{author}{Ban, X.J.},
  \bibinfo{author}{Jacobson, Q.}, \bibinfo{author}{Bayen, A.M.},
  \bibinfo{year}{2010}.
\newblock \bibinfo{title}{Evaluation of traffic data obtained via gps-enabled
  mobile phones: The mobile century field experiment}.
\newblock \bibinfo{journal}{Transportation Research Part C: Emerging
  Technologies} \bibinfo{volume}{18}, \bibinfo{pages}{568--583}.
\bibitem[{van Hinsbergen et~al.(2009)van Hinsbergen, van Lint, Hoogendoorn and
  van Zuylen}]{bayes_van2009bayesian}
\bibinfo{author}{van Hinsbergen, C.P.I.}, \bibinfo{author}{van Lint, H.W.},
  \bibinfo{author}{Hoogendoorn, S.P.}, \bibinfo{author}{van Zuylen, H.J.},
  \bibinfo{year}{2009}.
\newblock \bibinfo{title}{Bayesian calibration of car-following models}.
\newblock \bibinfo{journal}{IFAC Proceedings Volumes} \bibinfo{volume}{42},
  \bibinfo{pages}{91--97}.
\bibitem[{Ho and Ermon(2016)}]{Ho2-2016}
\bibinfo{author}{Ho, J.}, \bibinfo{author}{Ermon, S.}, \bibinfo{year}{2016}.
\newblock \bibinfo{title}{Generative adversarial imitation learning}, in:
  \bibinfo{booktitle}{Advances in Neural Information Processing Systems
  (NIPS)}, pp. \bibinfo{pages}{4565--4573}.
\bibitem[{Ho et~al.(2016)Ho, Gupta and Ermon}]{Ho-2016}
\bibinfo{author}{Ho, J.}, \bibinfo{author}{Gupta, J.}, \bibinfo{author}{Ermon,
  S.}, \bibinfo{year}{2016}.
\newblock \bibinfo{title}{Model-free imitation learning with policy
  optimization}, in: \bibinfo{booktitle}{Proceedings of International
  Conference on Machine Learning (ICML)}, pp. \bibinfo{pages}{2760--2769}.
\bibitem[{Hodson(2020)}]{hodson2019deepmind}
\bibinfo{author}{Hodson, H.}, \bibinfo{year}{2020}.
\newblock \bibinfo{title}{Deepmind and google: the battle to control artificial
  intelligence}.
\newblock
  \bibinfo{howpublished}{\url{https://www.1843magazine.com/features/deepmind-and-google-the-battle-to-control-artificial-intelligence/}}.
\newblock \bibinfo{note}{[Online; accessed 7.31.2020]}.
\bibitem[{Hoel et~al.(2020)Hoel, Driggs-Campbell, Wolff, Laine and
  Kochenderfer}]{Hoel-2020}
\bibinfo{author}{Hoel, C.J.}, \bibinfo{author}{Driggs-Campbell, K.},
  \bibinfo{author}{Wolff, K.}, \bibinfo{author}{Laine, L.},
  \bibinfo{author}{Kochenderfer, M.J.}, \bibinfo{year}{2020}.
\newblock \bibinfo{title}{Combining planning and deep reinforcement learning in
  tactical decision making for autonomous driving}.
\newblock \bibinfo{journal}{IEEE Transactions on Intelligent Vehicles}
  \bibinfo{volume}{5}, \bibinfo{pages}{294--305}.
\bibitem[{Hofleitner et~al.(2012)Hofleitner, Herring and
  Bayen}]{Hofleitner-2012}
\bibinfo{author}{Hofleitner, A.}, \bibinfo{author}{Herring, R.},
  \bibinfo{author}{Bayen, A.}, \bibinfo{year}{2012}.
\newblock \bibinfo{title}{Arterial travel time forecast with streaming data: A
  hybrid approach of flow modeling and machine learning}.
\newblock \bibinfo{journal}{Transportation Research Part B: Methodological}
  \bibinfo{volume}{46}, \bibinfo{pages}{1097--1122}.
\bibitem[{Hoogendoorn and
  Hoogendoorn(2010)}]{maximum_likelihood_hoogendoorn2010generic}
\bibinfo{author}{Hoogendoorn, S.P.}, \bibinfo{author}{Hoogendoorn, R.},
  \bibinfo{year}{2010}.
\newblock \bibinfo{title}{Generic calibration framework for joint estimation of
  car-following models by using microscopic data}.
\newblock \bibinfo{journal}{Transportation Research Record}
  \bibinfo{volume}{2188}, \bibinfo{pages}{37--45}.
\bibitem[{Huang et~al.(2019)Huang, Di, Du and Chen}]{huang2019stable}
\bibinfo{author}{Huang, K.}, \bibinfo{author}{Di, X.}, \bibinfo{author}{Du,
  Q.}, \bibinfo{author}{Chen, X.}, \bibinfo{year}{2019}.
\newblock \bibinfo{title}{Stabilizing traffic via autonomous vehicles: A
  continuum mean field game approach}.
\newblock \bibinfo{journal}{the 22nd IEEE International Conference on
  Intelligent Transportation Systems (ITSC) (DOI: 10.1109/ITSC.2019.8917021)} .
\bibitem[{Huang et~al.(2020a)Huang, Di, Du and Chen}]{huang2020game}
\bibinfo{author}{Huang, K.}, \bibinfo{author}{Di, X.}, \bibinfo{author}{Du,
  Q.}, \bibinfo{author}{Chen, X.}, \bibinfo{year}{2020}a.
\newblock \bibinfo{title}{A game-theoretic framework for autonomous vehicles
  velocity control: Bridging microscopic differential games and macroscopic
  mean field games}.
\newblock \bibinfo{journal}{Discrete and Continuous Dynamical Systems - Series
  B (accepted) (arXiv preprint arXiv:1903.06053)} .
\bibitem[{Huang et~al.(2020b)Huang, Di, Du and Chen}]{huang2020stable}
\bibinfo{author}{Huang, K.}, \bibinfo{author}{Di, X.}, \bibinfo{author}{Du,
  Q.}, \bibinfo{author}{Chen, X.}, \bibinfo{year}{2020}b.
\newblock \bibinfo{title}{Scalable traffic stability analysis in mixed-autonomy
  using continuum models}.
\newblock \bibinfo{journal}{Transportation Research Part C: Emerging
  Technologies} \bibinfo{volume}{111}, \bibinfo{pages}{616--630}.
\bibitem[{Huang et~al.(2006)Huang, Malham{\'e}, Caines et~al.}]{huang2006large}
\bibinfo{author}{Huang, M.}, \bibinfo{author}{Malham{\'e}, R.P.},
  \bibinfo{author}{Caines, P.E.}, et~al., \bibinfo{year}{2006}.
\newblock \bibinfo{title}{Large population stochastic dynamic games:
  closed-loop mckean-vlasov systems and the nash certainty equivalence
  principle}.
\newblock \bibinfo{journal}{Communications in Information \& Systems}
  \bibinfo{volume}{6}, \bibinfo{pages}{221--252}.
\bibitem[{Huang et~al.(2018)Huang, Sun and Sun}]{huang2018car}
\bibinfo{author}{Huang, X.}, \bibinfo{author}{Sun, J.}, \bibinfo{author}{Sun,
  J.}, \bibinfo{year}{2018}.
\newblock \bibinfo{title}{A car-following model considering asymmetric driving
  behavior based on long short-term memory neural networks}.
\newblock \bibinfo{journal}{Transportation research part C: emerging
  technologies} \bibinfo{volume}{95}, \bibinfo{pages}{346--362}.
\bibitem[{Ioannou and Chien(1993)}]{ioannou1993autonomous}
\bibinfo{author}{Ioannou, P.A.}, \bibinfo{author}{Chien, C.C.},
  \bibinfo{year}{1993}.
\newblock \bibinfo{title}{Autonomous intelligent cruise control}.
\newblock \bibinfo{journal}{IEEE Transactions on Vehicular technology}
  \bibinfo{volume}{42}, \bibinfo{pages}{657--672}.
\bibitem[{{I}nternational~{S}tandard {J}3016(2016)}]{sae}
\bibinfo{author}{{I}nternational~{S}tandard {J}3016, S.}, \bibinfo{year}{2016}.
\newblock \bibinfo{title}{S{AE} international}.
\bibitem[{Jang et~al.(2019)Jang, Vinitsky, Chalaki, Remer, Beaver, Malikopoulos
  and Bayen}]{Jang-2019}
\bibinfo{author}{Jang, K.}, \bibinfo{author}{Vinitsky, E.},
  \bibinfo{author}{Chalaki, B.}, \bibinfo{author}{Remer, B.},
  \bibinfo{author}{Beaver, L.}, \bibinfo{author}{Malikopoulos, A.A.},
  \bibinfo{author}{Bayen, A.}, \bibinfo{year}{2019}.
\newblock \bibinfo{title}{Simulation to scaled city: zero-shot policy transfer
  for traffic control via autonomous vehicles}, in:
  \bibinfo{booktitle}{Proceedings of the 10th ACM/IEEE International Conference
  on Cyber-Physical Systems}, pp. \bibinfo{pages}{291--300}.
\bibitem[{Jaritz et~al.(2018)Jaritz, Charette, Toromanoff, Perot and
  Nashashibi}]{Jaritz-2018}
\bibinfo{author}{Jaritz, M.}, \bibinfo{author}{Charette, R.D.},
  \bibinfo{author}{Toromanoff, M.}, \bibinfo{author}{Perot, E.},
  \bibinfo{author}{Nashashibi, F.}, \bibinfo{year}{2018}.
\newblock \bibinfo{title}{End-to-end race driving with deep reinforcement
  learning}, in: \bibinfo{booktitle}{2018 IEEE International Conference on
  Robotics and Automation (ICRA)}, pp. \bibinfo{pages}{2070--2075}.
\bibitem[{Jin et~al.(2018)Jin, Avedisov, He, Qin, Sadeghpour and
  Orosz}]{jin2018experimental}
\bibinfo{author}{Jin, I.G.}, \bibinfo{author}{Avedisov, S.S.},
  \bibinfo{author}{He, C.R.}, \bibinfo{author}{Qin, W.B.},
  \bibinfo{author}{Sadeghpour, M.}, \bibinfo{author}{Orosz, G.},
  \bibinfo{year}{2018}.
\newblock \bibinfo{title}{Experimental validation of connected automated
  vehicle design among human-driven vehicles}.
\newblock \bibinfo{journal}{Transportation research part C: emerging
  technologies} \bibinfo{volume}{91}, \bibinfo{pages}{335--352}.
\bibitem[{Jin and Orosz(2014)}]{jin2014dynamics}
\bibinfo{author}{Jin, I.G.}, \bibinfo{author}{Orosz, G.}, \bibinfo{year}{2014}.
\newblock \bibinfo{title}{Dynamics of connected vehicle systems with delayed
  acceleration feedback}.
\newblock \bibinfo{journal}{Transportation Research Part C: Emerging
  Technologies} \bibinfo{volume}{46}, \bibinfo{pages}{46--64}.
\bibitem[{Jin and Orosz(2018)}]{jin2018connected}
\bibinfo{author}{Jin, I.G.}, \bibinfo{author}{Orosz, G.}, \bibinfo{year}{2018}.
\newblock \bibinfo{title}{Connected cruise control among human-driven vehicles:
  Experiment-based parameter estimation and optimal control design}.
\newblock \bibinfo{journal}{Transportation Research Part C: Emerging
  Technologies} \bibinfo{volume}{95}, \bibinfo{pages}{445--459}.
\bibitem[{Johnson-Roberson et~al.(2017)Johnson-Roberson, Barto, Mehta, Sridhar,
  Rosaen and Vasudevan}]{Johnson-Roberson-2017}
\bibinfo{author}{Johnson-Roberson, M.}, \bibinfo{author}{Barto, C.},
  \bibinfo{author}{Mehta, R.}, \bibinfo{author}{Sridhar, S.N.},
  \bibinfo{author}{Rosaen, K.}, \bibinfo{author}{Vasudevan, R.},
  \bibinfo{year}{2017}.
\newblock \bibinfo{title}{Driving in the matrix: Can virtual worlds replace
  human-generated annotations for real world tasks?}, in:
  \bibinfo{booktitle}{Proceedings of the 2017 IEEE International Conference on
  Robotics and Automation (ICRA)}, pp. \bibinfo{pages}{746--753}.
\bibitem[{Kasai et~al.(2013)Kasai, Shibagaki and Terabe}]{kasai2013application}
\bibinfo{author}{Kasai, M.}, \bibinfo{author}{Shibagaki, S.},
  \bibinfo{author}{Terabe, S.}, \bibinfo{year}{2013}.
\newblock \bibinfo{title}{Application of hierarchical bayesian estimation to
  calibrating a car-following model with time-varying parameters}, in:
  \bibinfo{booktitle}{2013 IEEE Intelligent Vehicles Symposium (IV)},
  \bibinfo{organization}{IEEE}. pp. \bibinfo{pages}{870--875}.
\bibitem[{Kasper et~al.(2012)Kasper, Weidl, Dang, Breuel, Tamke, Wedel and
  Rosenstiel}]{kasper2012object}
\bibinfo{author}{Kasper, D.}, \bibinfo{author}{Weidl, G.},
  \bibinfo{author}{Dang, T.}, \bibinfo{author}{Breuel, G.},
  \bibinfo{author}{Tamke, A.}, \bibinfo{author}{Wedel, A.},
  \bibinfo{author}{Rosenstiel, W.}, \bibinfo{year}{2012}.
\newblock \bibinfo{title}{Object-oriented bayesian networks for detection of
  lane change maneuvers}.
\newblock \bibinfo{journal}{IEEE Intelligent Transportation Systems Magazine}
  \bibinfo{volume}{4}, \bibinfo{pages}{19--31}.
\bibitem[{Katrakazas et~al.(2015)Katrakazas, Quddus, Chen and
  Deka}]{katrakazas2015real}
\bibinfo{author}{Katrakazas, C.}, \bibinfo{author}{Quddus, M.},
  \bibinfo{author}{Chen, W.H.}, \bibinfo{author}{Deka, L.},
  \bibinfo{year}{2015}.
\newblock \bibinfo{title}{Real-time motion planning methods for autonomous
  on-road driving: State-of-the-art and future research directions}.
\newblock \bibinfo{journal}{Transportation Research Part C: Emerging
  Technologies} \bibinfo{volume}{60}, \bibinfo{pages}{416--442}.
\bibitem[{Kesten et~al.(2019)Kesten, Usman, Houston, Pandya, Nadhamuni,
  Ferreira, Yuan, Low, Jain, Ondruska, Omari, Shah, Kulkarni, Kazakova, Tao,
  Platinsky, Jiang and Shet}]{Lyft-2019}
\bibinfo{author}{Kesten, R.}, \bibinfo{author}{Usman, M.},
  \bibinfo{author}{Houston, J.}, \bibinfo{author}{Pandya, T.},
  \bibinfo{author}{Nadhamuni, K.}, \bibinfo{author}{Ferreira, A.},
  \bibinfo{author}{Yuan, M.}, \bibinfo{author}{Low, B.}, \bibinfo{author}{Jain,
  A.}, \bibinfo{author}{Ondruska, P.}, \bibinfo{author}{Omari, S.},
  \bibinfo{author}{Shah, S.}, \bibinfo{author}{Kulkarni, A.},
  \bibinfo{author}{Kazakova, A.}, \bibinfo{author}{Tao, C.},
  \bibinfo{author}{Platinsky, L.}, \bibinfo{author}{Jiang, W.},
  \bibinfo{author}{Shet, V.}, \bibinfo{year}{2019}.
\newblock \bibinfo{title}{Lyft {L}evel 5 {AV} dataset}.
\newblock
  \bibinfo{howpublished}{\url{https://self-driving.lyft.com/level5/data}}.
\bibitem[{Kesting et~al.(2007)Kesting, Treiber and Helbing}]{Kesting-2007}
\bibinfo{author}{Kesting, A.}, \bibinfo{author}{Treiber, M.},
  \bibinfo{author}{Helbing, D.}, \bibinfo{year}{2007}.
\newblock \bibinfo{title}{General lane-changing model mobil for car-following
  models}.
\newblock \bibinfo{journal}{Transportation Research Record}
  \bibinfo{volume}{1999}, \bibinfo{pages}{86--94}.
\bibitem[{Kesting et~al.(2010)Kesting, Treiber and
  Helbing}]{kesting2010enhanced}
\bibinfo{author}{Kesting, A.}, \bibinfo{author}{Treiber, M.},
  \bibinfo{author}{Helbing, D.}, \bibinfo{year}{2010}.
\newblock \bibinfo{title}{Enhanced intelligent driver model to access the
  impact of driving strategies on traffic capacity}.
\newblock \bibinfo{journal}{Philosophical Transactions of the Royal Society of
  London A: Mathematical, Physical and Engineering Sciences}
  \bibinfo{volume}{368}, \bibinfo{pages}{4585--4605}.
\bibitem[{Khodayari et~al.(2012)Khodayari, Ghaffari, Kazemi and
  Braunstingl}]{NN_sgd_khodayari2012modified}
\bibinfo{author}{Khodayari, A.}, \bibinfo{author}{Ghaffari, A.},
  \bibinfo{author}{Kazemi, R.}, \bibinfo{author}{Braunstingl, R.},
  \bibinfo{year}{2012}.
\newblock \bibinfo{title}{A modified car-following model based on a neural
  network model of the human driver effects}.
\newblock \bibinfo{journal}{IEEE Transactions on Systems, Man, and
  Cybernetics-Part A: Systems and Humans} \bibinfo{volume}{42},
  \bibinfo{pages}{1440--1449}.
\bibitem[{Kim and Langari(2014)}]{kim2014game}
\bibinfo{author}{Kim, C.}, \bibinfo{author}{Langari, R.}, \bibinfo{year}{2014}.
\newblock \bibinfo{title}{Game theory based autonomous vehicles operation}.
\newblock \bibinfo{journal}{International Journal of Vehicle Design}
  \bibinfo{volume}{65}, \bibinfo{pages}{360--383}.
\bibitem[{Kiran et~al.(2020)Kiran, Sobh, Talpaert, Mannion, Sallab, Yogamani
  and Pérez}]{Kiran-2020-arXiv}
\bibinfo{author}{Kiran, B.R.}, \bibinfo{author}{Sobh, I.},
  \bibinfo{author}{Talpaert, V.}, \bibinfo{author}{Mannion, P.},
  \bibinfo{author}{Sallab, A.A.A.}, \bibinfo{author}{Yogamani, S.},
  \bibinfo{author}{Pérez, P.}, \bibinfo{year}{2020}.
\newblock \bibinfo{title}{Deep reinforcement learning for autonomous driving: A
  survey}.
\newblock \bibinfo{journal}{arXiv preprint arXiv:2002.00444} .
\bibitem[{Kockelman(2017)}]{kockelman2017assessment}
\bibinfo{author}{Kockelman, K.}, \bibinfo{year}{2017}.
\newblock \bibinfo{title}{An Assessment of Autonomous Vehicles: Traffic Impacts
  and Infrastructure Needs--final Report}.
\newblock \bibinfo{publisher}{Center for Transportation Research, The
  University of Texas at Austin}.
\bibitem[{Kreidieh et~al.(2018a)Kreidieh, Wu and Bayen}]{Kreidieh-2018}
\bibinfo{author}{Kreidieh, A.}, \bibinfo{author}{Wu, C.},
  \bibinfo{author}{Bayen, A.M.}, \bibinfo{year}{2018}a.
\newblock \bibinfo{title}{Dissipating stop-and-go waves in closed and open
  networks via deep reinforcement learning}, in: \bibinfo{booktitle}{21st IEEE
  International Conference on Intelligent Transportation Systems (ITSC)}, pp.
  \bibinfo{pages}{1475--1480}.
\bibitem[{Kreidieh et~al.(2018b)Kreidieh, Wu and
  Bayen}]{kreidieh2018dissipating}
\bibinfo{author}{Kreidieh, A.R.}, \bibinfo{author}{Wu, C.},
  \bibinfo{author}{Bayen, A.M.}, \bibinfo{year}{2018}b.
\newblock \bibinfo{title}{Dissipating stop-and-go waves in closed and open
  networks via deep reinforcement learning}, in: \bibinfo{booktitle}{2018 21st
  International Conference on Intelligent Transportation Systems (ITSC)},
  \bibinfo{organization}{IEEE}. pp. \bibinfo{pages}{1475--1480}.
\bibitem[{Kuefler et~al.(2017)Kuefler, Morton, Wheeler and
  Kochenderfer}]{Kuefler-2017}
\bibinfo{author}{Kuefler, A.}, \bibinfo{author}{Morton, J.},
  \bibinfo{author}{Wheeler, T.}, \bibinfo{author}{Kochenderfer, M.},
  \bibinfo{year}{2017}.
\newblock \bibinfo{title}{Imitating driver behavior with generative adversarial
  networks}, in: \bibinfo{booktitle}{Proceedings of 2017 IEEE Intelligent
  Vehicles Symposium (IV)}, pp. \bibinfo{pages}{204--211}.
\bibitem[{Kumagai and Akamatsu(2006)}]{Toru-2006}
\bibinfo{author}{Kumagai, T.}, \bibinfo{author}{Akamatsu, M.},
  \bibinfo{year}{2006}.
\newblock \bibinfo{title}{Prediction of human driving behavior using dynamic
  bayesian networks}.
\newblock \bibinfo{journal}{IEICE TRANSACTIONS on Information and Systems}
  \bibinfo{volume}{89}, \bibinfo{pages}{857--860}.
\bibitem[{Kumar et~al.(2013)Kumar, Perrollaz, Lefevre and Laugier}]{Kumar-2013}
\bibinfo{author}{Kumar, P.}, \bibinfo{author}{Perrollaz, M.},
  \bibinfo{author}{Lefevre, S.}, \bibinfo{author}{Laugier, C.},
  \bibinfo{year}{2013}.
\newblock \bibinfo{title}{Learning-based approach for online lane change
  intention prediction}, in: \bibinfo{booktitle}{Proceedings of 2013 IEEE
  Intelligent Vehicles Symposium (IV)}, pp. \bibinfo{pages}{797--802}.
\bibitem[{Lachapelle et~al.(2010)Lachapelle, Salomon and
  Turinici}]{lachapelle2010computation}
\bibinfo{author}{Lachapelle, A.}, \bibinfo{author}{Salomon, J.},
  \bibinfo{author}{Turinici, G.}, \bibinfo{year}{2010}.
\newblock \bibinfo{title}{Computation of mean field equilibria in economics}.
\newblock \bibinfo{journal}{Mathematical Models and Methods in Applied
  Sciences} \bibinfo{volume}{20}, \bibinfo{pages}{567--588}.
\bibitem[{Lachapelle and Wolfram(2011)}]{lachapelle2011mean}
\bibinfo{author}{Lachapelle, A.}, \bibinfo{author}{Wolfram, M.T.},
  \bibinfo{year}{2011}.
\newblock \bibinfo{title}{On a mean field game approach modeling congestion and
  aversion in pedestrian crowds}.
\newblock \bibinfo{journal}{Transportation research part B: methodological}
  \bibinfo{volume}{45}, \bibinfo{pages}{1572--1589}.
\bibitem[{Lasry and Lions(2007)}]{lasry2007mean}
\bibinfo{author}{Lasry, J.M.}, \bibinfo{author}{Lions, P.L.},
  \bibinfo{year}{2007}.
\newblock \bibinfo{title}{Mean field games}.
\newblock \bibinfo{journal}{Japanese journal of mathematics}
  \bibinfo{volume}{2}, \bibinfo{pages}{229--260}.
\bibitem[{Lazar et~al.(2018a)Lazar, Chandrasekher, Pedarsani and
  Sadigh}]{lazar2018maximizing}
\bibinfo{author}{Lazar, D.A.}, \bibinfo{author}{Chandrasekher, K.},
  \bibinfo{author}{Pedarsani, R.}, \bibinfo{author}{Sadigh, D.},
  \bibinfo{year}{2018}a.
\newblock \bibinfo{title}{Maximizing road capacity using cars that influence
  people}.
\newblock \bibinfo{journal}{arXiv preprint arXiv:1807.04414} .
\bibitem[{Lazar et~al.(2018b)Lazar, Pedarsani, Chandrasekher and
  Sadigh}]{Lazar-2018}
\bibinfo{author}{Lazar, D.A.}, \bibinfo{author}{Pedarsani, R.},
  \bibinfo{author}{Chandrasekher, K.}, \bibinfo{author}{Sadigh, D.},
  \bibinfo{year}{2018}b.
\newblock \bibinfo{title}{Maximizing road capacity using cars that influence
  people}, in: \bibinfo{booktitle}{2018 IEEE Conference on Decision and Control
  (CDC)}, pp. \bibinfo{pages}{1801--1808}.
\bibitem[{Lee and Ozbay(2009a)}]{bayes_and_SPSA}
\bibinfo{author}{Lee, J.B.}, \bibinfo{author}{Ozbay, K.},
  \bibinfo{year}{2009}a.
\newblock \bibinfo{title}{New calibration methodology for microscopic traffic
  simulation using enhanced simultaneous perturbation stochastic approximation
  approach}.
\newblock \bibinfo{journal}{Transportation Research Record}
  \bibinfo{volume}{2124}, \bibinfo{pages}{233--240}.
\bibitem[{Lee and Ozbay(2009b)}]{search_SPAP_1}
\bibinfo{author}{Lee, J.B.}, \bibinfo{author}{Ozbay, K.},
  \bibinfo{year}{2009}b.
\newblock \bibinfo{title}{New calibration methodology for microscopic traffic
  simulation using enhanced simultaneous perturbation stochastic approximation
  approach}.
\newblock \bibinfo{journal}{Transportation Research Record}
  \bibinfo{volume}{2124}, \bibinfo{pages}{233--240}.
\bibitem[{Levin and Boyles(2015)}]{levin2015intersection}
\bibinfo{author}{Levin, M.W.}, \bibinfo{author}{Boyles, S.D.},
  \bibinfo{year}{2015}.
\newblock \bibinfo{title}{Intersection auctions and reservation-based control
  in dynamic traffic assignment}.
\newblock \bibinfo{journal}{Transportation Research Record: Journal of the
  Transportation Research Board} , \bibinfo{pages}{35--44}.
\bibitem[{Levin and Boyles(2016)}]{levin2016multiclass}
\bibinfo{author}{Levin, M.W.}, \bibinfo{author}{Boyles, S.D.},
  \bibinfo{year}{2016}.
\newblock \bibinfo{title}{A multiclass cell transmission model for shared human
  and autonomous vehicle roads}.
\newblock \bibinfo{journal}{Transportation Research Part C: Emerging
  Technologies} \bibinfo{volume}{62}, \bibinfo{pages}{103--116}.
\bibitem[{Levine and Koltun(2012)}]{levine2012continuous}
\bibinfo{author}{Levine, S.}, \bibinfo{author}{Koltun, V.},
  \bibinfo{year}{2012}.
\newblock \bibinfo{title}{Continuous inverse optimal control with locally
  optimal examples}.
\newblock \bibinfo{journal}{arXiv preprint arXiv:1206.4617} .
\bibitem[{Li et~al.(2016)Li, Wang, Xu and Wang}]{li2016lane}
\bibinfo{author}{Li, K.}, \bibinfo{author}{Wang, X.}, \bibinfo{author}{Xu, Y.},
  \bibinfo{author}{Wang, J.}, \bibinfo{year}{2016}.
\newblock \bibinfo{title}{Lane changing intention recognition based on speech
  recognition models}.
\newblock \bibinfo{journal}{Transportation research part C: emerging
  technologies} \bibinfo{volume}{69}, \bibinfo{pages}{497--514}.
\bibitem[{Li et~al.(2014)Li, Wen and Yao}]{li2014survey}
\bibinfo{author}{Li, L.}, \bibinfo{author}{Wen, D.}, \bibinfo{author}{Yao, D.},
  \bibinfo{year}{2014}.
\newblock \bibinfo{title}{A survey of traffic control with vehicular
  communications} .
\bibitem[{Li et~al.(2019)Li, Qin, Jiao, Yang, Wang, Wang, Wu and
  Ye}]{li_efficient_2019}
\bibinfo{author}{Li, M.}, \bibinfo{author}{Qin, Z.}, \bibinfo{author}{Jiao,
  Y.}, \bibinfo{author}{Yang, Y.}, \bibinfo{author}{Wang, J.},
  \bibinfo{author}{Wang, C.}, \bibinfo{author}{Wu, G.}, \bibinfo{author}{Ye,
  J.}, \bibinfo{year}{2019}.
\newblock \bibinfo{title}{Efficient {Ridesharing} {Order} {Dispatching} with
  {Mean} {Field} {Multi}-{Agent} {Reinforcement} {Learning}}, in:
  \bibinfo{booktitle}{The {World} {Wide} {Web} {Conference}},
  \bibinfo{publisher}{ACM}, \bibinfo{address}{New York, NY, USA}. pp.
  \bibinfo{pages}{983--994}.
\newblock \bibinfo{note}{Event-place: San Francisco, CA, USA}.
\bibitem[{Li et~al.(2018a)Li, Oyler, Zhang, Yildiz, Kolmanovsky and
  Girard}]{li2018game}
\bibinfo{author}{Li, N.}, \bibinfo{author}{Oyler, D.W.},
  \bibinfo{author}{Zhang, M.}, \bibinfo{author}{Yildiz, Y.},
  \bibinfo{author}{Kolmanovsky, I.}, \bibinfo{author}{Girard, A.R.},
  \bibinfo{year}{2018}a.
\newblock \bibinfo{title}{Game theoretic modeling of driver and vehicle
  interactions for verification and validation of autonomous vehicle control
  systems}.
\newblock \bibinfo{journal}{IEEE Transactions on control systems technology}
  \bibinfo{volume}{26}, \bibinfo{pages}{1782--1797}.
\bibitem[{Li et~al.(2017a)Li, Zheng, Li, Wang and Zhang}]{li2017platoon}
\bibinfo{author}{Li, S.E.}, \bibinfo{author}{Zheng, Y.}, \bibinfo{author}{Li,
  K.}, \bibinfo{author}{Wang, L.Y.}, \bibinfo{author}{Zhang, H.},
  \bibinfo{year}{2017}a.
\newblock \bibinfo{title}{Platoon control of connected vehicles from a
  networked control perspective: Literature review, component modeling, and
  controller synthesis}.
\newblock \bibinfo{journal}{IEEE Transactions on Vehicular Technology} .
\bibitem[{Li et~al.(2017b)Li, Zheng, Li, Wu, Hedrick, Gao and
  Zhang}]{li2017dynamical}
\bibinfo{author}{Li, S.E.}, \bibinfo{author}{Zheng, Y.}, \bibinfo{author}{Li,
  K.}, \bibinfo{author}{Wu, Y.}, \bibinfo{author}{Hedrick, J.K.},
  \bibinfo{author}{Gao, F.}, \bibinfo{author}{Zhang, H.},
  \bibinfo{year}{2017}b.
\newblock \bibinfo{title}{Dynamical modeling and distributed control of
  connected and automated vehicles: Challenges and opportunities}.
\newblock \bibinfo{journal}{IEEE Intelligent Transportation Systems Magazine}
  \bibinfo{volume}{9}, \bibinfo{pages}{46--58}.
\bibitem[{Li et~al.(2018b)Li, Ghiasi, Xu and Qu}]{li2018piecewise}
\bibinfo{author}{Li, X.}, \bibinfo{author}{Ghiasi, A.}, \bibinfo{author}{Xu,
  Z.}, \bibinfo{author}{Qu, X.}, \bibinfo{year}{2018}b.
\newblock \bibinfo{title}{A piecewise trajectory optimization model for
  connected automated vehicles: Exact optimization algorithm and queue
  propagation analysis}.
\newblock \bibinfo{journal}{Transportation Research Part B: Methodological}
  \bibinfo{volume}{118}, \bibinfo{pages}{429--456}.
\bibitem[{Li et~al.(2018c)Li, Tang, Li, He, Peeta and Wang}]{li2018consensus}
\bibinfo{author}{Li, Y.}, \bibinfo{author}{Tang, C.}, \bibinfo{author}{Li, K.},
  \bibinfo{author}{He, X.}, \bibinfo{author}{Peeta, S.}, \bibinfo{author}{Wang,
  Y.}, \bibinfo{year}{2018}c.
\newblock \bibinfo{title}{Consensus-based cooperative control for multi-platoon
  under the connected vehicles environment}.
\newblock \bibinfo{journal}{IEEE Transactions on Intelligent Transportation
  Systems} \bibinfo{volume}{20}, \bibinfo{pages}{2220--2229}.
\bibitem[{Li et~al.(2018d)Li, Tang, Peeta and Wang}]{li2018nonlinear}
\bibinfo{author}{Li, Y.}, \bibinfo{author}{Tang, C.}, \bibinfo{author}{Peeta,
  S.}, \bibinfo{author}{Wang, Y.}, \bibinfo{year}{2018}d.
\newblock \bibinfo{title}{Nonlinear consensus-based connected vehicle platoon
  control incorporating car-following interactions and heterogeneous time
  delays}.
\newblock \bibinfo{journal}{IEEE Transactions on Intelligent Transportation
  Systems} .
\bibitem[{Lighthill and Whitham(1955)}]{lighthill1955kinematic}
\bibinfo{author}{Lighthill, M.J.}, \bibinfo{author}{Whitham, G.B.},
  \bibinfo{year}{1955}.
\newblock \bibinfo{title}{On kinematic waves ii. a theory of traffic flow on
  long crowded roads}.
\newblock \bibinfo{journal}{Proc. R. Soc. Lond. A} \bibinfo{volume}{229},
  \bibinfo{pages}{317--345}.
\bibitem[{Lillicrap et~al.(2015)Lillicrap, Hunt, Pritzel, Heess, Erez, Tassa,
  Silver and Wierstra}]{Lill-2015}
\bibinfo{author}{Lillicrap, T.P.}, \bibinfo{author}{Hunt, J.J.},
  \bibinfo{author}{Pritzel, A.}, \bibinfo{author}{Heess, N.},
  \bibinfo{author}{Erez, T.}, \bibinfo{author}{Tassa, Y.},
  \bibinfo{author}{Silver, D.}, \bibinfo{author}{Wierstra, D.},
  \bibinfo{year}{2015}.
\newblock \bibinfo{title}{Continuous control with deep reinforcement learning},
  in: \bibinfo{booktitle}{arXiv preprint arXiv:1509.02971}.
\bibitem[{Lin et~al.(2018)Lin, Zhao, Xu and Zhou}]{lin_efficient_2018}
\bibinfo{author}{Lin, K.}, \bibinfo{author}{Zhao, R.}, \bibinfo{author}{Xu,
  Z.}, \bibinfo{author}{Zhou, J.}, \bibinfo{year}{2018}.
\newblock \bibinfo{title}{Efficient {Large}-{Scale} {Fleet} {Management} via
  {Multi}-{Agent} {Deep} {Reinforcement} {Learning}}, in:
  \bibinfo{booktitle}{Proceedings of the 24th {ACM} {SIGKDD} {International}
  {Conference} on {Knowledge} {Discovery} \& {Data} {Mining}},
  \bibinfo{publisher}{ACM}, \bibinfo{address}{New York, NY, USA}. pp.
  \bibinfo{pages}{1774--1783}.
\bibitem[{Littman(1994)}]{littman_markov_1994}
\bibinfo{author}{Littman, M.L.}, \bibinfo{year}{1994}.
\newblock \bibinfo{title}{Markov {Games} {As} a {Framework} for {Multi}-agent
  {Reinforcement} {Learning}}, in: \bibinfo{booktitle}{Proceedings of the
  {Eleventh} {International} {Conference} on {International} {Conference} on
  {Machine} {Learning}}, \bibinfo{publisher}{Morgan Kaufmann Publishers Inc.},
  \bibinfo{address}{San Francisco, CA, USA}. pp. \bibinfo{pages}{157--163}.
\newblock \bibinfo{note}{Event-place: New Brunswick, NJ, USA}.
\bibitem[{Liu et~al.(2018)Liu, Lin, Shiraishi and Tomizuka}]{liu2018improving}
\bibinfo{author}{Liu, C.}, \bibinfo{author}{Lin, C.W.},
  \bibinfo{author}{Shiraishi, S.}, \bibinfo{author}{Tomizuka, M.},
  \bibinfo{year}{2018}.
\newblock \bibinfo{title}{Improving efficiency of autonomous vehicles by v2v
  communication}, in: \bibinfo{booktitle}{2018 Annual American Control
  Conference (ACC)}, \bibinfo{organization}{IEEE}. pp.
  \bibinfo{pages}{4778--4783}.
\bibitem[{Liu and Tomizuka(2015)}]{liu2015safe}
\bibinfo{author}{Liu, C.}, \bibinfo{author}{Tomizuka, M.},
  \bibinfo{year}{2015}.
\newblock \bibinfo{title}{Safe exploration: Addressing various uncertainty
  levels in human robot interactions.}, in: \bibinfo{booktitle}{ACC}, pp.
  \bibinfo{pages}{465--470}.
\bibitem[{Liu and Tomizuka(2016)}]{liu2016enabling}
\bibinfo{author}{Liu, C.}, \bibinfo{author}{Tomizuka, M.},
  \bibinfo{year}{2016}.
\newblock \bibinfo{title}{Enabling safe freeway driving for automated
  vehicles}, in: \bibinfo{booktitle}{American Control Conference (ACC), 2016},
  \bibinfo{organization}{IEEE}. pp. \bibinfo{pages}{3461--3467}.
\bibitem[{Liu et~al.(2007)Liu, Xin, Adam and Ban}]{liu2007game}
\bibinfo{author}{Liu, H.X.}, \bibinfo{author}{Xin, W.}, \bibinfo{author}{Adam,
  Z.}, \bibinfo{author}{Ban, J.}, \bibinfo{year}{2007}.
\newblock \bibinfo{title}{A game theoretical approach for modelling merging and
  yielding behaviour at freeway on-ramp sections}.
\newblock \bibinfo{journal}{Transportation and traffic theory}
  \bibinfo{volume}{3}, \bibinfo{pages}{197--211}.
\bibitem[{Lowe et~al.(2017)Lowe, Wu, Tamar, Harb, Abbeel and
  Mordatch}]{lowe_multi-agent_2017}
\bibinfo{author}{Lowe, R.}, \bibinfo{author}{Wu, Y.}, \bibinfo{author}{Tamar,
  A.}, \bibinfo{author}{Harb, J.}, \bibinfo{author}{Abbeel, P.},
  \bibinfo{author}{Mordatch, I.}, \bibinfo{year}{2017}.
\newblock \bibinfo{title}{Multi-agent {Actor}-critic for {Mixed}
  {Cooperative}-competitive {Environments}}, in:
  \bibinfo{booktitle}{Proceedings of the 31st {International} {Conference} on
  {Neural} {Information} {Processing} {Systems}}, \bibinfo{publisher}{Curran
  Associates Inc.}, \bibinfo{address}{USA}. pp. \bibinfo{pages}{6382--6393}.
\newblock \bibinfo{note}{Event-place: Long Beach, California, USA}.
\bibitem[{Lu and Skabardonis(2007)}]{camera_lu2007freeway}
\bibinfo{author}{Lu, X.Y.}, \bibinfo{author}{Skabardonis, A.},
  \bibinfo{year}{2007}.
\newblock \bibinfo{title}{Freeway traffic shockwave analysis: exploring the
  ngsim trajectory data}, in: \bibinfo{booktitle}{86th Annual Meeting of the
  Transportation Research Board, Washington, DC}.
\bibitem[{Ma et~al.(2016)Ma, Li, Shladover, Rakha, Lu, Jagannathan and
  Dailey}]{ma2016freeway}
\bibinfo{author}{Ma, J.}, \bibinfo{author}{Li, X.}, \bibinfo{author}{Shladover,
  S.E.}, \bibinfo{author}{Rakha, H.A.}, \bibinfo{author}{Lu, X.Y.},
  \bibinfo{author}{Jagannathan, R.}, \bibinfo{author}{Dailey, D.J.},
  \bibinfo{year}{2016}.
\newblock \bibinfo{title}{Freeway speed harmonization}.
\newblock \bibinfo{journal}{IEEE Trans. Intelligent Vehicles}
  \bibinfo{volume}{1}, \bibinfo{pages}{78--89}.
\bibitem[{Ma and Abdulhai(2002)}]{search_GA_1}
\bibinfo{author}{Ma, T.}, \bibinfo{author}{Abdulhai, B.}, \bibinfo{year}{2002}.
\newblock \bibinfo{title}{Genetic algorithm-based optimization approach and
  generic tool for calibrating traffic microscopic simulation parameters}.
\newblock \bibinfo{journal}{Transportation Research Record: Journal of the
  Transportation Research Board} , \bibinfo{pages}{6--15}.
\bibitem[{Malikopoulos et~al.(2018)Malikopoulos, Hong, Park, Lee and
  Ryu}]{malikopoulos2018optimal}
\bibinfo{author}{Malikopoulos, A.A.}, \bibinfo{author}{Hong, S.},
  \bibinfo{author}{Park, B.B.}, \bibinfo{author}{Lee, J.},
  \bibinfo{author}{Ryu, S.}, \bibinfo{year}{2018}.
\newblock \bibinfo{title}{Optimal control for speed harmonization of automated
  vehicles}.
\newblock \bibinfo{journal}{IEEE Transactions on Intelligent Transportation
  Systems} , \bibinfo{pages}{1--13}.
\bibitem[{Mao et~al.(2018)Mao, Zhang, He, Lin, Kale, Stein and
  Kostic}]{camera_mao2018aic2018}
\bibinfo{author}{Mao, T.}, \bibinfo{author}{Zhang, W.}, \bibinfo{author}{He,
  H.}, \bibinfo{author}{Lin, Y.}, \bibinfo{author}{Kale, V.},
  \bibinfo{author}{Stein, A.}, \bibinfo{author}{Kostic, Z.},
  \bibinfo{year}{2018}.
\newblock \bibinfo{title}{Aic2018 report: Traffic surveillance research}, in:
  \bibinfo{booktitle}{Proceedings of the IEEE Conference on Computer Vision and
  Pattern Recognition Workshops}, pp. \bibinfo{pages}{85--92}.
\bibitem[{Matignon et~al.(2012)Matignon, Laurent and
  Fort-Piat}]{matignon_independent_2012}
\bibinfo{author}{Matignon, L.}, \bibinfo{author}{Laurent, G.J.},
  \bibinfo{author}{Fort-Piat, N.L.}, \bibinfo{year}{2012}.
\newblock \bibinfo{title}{Independent reinforcement learners in cooperative
  {Markov} games: a survey regarding coordination problems}.
\newblock \bibinfo{journal}{The Knowledge Engineering Review}
  \bibinfo{volume}{27}, \bibinfo{pages}{1--31}.
\bibitem[{McLaughlin and Hankey(2015)}]{mclaughlin2015naturalistic}
\bibinfo{author}{McLaughlin, S.B.}, \bibinfo{author}{Hankey, J.M.},
  \bibinfo{year}{2015}.
\newblock \bibinfo{title}{Naturalistic Driving Study: Linking the Study Data to
  the Roadway Information Database}.
\newblock \bibinfo{number}{SHRP 2 Report S2-S31-RW-3}.
\bibitem[{Melson et~al.(2018)Melson, Levin, Hammit and
  Boyles}]{melson2018dynamic}
\bibinfo{author}{Melson, C.L.}, \bibinfo{author}{Levin, M.W.},
  \bibinfo{author}{Hammit, B.E.}, \bibinfo{author}{Boyles, S.D.},
  \bibinfo{year}{2018}.
\newblock \bibinfo{title}{Dynamic traffic assignment of cooperative adaptive
  cruise control}.
\newblock \bibinfo{journal}{Transportation Research Part C: Emerging
  Technologies} \bibinfo{volume}{90}, \bibinfo{pages}{114--133}.
\bibitem[{Milan{\'e}s and Shladover(2014)}]{milanes2014modeling}
\bibinfo{author}{Milan{\'e}s, V.}, \bibinfo{author}{Shladover, S.E.},
  \bibinfo{year}{2014}.
\newblock \bibinfo{title}{Modeling cooperative and autonomous adaptive cruise
  control dynamic responses using experimental data}.
\newblock \bibinfo{journal}{Transportation Research Part C: Emerging
  Technologies} \bibinfo{volume}{48}, \bibinfo{pages}{285--300}.
\bibitem[{Milan{\'e}s et~al.(2014)Milan{\'e}s, Shladover, Spring, Nowakowski,
  Kawazoe and Nakamura}]{milanes2014cooperative}
\bibinfo{author}{Milan{\'e}s, V.}, \bibinfo{author}{Shladover, S.E.},
  \bibinfo{author}{Spring, J.}, \bibinfo{author}{Nowakowski, C.},
  \bibinfo{author}{Kawazoe, H.}, \bibinfo{author}{Nakamura, M.},
  \bibinfo{year}{2014}.
\newblock \bibinfo{title}{Cooperative adaptive cruise control in real traffic
  situations}.
\newblock \bibinfo{journal}{IEEE Transactions on Intelligent Transportation
  Systems} \bibinfo{volume}{15}, \bibinfo{pages}{296--305}.
\bibitem[{Millard-Ball(2016)}]{millard2016pedestrians}
\bibinfo{author}{Millard-Ball, A.}, \bibinfo{year}{2016}.
\newblock \bibinfo{title}{Pedestrians, autonomous vehicles, and cities}.
\newblock \bibinfo{journal}{Journal of Planning Education and Research} ,
  \bibinfo{pages}{0739456X16675674}.
\bibitem[{Mnih et~al.(2015)Mnih, Kavukcuoglu, Silver, Rusu, Veness, Bellemare,
  Graves, Riedmiller, Fidjeland, Ostrovski, Petersen, Beattie, Sadik,
  Antonoglou, King, Kumaran, Wierstra, Legg and Hassabis}]{Mnih-2015}
\bibinfo{author}{Mnih, V.}, \bibinfo{author}{Kavukcuoglu, K.},
  \bibinfo{author}{Silver, D.}, \bibinfo{author}{Rusu, A.A.},
  \bibinfo{author}{Veness, J.}, \bibinfo{author}{Bellemare, M.G.},
  \bibinfo{author}{Graves, A.}, \bibinfo{author}{Riedmiller, M.},
  \bibinfo{author}{Fidjeland, A.K.}, \bibinfo{author}{Ostrovski, G.},
  \bibinfo{author}{Petersen, S.}, \bibinfo{author}{Beattie, C.},
  \bibinfo{author}{Sadik, A.}, \bibinfo{author}{Antonoglou, I.},
  \bibinfo{author}{King, H.}, \bibinfo{author}{Kumaran, D.},
  \bibinfo{author}{Wierstra, D.}, \bibinfo{author}{Legg, S.},
  \bibinfo{author}{Hassabis, D.}, \bibinfo{year}{2015}.
\newblock \bibinfo{title}{Human-level control through deep reinforcement
  learning}.
\newblock \bibinfo{journal}{Nature} \bibinfo{volume}{518},
  \bibinfo{pages}{529--533}.
\bibitem[{Mukadam et~al.(2017)Mukadam, Cosgun, Nakhaei and
  Fujimura}]{mukadam2017tactical}
\bibinfo{author}{Mukadam, M.}, \bibinfo{author}{Cosgun, A.},
  \bibinfo{author}{Nakhaei, A.}, \bibinfo{author}{Fujimura, K.},
  \bibinfo{year}{2017}.
\newblock \bibinfo{title}{Tactical decision making for lane changing with deep
  reinforcement learning} .
\bibitem[{M{\"u}ller et~al.(2016)M{\"u}ller, Risto and
  Emmenegger}]{muller2016social}
\bibinfo{author}{M{\"u}ller, L.}, \bibinfo{author}{Risto, M.},
  \bibinfo{author}{Emmenegger, C.}, \bibinfo{year}{2016}.
\newblock \bibinfo{title}{The social behavior of autonomous vehicles}, in:
  \bibinfo{booktitle}{Proceedings of the 2016 ACM International Joint
  Conference on Pervasive and Ubiquitous Computing: Adjunct}, pp.
  \bibinfo{pages}{686--689}.
\bibitem[{Muller et~al.(2006)Muller, Ben, Cosatto, Flepp and Cun}]{Muller-2006}
\bibinfo{author}{Muller, U.}, \bibinfo{author}{Ben, J.},
  \bibinfo{author}{Cosatto, E.}, \bibinfo{author}{Flepp, B.},
  \bibinfo{author}{Cun, Y.}, \bibinfo{year}{2006}.
\newblock \bibinfo{title}{Off-road obstacle avoidance through end-to-end
  learning}, in: \bibinfo{booktitle}{Advances in Neural Information Processing
  Systems (NIPS)}, pp. \bibinfo{pages}{739--746}.
\bibitem[{Naus et~al.(2010)Naus, Vugts, Ploeg, van~de Molengraft and
  Steinbuch}]{naus2010string}
\bibinfo{author}{Naus, G.J.}, \bibinfo{author}{Vugts, R.P.},
  \bibinfo{author}{Ploeg, J.}, \bibinfo{author}{van~de Molengraft, M.J.},
  \bibinfo{author}{Steinbuch, M.}, \bibinfo{year}{2010}.
\newblock \bibinfo{title}{String-stable cacc design and experimental
  validation: A frequency-domain approach}.
\newblock \bibinfo{journal}{IEEE Transactions on vehicular technology}
  \bibinfo{volume}{59}, \bibinfo{pages}{4268--4279}.
\bibitem[{NDS(2018)}]{SHRP2}
\bibinfo{author}{NDS, I.S.}, \bibinfo{year}{2018}.
\newblock \bibinfo{howpublished}{\url{insight.shrp2nds.us/home}}.
\newblock \bibinfo{note}{[Online; accessed 07.19.2018]}.
\bibitem[{Newell(1961)}]{newell1961nonlinear}
\bibinfo{author}{Newell, G.F.}, \bibinfo{year}{1961}.
\newblock \bibinfo{title}{Nonlinear effects in the dynamics of car following}.
\newblock \bibinfo{journal}{Operations research} \bibinfo{volume}{9},
  \bibinfo{pages}{209--229}.
\bibitem[{Ng and Russell(2000)}]{ng2000algorithms}
\bibinfo{author}{Ng, A.Y.}, \bibinfo{author}{Russell, S.J.},
  \bibinfo{year}{2000}.
\newblock \bibinfo{title}{Algorithms for inverse reinforcement learning.}, in:
  \bibinfo{booktitle}{Icml}, p.~\bibinfo{pages}{2}.
\bibitem[{Ngoduy(2013a)}]{ngoduy2013analytical}
\bibinfo{author}{Ngoduy, D.}, \bibinfo{year}{2013}a.
\newblock \bibinfo{title}{Analytical studies on the instabilities of
  heterogeneous intelligent traffic flow}.
\newblock \bibinfo{journal}{Communications in Nonlinear Science and Numerical
  Simulation} \bibinfo{volume}{18}, \bibinfo{pages}{2699--2706}.
\bibitem[{Ngoduy(2013b)}]{ngoduy2013instability}
\bibinfo{author}{Ngoduy, D.}, \bibinfo{year}{2013}b.
\newblock \bibinfo{title}{Instability of cooperative adaptive cruise control
  traffic flow: A macroscopic approach}.
\newblock \bibinfo{journal}{Communications in Nonlinear Science and Numerical
  Simulation} \bibinfo{volume}{18}, \bibinfo{pages}{2838--2851}.
\bibitem[{Ngoduy et~al.(2009)Ngoduy, Hoogendoorn and Liu}]{ngoduy2009continuum}
\bibinfo{author}{Ngoduy, D.}, \bibinfo{author}{Hoogendoorn, S.},
  \bibinfo{author}{Liu, R.}, \bibinfo{year}{2009}.
\newblock \bibinfo{title}{Continuum modeling of cooperative traffic flow
  dynamics}.
\newblock \bibinfo{journal}{Physica A: Statistical Mechanics and its
  Applications} \bibinfo{volume}{388}, \bibinfo{pages}{2705--2716}.
\bibitem[{Nguyen et~al.(2018)Nguyen, Nguyen and Nahavandi}]{Nguyen-2018}
\bibinfo{author}{Nguyen, T.}, \bibinfo{author}{Nguyen, N.},
  \bibinfo{author}{Nahavandi, S.}, \bibinfo{year}{2018}.
\newblock \bibinfo{title}{Deep reinforcement learning for multi-agent systems:
  a review of challenges}, in: \bibinfo{booktitle}{arXiv preprint
  arXiv:1812.11794}.
\bibitem[{Omidshafiei et~al.(2017)Omidshafiei, Pazis, Amato, How and
  Vian}]{Omidshafiei-2017}
\bibinfo{author}{Omidshafiei, S.}, \bibinfo{author}{Pazis, J.},
  \bibinfo{author}{Amato, C.}, \bibinfo{author}{How, J.},
  \bibinfo{author}{Vian, J.}, \bibinfo{year}{2017}.
\newblock \bibinfo{title}{{Deep decentralized multi-task multi-agent
  reinforcement learning under partial observability}}, in:
  \bibinfo{booktitle}{the 34th International Conference on Machine Learning},
  pp. \bibinfo{pages}{2681--2690}.
\bibitem[{OpenAI(2018)}]{OpenAI_dota}
\bibinfo{author}{OpenAI}, \bibinfo{year}{2018}.
\newblock \bibinfo{title}{Openai five}.
\newblock \bibinfo{howpublished}{\url{https://blog.openai.com/openai-five/}}.
\bibitem[{Orosz et~al.(2010)Orosz, Wilson and
  St{\'e}p{\'a}n}]{orosz2010traffic}
\bibinfo{author}{Orosz, G.}, \bibinfo{author}{Wilson, R.E.},
  \bibinfo{author}{St{\'e}p{\'a}n, G.}, \bibinfo{year}{2010}.
\newblock \bibinfo{title}{Traffic jams: dynamics and control}.
\bibitem[{Ossen and Hoogendoorn(2011)}]{ossen2011heterogeneity}
\bibinfo{author}{Ossen, S.}, \bibinfo{author}{Hoogendoorn, S.P.},
  \bibinfo{year}{2011}.
\newblock \bibinfo{title}{Heterogeneity in car-following behavior: Theory and
  empirics}.
\newblock \bibinfo{journal}{Transportation research part C: emerging
  technologies} \bibinfo{volume}{19}, \bibinfo{pages}{182--195}.
\bibitem[{Palmer et~al.(2018)Palmer, Tuyls, Bloembergen and
  Savani}]{Palmer-2018}
\bibinfo{author}{Palmer, G.}, \bibinfo{author}{Tuyls, K.},
  \bibinfo{author}{Bloembergen, D.}, \bibinfo{author}{Savani, R.},
  \bibinfo{year}{2018}.
\newblock \bibinfo{title}{Lenient multi-agent deep reinforcement learning}, in:
  \bibinfo{booktitle}{the 17th International Conference on Autonomous Agents
  and MultiAgent Systems}, pp. \bibinfo{pages}{443--451}.
\bibitem[{Pan et~al.(2018)Pan, Cheng, Saigol, Lee, Yan, Theodorou and
  Boots}]{Pan-2018}
\bibinfo{author}{Pan, Y.}, \bibinfo{author}{Cheng, C.A.},
  \bibinfo{author}{Saigol, K.}, \bibinfo{author}{Lee, K.},
  \bibinfo{author}{Yan, X.}, \bibinfo{author}{Theodorou, E.},
  \bibinfo{author}{Boots, B.}, \bibinfo{year}{2018}.
\newblock \bibinfo{title}{Agile autonomous driving using end-to-end deep
  imitation learning}, in: \bibinfo{booktitle}{Robotics: science and systems}.
\bibitem[{Panwai and Dia(2007)}]{ann}
\bibinfo{author}{Panwai, S.}, \bibinfo{author}{Dia, H.}, \bibinfo{year}{2007}.
\newblock \bibinfo{title}{Neural agent car-following models}.
\newblock \bibinfo{journal}{IEEE Transactions on Intelligent Transportation
  Systems} \bibinfo{volume}{8}, \bibinfo{pages}{60--70}.
\bibitem[{Patel et~al.(2016)Patel, Levin and Boyles}]{patel2016effects}
\bibinfo{author}{Patel, R.}, \bibinfo{author}{Levin, M.W.},
  \bibinfo{author}{Boyles, S.D.}, \bibinfo{year}{2016}.
\newblock \bibinfo{title}{Effects of autonomous vehicle behavior on arterial
  and freeway networks}.
\newblock \bibinfo{journal}{Transportation Research Record: Journal of the
  Transportation Research Board} .
\bibitem[{Patil et~al.(2019)Patil, Malla, Gang and Chen}]{Patil-2019}
\bibinfo{author}{Patil, A.}, \bibinfo{author}{Malla, S.},
  \bibinfo{author}{Gang, H.}, \bibinfo{author}{Chen, Y.T.},
  \bibinfo{year}{2019}.
\newblock \bibinfo{title}{The {H3D} dataset for full-surround {3D} multi-object
  detection and tracking in crowded urban scenes}, in: \bibinfo{booktitle}{2019
  International Conference on Robotics and Automation (ICRA)}, pp.
  \bibinfo{pages}{9552--9557}.
\bibitem[{Paxton et~al.(2017)Paxton, Raman, Hager and Kobilarov}]{Paxton-2017}
\bibinfo{author}{Paxton, C.}, \bibinfo{author}{Raman, V.},
  \bibinfo{author}{Hager, G.D.}, \bibinfo{author}{Kobilarov, M.},
  \bibinfo{year}{2017}.
\newblock \bibinfo{title}{Combining neural networks and tree search for task
  and motion planning in challenging environments}, in:
  \bibinfo{booktitle}{2017 IEEE/RSJ International Conference on Intelligent
  Robots and Systems (IROS)}, pp. \bibinfo{pages}{6059--6066}.
\bibitem[{Payne(1971)}]{payne1971model}
\bibinfo{author}{Payne, H.J.}, \bibinfo{year}{1971}.
\newblock \bibinfo{title}{Model of freeway traffic and control}.
\newblock \bibinfo{journal}{Mathematical Model of Public System} ,
  \bibinfo{pages}{51--61}.
\bibitem[{Pedersen(2001)}]{pedersen2001game}
\bibinfo{author}{Pedersen, P.A.}, \bibinfo{year}{2001}.
\newblock \bibinfo{title}{A game theoretical approach to road safety}.
\newblock \bibinfo{type}{Technical Report}. Department of Economics Discussion
  Paper, University of Kent.
\bibitem[{Pedersen(2003)}]{pedersen2003moral}
\bibinfo{author}{Pedersen, P.A.}, \bibinfo{year}{2003}.
\newblock \bibinfo{title}{Moral hazard in traffic games}.
\newblock \bibinfo{journal}{Journal of Transport Economics and Policy (JTEP)}
  \bibinfo{volume}{37}, \bibinfo{pages}{47--68}.
\bibitem[{Perot et~al.(2017)Perot, Jaritz, Toromanoff and
  de~Charette}]{Perot-2017}
\bibinfo{author}{Perot, E.}, \bibinfo{author}{Jaritz, M.},
  \bibinfo{author}{Toromanoff, M.}, \bibinfo{author}{de~Charette, R.},
  \bibinfo{year}{2017}.
\newblock \bibinfo{title}{End-to-end driving in a realistic racing game with
  deep reinforcement learning}, in: \bibinfo{booktitle}{2017 IEEE Conference on
  Computer Vision and Pattern Recognition Workshops (CVPRW)}, pp.
  \bibinfo{pages}{474--475}.
\bibitem[{Pham et~al.(2019)Pham, Sevestre, Pahwa, Zhan, Pang, Chen, Mustafa,
  Chandrasekhar and Lin}]{Pham-2019}
\bibinfo{author}{Pham, Q.H.}, \bibinfo{author}{Sevestre, P.},
  \bibinfo{author}{Pahwa, R.S.}, \bibinfo{author}{Zhan, H.},
  \bibinfo{author}{Pang, C.H.}, \bibinfo{author}{Chen, Y.},
  \bibinfo{author}{Mustafa, A.}, \bibinfo{author}{Chandrasekhar, V.},
  \bibinfo{author}{Lin, J.}, \bibinfo{year}{2019}.
\newblock \bibinfo{title}{{A*3D} {D}ataset: Towards autonomous driving in
  challenging environments}.
\newblock \bibinfo{journal}{arXiv preprint arXiv:1909.07541} .
\bibitem[{Phegley et~al.(2014)Phegley, Gomes and
  Horowitz}]{macro_stochastic_phegley2014fundamental}
\bibinfo{author}{Phegley, B.}, \bibinfo{author}{Gomes, G.},
  \bibinfo{author}{Horowitz, R.}, \bibinfo{year}{2014}.
\newblock \bibinfo{title}{Fundamental diagram calibration: A stochastic
  approach to linear fitting}, in: \bibinfo{booktitle}{Proceedings of the 93rd
  Annual Meeting of the Transportation Research Board, Washington, DC}.
\bibitem[{Ploeg et~al.(2011)Ploeg, Scheepers, Van~Nunen, Van~de Wouw and
  Nijmeijer}]{ploeg2011design}
\bibinfo{author}{Ploeg, J.}, \bibinfo{author}{Scheepers, B.T.},
  \bibinfo{author}{Van~Nunen, E.}, \bibinfo{author}{Van~de Wouw, N.},
  \bibinfo{author}{Nijmeijer, H.}, \bibinfo{year}{2011}.
\newblock \bibinfo{title}{Design and experimental evaluation of cooperative
  adaptive cruise control}, in: \bibinfo{booktitle}{Intelligent Transportation
  Systems (ITSC), 2011 14th International IEEE Conference on},
  \bibinfo{organization}{IEEE}. pp. \bibinfo{pages}{260--265}.
\bibitem[{Pomerleau(1989)}]{Pomerleau-1989}
\bibinfo{author}{Pomerleau, D.A.}, \bibinfo{year}{1989}.
\newblock \bibinfo{title}{{ALVINN}: An autonomous land vehicle in a neural
  network}, in: \bibinfo{booktitle}{Advances in Neural Information Processing
  Systems (NIPS)}, pp. \bibinfo{pages}{305--313}.
\bibitem[{Puterman(1994)}]{puterman_markov_1994}
\bibinfo{author}{Puterman, M.L.}, \bibinfo{year}{1994}.
\newblock \bibinfo{title}{Markov {Decision} {Processes}: {Discrete}
  {Stochastic} {Dynamic} {Programming}}.
\newblock \bibinfo{edition}{1st} ed., \bibinfo{publisher}{John Wiley \& Sons,
  Inc.}, \bibinfo{address}{New York, NY, USA}.
\bibitem[{Qin and Orosz(2013)}]{qin2013digital}
\bibinfo{author}{Qin, W.B.}, \bibinfo{author}{Orosz, G.}, \bibinfo{year}{2013}.
\newblock \bibinfo{title}{Digital effects and delays in connected vehicles:
  linear stability and simulations}, in: \bibinfo{booktitle}{ASME 2013 Dynamic
  Systems and Control Conference}, \bibinfo{organization}{American Society of
  Mechanical Engineers}. pp. \bibinfo{pages}{V002T30A001--V002T30A001}.
\bibitem[{Qin and Orosz(2017)}]{qin2017scalable}
\bibinfo{author}{Qin, W.B.}, \bibinfo{author}{Orosz, G.}, \bibinfo{year}{2017}.
\newblock \bibinfo{title}{Scalable stability analysis on large connected
  vehicle systems subject to stochastic communication delays}.
\newblock \bibinfo{journal}{Transportation Research Part C: Emerging
  Technologies} \bibinfo{volume}{83}, \bibinfo{pages}{39--60}.
\bibitem[{Qu et~al.(2015)Qu, Wang and Zhang}]{macro_lsm_qu2015fundamental}
\bibinfo{author}{Qu, X.}, \bibinfo{author}{Wang, S.}, \bibinfo{author}{Zhang,
  J.}, \bibinfo{year}{2015}.
\newblock \bibinfo{title}{On the fundamental diagram for freeway traffic: a
  novel calibration approach for single-regime models}.
\newblock \bibinfo{journal}{Transportation Research Part B: Methodological}
  \bibinfo{volume}{73}, \bibinfo{pages}{91--102}.
\bibitem[{Rahman et~al.(2015)Rahman, Chowdhury, Khan and Bhavsar}]{bayes}
\bibinfo{author}{Rahman, M.}, \bibinfo{author}{Chowdhury, M.},
  \bibinfo{author}{Khan, T.}, \bibinfo{author}{Bhavsar, P.},
  \bibinfo{year}{2015}.
\newblock \bibinfo{title}{Improving the efficacy of car-following models with a
  new stochastic parameter estimation and calibration method}.
\newblock \bibinfo{journal}{IEEE Transactions on Intelligent Transportation
  Systems} \bibinfo{volume}{16}, \bibinfo{pages}{2687--2699}.
\bibitem[{Rahmati and Talebpour(2017)}]{rahmati2017towards}
\bibinfo{author}{Rahmati, Y.}, \bibinfo{author}{Talebpour, A.},
  \bibinfo{year}{2017}.
\newblock \bibinfo{title}{Towards a collaborative connected, automated driving
  environment: A game theory based decision framework for unprotected left turn
  maneuvers}, in: \bibinfo{booktitle}{2017 IEEE Intelligent Vehicles Symposium
  (IV)}, \bibinfo{organization}{IEEE}. pp. \bibinfo{pages}{1316--1321}.
\bibitem[{Rahwan et~al.(2019)Rahwan, Cebrian, Obradovich, Bongard, Bonnefon,
  Breazeal, Crandall, Christakis, Couzin, Jackson et~al.}]{rahwan2019machine}
\bibinfo{author}{Rahwan, I.}, \bibinfo{author}{Cebrian, M.},
  \bibinfo{author}{Obradovich, N.}, \bibinfo{author}{Bongard, J.},
  \bibinfo{author}{Bonnefon, J.F.}, \bibinfo{author}{Breazeal, C.},
  \bibinfo{author}{Crandall, J.W.}, \bibinfo{author}{Christakis, N.A.},
  \bibinfo{author}{Couzin, I.D.}, \bibinfo{author}{Jackson, M.O.}, et~al.,
  \bibinfo{year}{2019}.
\newblock \bibinfo{title}{Machine behaviour}.
\newblock \bibinfo{journal}{Nature} \bibinfo{volume}{568},
  \bibinfo{pages}{477--486}.
\bibitem[{Raissi(2018)}]{Raissi-2018a}
\bibinfo{author}{Raissi, M.}, \bibinfo{year}{2018}.
\newblock \bibinfo{title}{Deep hidden physics models: Deep learning of
  nonlinear partial differential equations}.
\newblock \bibinfo{journal}{Journal of Machine Learning Research}
  \bibinfo{volume}{19}, \bibinfo{pages}{932--955}.
\bibitem[{Raissi and Karniadakis(2018)}]{Raissi-2018b}
\bibinfo{author}{Raissi, M.}, \bibinfo{author}{Karniadakis, G.E.},
  \bibinfo{year}{2018}.
\newblock \bibinfo{title}{Hidden physics models: Machine learning of nonlinear
  partial differential equations}.
\newblock \bibinfo{journal}{Journal of Computational Physics}
  \bibinfo{volume}{357}, \bibinfo{pages}{125--141}.
\bibitem[{Raissi et~al.(2019)Raissi, Wang, Triantafyllou and
  Karniadakis}]{Maziar-2019}
\bibinfo{author}{Raissi, M.}, \bibinfo{author}{Wang, Z.},
  \bibinfo{author}{Triantafyllou, M.S.}, \bibinfo{author}{Karniadakis, G.E.},
  \bibinfo{year}{2019}.
\newblock \bibinfo{title}{Deep learning of vortex-induced vibrations}.
\newblock \bibinfo{journal}{Journal of Fluid Mechanics} \bibinfo{volume}{861},
  \bibinfo{pages}{119--137}.
\bibitem[{Raissi et~al.(2020)Raissi, Yazdani and Karniadakis}]{Maziar-2020}
\bibinfo{author}{Raissi, M.}, \bibinfo{author}{Yazdani, A.},
  \bibinfo{author}{Karniadakis, G.E.}, \bibinfo{year}{2020}.
\newblock \bibinfo{title}{Hidden fluid mechanics: Learning velocity and
  pressure fields from flow visualizations}.
\newblock \bibinfo{journal}{Science} \bibinfo{volume}{367},
  \bibinfo{pages}{1026--1030}.
\bibitem[{Rajamani and Shladover(2001)}]{rajamani2001experimental}
\bibinfo{author}{Rajamani, R.}, \bibinfo{author}{Shladover, S.E.},
  \bibinfo{year}{2001}.
\newblock \bibinfo{title}{An experimental comparative study of autonomous and
  co-operative vehicle-follower control systems}.
\newblock \bibinfo{journal}{Transportation Research Part C: Emerging
  Technologies} \bibinfo{volume}{9}, \bibinfo{pages}{15--31}.
\bibitem[{Rakha and Crowther(2002)}]{loop_detector_rakha2002comparison}
\bibinfo{author}{Rakha, H.}, \bibinfo{author}{Crowther, B.},
  \bibinfo{year}{2002}.
\newblock \bibinfo{title}{Comparison of greenshields, pipes, and van aerde
  car-following and traffic stream models}.
\newblock \bibinfo{journal}{Transportation Research Record}
  \bibinfo{volume}{1802}, \bibinfo{pages}{248--262}.
\bibitem[{Rakha and Crowther(2003)}]{loop_detector_rakha2003comparison}
\bibinfo{author}{Rakha, H.}, \bibinfo{author}{Crowther, B.},
  \bibinfo{year}{2003}.
\newblock \bibinfo{title}{Comparison and calibration of fresim and integration
  steady-state car-following behavior}.
\newblock \bibinfo{journal}{Transportation Research Part A: Policy and
  Practice} \bibinfo{volume}{37}, \bibinfo{pages}{1--27}.
\bibitem[{Rakha et~al.(2010)Rakha, Gao
  et~al.}]{loop_detector_rakha2010calibration}
\bibinfo{author}{Rakha, H.A.}, \bibinfo{author}{Gao, Y.}, et~al.,
  \bibinfo{year}{2010}.
\newblock \bibinfo{title}{Calibration of steady-state car-following models
  using macroscopic loop detector data} .
\bibitem[{Ramamoorthy and Yampolskiy(2018)}]{ramamoorthy2018beyond}
\bibinfo{author}{Ramamoorthy, A.}, \bibinfo{author}{Yampolskiy, R.},
  \bibinfo{year}{2018}.
\newblock \bibinfo{title}{Beyond mad? the race for artificial general
  intelligence}.
\newblock \bibinfo{journal}{ITU J} \bibinfo{volume}{1}, \bibinfo{pages}{1--8}.
\bibitem[{Rausch et~al.(2017)Rausch, Hansen, Solowjow, Liu, Kreuzer and
  Hedrick}]{Rausch-2017}
\bibinfo{author}{Rausch, V.}, \bibinfo{author}{Hansen, A.},
  \bibinfo{author}{Solowjow, E.}, \bibinfo{author}{Liu, C.},
  \bibinfo{author}{Kreuzer, E.}, \bibinfo{author}{Hedrick, J.K.},
  \bibinfo{year}{2017}.
\newblock \bibinfo{title}{Learning a deep neural net policy for end-to-end
  control of autonomous vehicles}, in: \bibinfo{booktitle}{American Control
  Conference (ACC)}, pp. \bibinfo{pages}{4914--4919}.
\bibitem[{Richards(1956)}]{richards1956shock}
\bibinfo{author}{Richards, P.I.}, \bibinfo{year}{1956}.
\newblock \bibinfo{title}{Shock waves on the highway}.
\newblock \bibinfo{journal}{Operations research} \bibinfo{volume}{4},
  \bibinfo{pages}{42--51}.
\bibitem[{Richter et~al.(2017)Richter, Hayder and Koltun}]{richter-2017}
\bibinfo{author}{Richter, S.R.}, \bibinfo{author}{Hayder, Z.},
  \bibinfo{author}{Koltun, V.}, \bibinfo{year}{2017}.
\newblock \bibinfo{title}{Playing for benchmarks}, in: \bibinfo{booktitle}{IEEE
  International Conference on Computer Vision}, pp.
  \bibinfo{pages}{2213--2222}.
\bibitem[{Richter et~al.(2016)Richter, Vineet, Roth and Koltun}]{Richter-2016}
\bibinfo{author}{Richter, S.R.}, \bibinfo{author}{Vineet, V.},
  \bibinfo{author}{Roth, S.}, \bibinfo{author}{Koltun, V.},
  \bibinfo{year}{2016}.
\newblock \bibinfo{title}{Playing for data: Ground truth from computer games},
  in: \bibinfo{booktitle}{European Conference on Computer Vision}, pp.
  \bibinfo{pages}{102--118}.
\bibitem[{Rios-Torres and Malikopoulos(2016)}]{rios2016survey}
\bibinfo{author}{Rios-Torres, J.}, \bibinfo{author}{Malikopoulos, A.A.},
  \bibinfo{year}{2016}.
\newblock \bibinfo{title}{A survey on the coordination of connected and
  automated vehicles at intersections and merging at highway on-ramps}.
\newblock \bibinfo{journal}{IEEE Transactions on Intelligent Transportation
  Systems} \bibinfo{volume}{18}, \bibinfo{pages}{1066--1077}.
\bibitem[{Ross and Bagnell(2010)}]{Ross-2010}
\bibinfo{author}{Ross, S.}, \bibinfo{author}{Bagnell, D.},
  \bibinfo{year}{2010}.
\newblock \bibinfo{title}{Efficient reductions for imitation learning}, in:
  \bibinfo{booktitle}{Proceedings of the thirteenth International Conference on
  Artificial Intelligence and Statistics (AISTATS)}, pp.
  \bibinfo{pages}{661--668}.
\bibitem[{Ross et~al.(2011)Ross, Gordon and Bagnell}]{ross2011reduction}
\bibinfo{author}{Ross, S.}, \bibinfo{author}{Gordon, G.},
  \bibinfo{author}{Bagnell, D.}, \bibinfo{year}{2011}.
\newblock \bibinfo{title}{A reduction of imitation learning and structured
  prediction to no-regret online learning}, in: \bibinfo{booktitle}{Proceedings
  of the fourteenth international conference on artificial intelligence and
  statistics}, pp. \bibinfo{pages}{627--635}.
\bibitem[{Sadigh et~al.(2016a)Sadigh, Sastry, Seshia and Dragan}]{sadigh-2016}
\bibinfo{author}{Sadigh, D.}, \bibinfo{author}{Sastry, S.},
  \bibinfo{author}{Seshia, S.A.}, \bibinfo{author}{Dragan, A.D.},
  \bibinfo{year}{2016}a.
\newblock \bibinfo{title}{High-speed highway scene prediction based on driver
  models learned from demonstrations}.
\newblock \bibinfo{journal}{Robotics: Science and Systems} \bibinfo{volume}{2}.
\bibitem[{Sadigh et~al.(2016b)Sadigh, Sastry, Seshia and
  Dragan}]{sadigh2016planning}
\bibinfo{author}{Sadigh, D.}, \bibinfo{author}{Sastry, S.},
  \bibinfo{author}{Seshia, S.A.}, \bibinfo{author}{Dragan, A.D.},
  \bibinfo{year}{2016}b.
\newblock \bibinfo{title}{Planning for autonomous cars that leverage effects on
  human actions.}, in: \bibinfo{booktitle}{Robotics: Science and Systems}.
\bibitem[{Sadigh et~al.()Sadigh, Sastry, Seshia and Berkeley}]{sadighverifying}
\bibinfo{author}{Sadigh, D.}, \bibinfo{author}{Sastry, S.S.},
  \bibinfo{author}{Seshia, S.A.}, \bibinfo{author}{Berkeley, U.}, .
\newblock \bibinfo{title}{Verifying robustness of human-aware autonomous cars}
  .
\bibitem[{Sallab et~al.(2017)Sallab, Abdou, Perot and Yogamani}]{Sallab-2017}
\bibinfo{author}{Sallab, A.E.}, \bibinfo{author}{Abdou, M.},
  \bibinfo{author}{Perot, E.}, \bibinfo{author}{Yogamani, S.},
  \bibinfo{year}{2017}.
\newblock \bibinfo{title}{Deep reinforcement learning framework for autonomous
  driving}, in: \bibinfo{booktitle}{arXiv preprint arXiv:1704.02532}.
\bibitem[{SAS(2015)}]{cv_sas}
\bibinfo{author}{SAS}, \bibinfo{year}{2015}.
\newblock \bibinfo{title}{The connected vehicle: Big data, big opportunities}.
\newblock
  \bibinfo{howpublished}{\url{https://www.sas.com/content/dam/SAS/en_us/doc/whitepaper1/connected-vehicle-107832.pdf}}.
\newblock \bibinfo{note}{[Online; accessed 06.22.2020]}.
\bibitem[{Schakel et~al.(2010)Schakel, Van~Arem and
  Netten}]{schakel2010effects}
\bibinfo{author}{Schakel, W.J.}, \bibinfo{author}{Van~Arem, B.},
  \bibinfo{author}{Netten, B.D.}, \bibinfo{year}{2010}.
\newblock \bibinfo{title}{Effects of cooperative adaptive cruise control on
  traffic flow stability}, in: \bibinfo{booktitle}{Intelligent Transportation
  Systems (ITSC), 2010 13th International IEEE Conference on},
  \bibinfo{organization}{IEEE}. pp. \bibinfo{pages}{759--764}.
\bibitem[{Schulman et~al.(2015a)Schulman, Levine, Abbeel, Jordan and
  Moritz}]{Schulman-2015}
\bibinfo{author}{Schulman, J.}, \bibinfo{author}{Levine, S.},
  \bibinfo{author}{Abbeel, P.}, \bibinfo{author}{Jordan, M.},
  \bibinfo{author}{Moritz, P.}, \bibinfo{year}{2015}a.
\newblock \bibinfo{title}{Trust region policy optimization}, in:
  \bibinfo{booktitle}{International Conference on Machine Learning (ICML)}, pp.
  \bibinfo{pages}{1889--1897}.
\bibitem[{Schulman et~al.(2015b)Schulman, Moritz, Levine, Jordan and
  Abbeel}]{Schulman2-2015}
\bibinfo{author}{Schulman, J.}, \bibinfo{author}{Moritz, P.},
  \bibinfo{author}{Levine, S.}, \bibinfo{author}{Jordan, M.},
  \bibinfo{author}{Abbeel, P.}, \bibinfo{year}{2015}b.
\newblock \bibinfo{title}{High-dimensional continuous control using generalized
  advantage estimation}, in: \bibinfo{booktitle}{arXiv preprint
  arXiv:1506.02438}.
\bibitem[{Schwarting et~al.(2019)Schwarting, Pierson, Alonso-Mora, Karaman and
  Rus}]{schwarting2019social}
\bibinfo{author}{Schwarting, W.}, \bibinfo{author}{Pierson, A.},
  \bibinfo{author}{Alonso-Mora, J.}, \bibinfo{author}{Karaman, S.},
  \bibinfo{author}{Rus, D.}, \bibinfo{year}{2019}.
\newblock \bibinfo{title}{Social behavior for autonomous vehicles}.
\newblock \bibinfo{journal}{Proceedings of the National Academy of Sciences}
  \bibinfo{volume}{116}, \bibinfo{pages}{24972--24978}.
\bibitem[{Shladover et~al.(2015)Shladover, Nowakowski, Lu and
  Ferlis}]{shladover2015cooperative}
\bibinfo{author}{Shladover, S.E.}, \bibinfo{author}{Nowakowski, C.},
  \bibinfo{author}{Lu, X.Y.}, \bibinfo{author}{Ferlis, R.},
  \bibinfo{year}{2015}.
\newblock \bibinfo{title}{Cooperative adaptive cruise control: Definitions and
  operating concepts}.
\newblock \bibinfo{journal}{Transportation Research Record: Journal of the
  Transportation Research Board} , \bibinfo{pages}{145--152}.
\bibitem[{Shou et~al.(2020)Shou, Wang, Han, Liu, Tiwari and Di}]{shou2020long}
\bibinfo{author}{Shou, Z.}, \bibinfo{author}{Wang, Z.}, \bibinfo{author}{Han,
  K.}, \bibinfo{author}{Liu, Y.}, \bibinfo{author}{Tiwari, P.},
  \bibinfo{author}{Di, X.}, \bibinfo{year}{2020}.
\newblock \bibinfo{title}{Long-term prediction of lane change maneuver through
  a multilayer perceptron}.
\newblock \bibinfo{journal}{2020 IEEE Intelligent Vehicles Symposium (IEEE IV
  2020)} .
\bibitem[{Silver et~al.(2016)Silver, Huang, Maddison, Guez, Sifre, Van
  Den~Driessche, Schrittwieser, Antonoglou, Panneershelvam, Lanctot, Dieleman,
  Grewe, Nham, Kalchbrenner, Sutskever, Lillicrap, Leach, Kavukcuoglu, Graepel
  and Hassabis}]{Silver-2016}
\bibinfo{author}{Silver, D.}, \bibinfo{author}{Huang, A.},
  \bibinfo{author}{Maddison, C.J.}, \bibinfo{author}{Guez, A.},
  \bibinfo{author}{Sifre, L.}, \bibinfo{author}{Van Den~Driessche, G.},
  \bibinfo{author}{Schrittwieser, J.}, \bibinfo{author}{Antonoglou, I.},
  \bibinfo{author}{Panneershelvam, V.}, \bibinfo{author}{Lanctot, M.},
  \bibinfo{author}{Dieleman, S.}, \bibinfo{author}{Grewe, D.},
  \bibinfo{author}{Nham, J.}, \bibinfo{author}{Kalchbrenner, N.},
  \bibinfo{author}{Sutskever, I.}, \bibinfo{author}{Lillicrap, T.},
  \bibinfo{author}{Leach, M.}, \bibinfo{author}{Kavukcuoglu, K.},
  \bibinfo{author}{Graepel, T.}, \bibinfo{author}{Hassabis, D.},
  \bibinfo{year}{2016}.
\newblock \bibinfo{title}{Mastering the game of go with deep neural networks
  and tree search}.
\newblock \bibinfo{journal}{nature} \bibinfo{volume}{529},
  \bibinfo{pages}{484}.
\bibitem[{Singer et~al.(2013)Singer, Atkinson, Myers, Robinson, Krueger
  et~al.}]{singer2013travel}
\bibinfo{author}{Singer, J.P.}, \bibinfo{author}{Atkinson, J.},
  \bibinfo{author}{Myers, M.}, \bibinfo{author}{Robinson, A.E.},
  \bibinfo{author}{Krueger, J.}, et~al., \bibinfo{year}{2013}.
\newblock \bibinfo{title}{Travel time on arterials and rural highways:
  state-of-the-practice synthesis on rural data collection technology.}
\newblock \bibinfo{type}{Technical Report}. United States. Federal Highway
  Administration.
\bibitem[{Sivaraman and Trivedi(2013)}]{sivaraman2013looking}
\bibinfo{author}{Sivaraman, S.}, \bibinfo{author}{Trivedi, M.M.},
  \bibinfo{year}{2013}.
\newblock \bibinfo{title}{Looking at vehicles on the road: A survey of
  vision-based vehicle detection, tracking, and behavior analysis}.
\newblock \bibinfo{journal}{IEEE transactions on intelligent transportation
  systems} \bibinfo{volume}{14}, \bibinfo{pages}{1773--1795}.
\bibitem[{Song et~al.(2018)Song, Ren, Sadigh and Ermon}]{song2018multi}
\bibinfo{author}{Song, J.}, \bibinfo{author}{Ren, H.}, \bibinfo{author}{Sadigh,
  D.}, \bibinfo{author}{Ermon, S.}, \bibinfo{year}{2018}.
\newblock \bibinfo{title}{Multi-agent generative adversarial imitation
  learning} .
\bibitem[{Song et~al.(2019)Song, Wang, Zhou, Zhu, Guan, Dai, Su, Li and
  Yang}]{Xibin-2019}
\bibinfo{author}{Song, X.}, \bibinfo{author}{Wang, P.}, \bibinfo{author}{Zhou,
  D.}, \bibinfo{author}{Zhu, R.}, \bibinfo{author}{Guan, C.},
  \bibinfo{author}{Dai, Y.}, \bibinfo{author}{Su, H.}, \bibinfo{author}{Li,
  H.}, \bibinfo{author}{Yang, R.}, \bibinfo{year}{2019}.
\newblock \bibinfo{title}{Apollocar3{D}: A large 3{D} car instance
  understanding benchmark for autonomous driving}, in:
  \bibinfo{booktitle}{Proceedings of the IEEE Conference on Computer Vision and
  Pattern Recognition (CVPR)}, pp. \bibinfo{pages}{5452--5462}.
\bibitem[{Stern et~al.(2018)Stern, Cui, Delle~Monache, Bhadani, Bunting,
  Churchill, Hamilton, Pohlmann, Wu, Piccoli et~al.}]{stern2018dissipation}
\bibinfo{author}{Stern, R.E.}, \bibinfo{author}{Cui, S.},
  \bibinfo{author}{Delle~Monache, M.L.}, \bibinfo{author}{Bhadani, R.},
  \bibinfo{author}{Bunting, M.}, \bibinfo{author}{Churchill, M.},
  \bibinfo{author}{Hamilton, N.}, \bibinfo{author}{Pohlmann, H.},
  \bibinfo{author}{Wu, F.}, \bibinfo{author}{Piccoli, B.}, et~al.,
  \bibinfo{year}{2018}.
\newblock \bibinfo{title}{Dissipation of stop-and-go waves via control of
  autonomous vehicles: Field experiments}.
\newblock \bibinfo{journal}{Transportation Research Part C: Emerging
  Technologies} \bibinfo{volume}{89}, \bibinfo{pages}{205--221}.
\bibitem[{Sunberg et~al.(2017)Sunberg, Ho and Kochenderfer}]{Sunberg-2017}
\bibinfo{author}{Sunberg, Z.N.}, \bibinfo{author}{Ho, C.J.},
  \bibinfo{author}{Kochenderfer, M.J.}, \bibinfo{year}{2017}.
\newblock \bibinfo{title}{The value of inferring the internal state of traffic
  participants for autonomous freeway driving}, in: \bibinfo{booktitle}{2017
  American Control Conference (ACC)}, pp. \bibinfo{pages}{3004--301}.
\bibitem[{Sutton and Barto(1998)}]{sutton_introduction_1998}
\bibinfo{author}{Sutton, R.S.}, \bibinfo{author}{Barto, A.G.},
  \bibinfo{year}{1998}.
\newblock \bibinfo{title}{Introduction to {Reinforcement} {Learning}}.
\newblock \bibinfo{edition}{1st} ed., \bibinfo{publisher}{MIT Press},
  \bibinfo{address}{Cambridge, MA, USA}.
\bibitem[{Swaroop and Hedrick(1996)}]{swaroop1996string}
\bibinfo{author}{Swaroop, D.}, \bibinfo{author}{Hedrick, J.K.},
  \bibinfo{year}{1996}.
\newblock \bibinfo{title}{String stability of interconnected systems}.
\newblock \bibinfo{journal}{IEEE transactions on automatic control}
  \bibinfo{volume}{41}, \bibinfo{pages}{349--357}.
\bibitem[{Swaroop et~al.(1994)Swaroop, Hedrick, Chien and
  Ioannou}]{swaroop1994comparision}
\bibinfo{author}{Swaroop, D.}, \bibinfo{author}{Hedrick, J.K.},
  \bibinfo{author}{Chien, C.}, \bibinfo{author}{Ioannou, P.},
  \bibinfo{year}{1994}.
\newblock \bibinfo{title}{A comparision of spacing and headway control laws for
  automatically controlled vehicles1}.
\newblock \bibinfo{journal}{Vehicle system dynamics} \bibinfo{volume}{23},
  \bibinfo{pages}{597--625}.
\bibitem[{Swaroop et~al.(2001)Swaroop, Hedrick and Choi}]{swaroop2001direct}
\bibinfo{author}{Swaroop, D.}, \bibinfo{author}{Hedrick, J.K.},
  \bibinfo{author}{Choi, S.B.}, \bibinfo{year}{2001}.
\newblock \bibinfo{title}{Direct adaptive longitudinal control of vehicle
  platoons}.
\newblock \bibinfo{journal}{IEEE Transactions on Vehicular Technology}
  \bibinfo{volume}{50}, \bibinfo{pages}{150--161}.
\bibitem[{Syed and Schapire(2008)}]{Syed-2008}
\bibinfo{author}{Syed, U.}, \bibinfo{author}{Schapire, R.E.},
  \bibinfo{year}{2008}.
\newblock \bibinfo{title}{A game-theoretic approach to apprenticeship
  learning}, in: \bibinfo{booktitle}{Advances in Neural Information Processing
  Systems (NIPS)}, pp. \bibinfo{pages}{1449--1456}.
\bibitem[{DLR {I}nstitute of~{T}ransportation {S}ystems()}]{SUMO}
\bibinfo{author}{DLR {I}nstitute of~{T}ransportation {S}ystems, year =~{2018},
  t..S.h..{\url{https://www.dlr.de/ts/en/desktopdefault.aspx/tabid-9883/16931_read-41000/}}.n..O.},
  .
\bibitem[{Talebpour and Mahmassani(2016)}]{talebpour2016influence}
\bibinfo{author}{Talebpour, A.}, \bibinfo{author}{Mahmassani, H.S.},
  \bibinfo{year}{2016}.
\newblock \bibinfo{title}{Influence of connected and autonomous vehicles on
  traffic flow stability and throughput}.
\newblock \bibinfo{journal}{Transportation Research Part C: Emerging
  Technologies} \bibinfo{volume}{71}, \bibinfo{pages}{143--163}.
\bibitem[{Talebpour et~al.(2015)Talebpour, Mahmassani and
  Hamdar}]{Talebpour2015}
\bibinfo{author}{Talebpour, A.}, \bibinfo{author}{Mahmassani, H.S.},
  \bibinfo{author}{Hamdar, S.H.}, \bibinfo{year}{2015}.
\newblock \bibinfo{title}{{Modeling Lane-Changing Behavior in a Connected
  Environment: A Game Theory Approach}}.
\newblock \bibinfo{journal}{Transportation Research Procedia}
  \bibinfo{volume}{7}, \bibinfo{pages}{420--440}.
\newblock \DOIprefix\doi{10.1016/j.trpro.2015.06.022}.
\bibitem[{Tampuu et~al.(2017)Tampuu, Matiisen, Kodelja, Kuzovkin, Korjus, Aru,
  Aru and Vicente}]{tampuu_multiagent_2017}
\bibinfo{author}{Tampuu, A.}, \bibinfo{author}{Matiisen, T.},
  \bibinfo{author}{Kodelja, D.}, \bibinfo{author}{Kuzovkin, I.},
  \bibinfo{author}{Korjus, K.}, \bibinfo{author}{Aru, J.},
  \bibinfo{author}{Aru, J.}, \bibinfo{author}{Vicente, R.},
  \bibinfo{year}{2017}.
\newblock \bibinfo{title}{Multiagent cooperation and competition with deep
  reinforcement learning}.
\newblock \bibinfo{journal}{PLOS ONE} \bibinfo{volume}{12},
  \bibinfo{pages}{e0172395}.
\bibitem[{Tanaka(2013)}]{NN_moment_tanaka2013development}
\bibinfo{author}{Tanaka, M.}, \bibinfo{year}{2013}.
\newblock \bibinfo{title}{Development of various artificial neural network
  car-following models with converted data sets by a self-organization neural
  network}.
\newblock \bibinfo{journal}{Journal of the Eastern Asia Society for
  Transportation Studies} \bibinfo{volume}{10}, \bibinfo{pages}{1614--1630}.
\bibitem[{Tang et~al.(2017)Tang, Zhang, Gu, Li and
  Yang}]{camera_tang2017vehicle}
\bibinfo{author}{Tang, Y.}, \bibinfo{author}{Zhang, C.}, \bibinfo{author}{Gu,
  R.}, \bibinfo{author}{Li, P.}, \bibinfo{author}{Yang, B.},
  \bibinfo{year}{2017}.
\newblock \bibinfo{title}{Vehicle detection and recognition for intelligent
  traffic surveillance system}.
\newblock \bibinfo{journal}{Multimedia tools and applications}
  \bibinfo{volume}{76}, \bibinfo{pages}{5817--5832}.
\bibitem[{Tennenholtz(2002)}]{tennenholtz2002game}
\bibinfo{author}{Tennenholtz, M.}, \bibinfo{year}{2002}.
\newblock \bibinfo{title}{Game theory and artificial intelligence}, in:
  \bibinfo{booktitle}{Foundations and applications of multi-agent systems}.
  \bibinfo{publisher}{Springer}, pp. \bibinfo{pages}{49--58}.
\bibitem[{Tian et~al.(2019)Tian, Li, Kolmanovsky, Yildiz and
  Girard}]{tian2019game}
\bibinfo{author}{Tian, R.}, \bibinfo{author}{Li, N.},
  \bibinfo{author}{Kolmanovsky, I.}, \bibinfo{author}{Yildiz, Y.},
  \bibinfo{author}{Girard, A.}, \bibinfo{year}{2019}.
\newblock \bibinfo{title}{Game-theoretic modeling of traffic in unsignalized
  intersection network for autonomous vehicle control verification and
  validation}.
\newblock \bibinfo{journal}{arXiv preprint arXiv:1910.07141} .
\bibitem[{Tian et~al.(2018)Tian, Li, Li, Kolmanovsky, Girard and
  Yildiz}]{tian2018adaptive}
\bibinfo{author}{Tian, R.}, \bibinfo{author}{Li, S.}, \bibinfo{author}{Li, N.},
  \bibinfo{author}{Kolmanovsky, I.}, \bibinfo{author}{Girard, A.},
  \bibinfo{author}{Yildiz, Y.}, \bibinfo{year}{2018}.
\newblock \bibinfo{title}{Adaptive game-theoretic decision making for
  autonomous vehicle control at roundabouts}, in: \bibinfo{booktitle}{2018 IEEE
  Conference on Decision and Control (CDC)}, \bibinfo{organization}{IEEE}. pp.
  \bibinfo{pages}{321--326}.
\bibitem[{Tilbury et~al.(2020)Tilbury, Yang, Pradhan, Robert
  et~al.}]{tilbury2020analysis}
\bibinfo{author}{Tilbury, D.}, \bibinfo{author}{Yang, J.},
  \bibinfo{author}{Pradhan, A.}, \bibinfo{author}{Robert, L.}, et~al.,
  \bibinfo{year}{2020}.
\newblock \bibinfo{title}{Analysis and prediction of pedestrian crosswalk
  behavior during automated vehicle interactions}, in:
  \bibinfo{booktitle}{Proceedings of the International Conference on Robotics
  and Automation}.
\bibitem[{Treiber et~al.(2000)Treiber, Hennecke and
  Helbing}]{treiber2000congested}
\bibinfo{author}{Treiber, M.}, \bibinfo{author}{Hennecke, A.},
  \bibinfo{author}{Helbing, D.}, \bibinfo{year}{2000}.
\newblock \bibinfo{title}{Congested traffic states in empirical observations
  and microscopic simulations}.
\newblock \bibinfo{journal}{Physical review E} \bibinfo{volume}{62},
  \bibinfo{pages}{1805}.
\bibitem[{Van~Arem et~al.(2006)Van~Arem, Van~Driel and Visser}]{van2006impact}
\bibinfo{author}{Van~Arem, B.}, \bibinfo{author}{Van~Driel, C.J.},
  \bibinfo{author}{Visser, R.}, \bibinfo{year}{2006}.
\newblock \bibinfo{title}{The impact of cooperative adaptive cruise control on
  traffic-flow characteristics}.
\newblock \bibinfo{journal}{IEEE Transactions on Intelligent Transportation
  Systems} \bibinfo{volume}{7}, \bibinfo{pages}{429--436}.
\bibitem[{VanderWerf et~al.(2001)VanderWerf, Shladover, Kourjanskaia, Miller
  and Krishnan}]{vanderwerf2001modeling}
\bibinfo{author}{VanderWerf, J.}, \bibinfo{author}{Shladover, S.},
  \bibinfo{author}{Kourjanskaia, N.}, \bibinfo{author}{Miller, M.},
  \bibinfo{author}{Krishnan, H.}, \bibinfo{year}{2001}.
\newblock \bibinfo{title}{Modeling effects of driver control assistance systems
  on traffic}.
\newblock \bibinfo{journal}{Transportation Research Record: Journal of the
  Transportation Research Board} , \bibinfo{pages}{167--174}.
\bibitem[{Venayagamoorthy and Doctor(2004)}]{venayagamoorthy2004unmanned}
\bibinfo{author}{Venayagamoorthy, G.}, \bibinfo{author}{Doctor, S.},
  \bibinfo{year}{2004}.
\newblock \bibinfo{title}{Unmanned vehicle navigation using swarm
  intelligence}, in: \bibinfo{booktitle}{International Conference on
  Intelligent Sensing and Information Processing, 2004. Proceedings of},
  \bibinfo{organization}{IEEE}. pp. \bibinfo{pages}{249--253}.
\bibitem[{Vinitsky et~al.(2018)Vinitsky, Kreidieh, Le~Flem, Kheterpal, Jang,
  Wu, Wu, Liaw, Liang and Bayen}]{CWu-2018}
\bibinfo{author}{Vinitsky, E.}, \bibinfo{author}{Kreidieh, A.},
  \bibinfo{author}{Le~Flem, L.}, \bibinfo{author}{Kheterpal, N.},
  \bibinfo{author}{Jang, K.}, \bibinfo{author}{Wu, C.}, \bibinfo{author}{Wu,
  F.}, \bibinfo{author}{Liaw, R.}, \bibinfo{author}{Liang, E.},
  \bibinfo{author}{Bayen, A.M.}, \bibinfo{year}{2018}.
\newblock \bibinfo{title}{Benchmarks for reinforcement learning in
  mixed-autonomy traffic}, in: \bibinfo{booktitle}{Conference on Robot
  Learning}, pp. \bibinfo{pages}{399--409}.
\bibitem[{Vinkhuyzen and Cefkin(2016)}]{vinkhuyzen2016developing}
\bibinfo{author}{Vinkhuyzen, E.}, \bibinfo{author}{Cefkin, M.},
  \bibinfo{year}{2016}.
\newblock \bibinfo{title}{Developing socially acceptable autonomous vehicles},
  in: \bibinfo{booktitle}{Ethnographic Praxis in Industry Conference
  Proceedings}, \bibinfo{organization}{Wiley Online Library}. pp.
  \bibinfo{pages}{522--534}.
\bibitem[{Vinyals et~al.(2019)Vinyals, Babuschkin, Czarnecki, Mathieu, Dudzik,
  Chung, Choi, Powell, Ewalds, Georgiev, Oh, Horgan, Kroiss, Danihelka, Huang,
  Sifre, Cai, Agapiou, Jaderberg, Vezhnevets, Leblond, Pohlen, Dalibard,
  Budden, Sulsky, Molloy, Paine, Gulcehre, Wang, Pfaff, Wu, Ring, Yogatama,
  Wünsch, McKinney, Smith, Schaul, Lillicrap, Kavukcuoglu, Hassabis, Apps and
  Silver}]{vinyals_grandmaster_2019}
\bibinfo{author}{Vinyals, O.}, \bibinfo{author}{Babuschkin, I.},
  \bibinfo{author}{Czarnecki, W.M.}, \bibinfo{author}{Mathieu, M.},
  \bibinfo{author}{Dudzik, A.}, \bibinfo{author}{Chung, J.},
  \bibinfo{author}{Choi, D.H.}, \bibinfo{author}{Powell, R.},
  \bibinfo{author}{Ewalds, T.}, \bibinfo{author}{Georgiev, P.},
  \bibinfo{author}{Oh, J.}, \bibinfo{author}{Horgan, D.},
  \bibinfo{author}{Kroiss, M.}, \bibinfo{author}{Danihelka, I.},
  \bibinfo{author}{Huang, A.}, \bibinfo{author}{Sifre, L.},
  \bibinfo{author}{Cai, T.}, \bibinfo{author}{Agapiou, J.P.},
  \bibinfo{author}{Jaderberg, M.}, \bibinfo{author}{Vezhnevets, A.S.},
  \bibinfo{author}{Leblond, R.}, \bibinfo{author}{Pohlen, T.},
  \bibinfo{author}{Dalibard, V.}, \bibinfo{author}{Budden, D.},
  \bibinfo{author}{Sulsky, Y.}, \bibinfo{author}{Molloy, J.},
  \bibinfo{author}{Paine, T.L.}, \bibinfo{author}{Gulcehre, C.},
  \bibinfo{author}{Wang, Z.}, \bibinfo{author}{Pfaff, T.}, \bibinfo{author}{Wu,
  Y.}, \bibinfo{author}{Ring, R.}, \bibinfo{author}{Yogatama, D.},
  \bibinfo{author}{Wünsch, D.}, \bibinfo{author}{McKinney, K.},
  \bibinfo{author}{Smith, O.}, \bibinfo{author}{Schaul, T.},
  \bibinfo{author}{Lillicrap, T.}, \bibinfo{author}{Kavukcuoglu, K.},
  \bibinfo{author}{Hassabis, D.}, \bibinfo{author}{Apps, C.},
  \bibinfo{author}{Silver, D.}, \bibinfo{year}{2019}.
\newblock \bibinfo{title}{Grandmaster level in {StarCraft} {II} using
  multi-agent reinforcement learning}.
\newblock \bibinfo{journal}{Nature} \bibinfo{volume}{575},
  \bibinfo{pages}{350--354}.
\bibitem[{Wang(2018)}]{wang2018infrastructure}
\bibinfo{author}{Wang, M.}, \bibinfo{year}{2018}.
\newblock \bibinfo{title}{Infrastructure assisted adaptive driving to stabilise
  heterogeneous vehicle strings}.
\newblock \bibinfo{journal}{Transportation Research Part C: Emerging
  Technologies} \bibinfo{volume}{91}, \bibinfo{pages}{276--295}.
\bibitem[{Wang et~al.(2014a)Wang, Daamen, Hoogendoorn and van
  Arem}]{wang2014rolling1}
\bibinfo{author}{Wang, M.}, \bibinfo{author}{Daamen, W.},
  \bibinfo{author}{Hoogendoorn, S.P.}, \bibinfo{author}{van Arem, B.},
  \bibinfo{year}{2014}a.
\newblock \bibinfo{title}{Rolling horizon control framework for driver
  assistance systems. part i: Mathematical formulation and non-cooperative
  systems}.
\newblock \bibinfo{journal}{Transportation research part C: emerging
  technologies} \bibinfo{volume}{40}, \bibinfo{pages}{271--289}.
\bibitem[{Wang et~al.(2014b)Wang, Daamen, Hoogendoorn and van
  Arem}]{wang2014rolling2}
\bibinfo{author}{Wang, M.}, \bibinfo{author}{Daamen, W.},
  \bibinfo{author}{Hoogendoorn, S.P.}, \bibinfo{author}{van Arem, B.},
  \bibinfo{year}{2014}b.
\newblock \bibinfo{title}{Rolling horizon control framework for driver
  assistance systems. part ii: Cooperative sensing and cooperative control}.
\newblock \bibinfo{journal}{Transportation research part C: emerging
  technologies} \bibinfo{volume}{40}, \bibinfo{pages}{290--311}.
\bibitem[{Wang et~al.(2016)Wang, Daamen, Hoogendoorn and van
  Arem}]{wang2016cooperative}
\bibinfo{author}{Wang, M.}, \bibinfo{author}{Daamen, W.},
  \bibinfo{author}{Hoogendoorn, S.P.}, \bibinfo{author}{van Arem, B.},
  \bibinfo{year}{2016}.
\newblock \bibinfo{title}{Cooperative car-following control: Distributed
  algorithm and impact on moving jam features}.
\newblock \bibinfo{journal}{IEEE Transactions on Intelligent Transportation
  Systems} \bibinfo{volume}{17}, \bibinfo{pages}{1459--1471}.
\bibitem[{Wang et~al.(2015)Wang, Hoogendoorn, Daamen, van Arem and
  Happee}]{wang2015game}
\bibinfo{author}{Wang, M.}, \bibinfo{author}{Hoogendoorn, S.P.},
  \bibinfo{author}{Daamen, W.}, \bibinfo{author}{van Arem, B.},
  \bibinfo{author}{Happee, R.}, \bibinfo{year}{2015}.
\newblock \bibinfo{title}{Game theoretic approach for predictive lane-changing
  and car-following control}.
\newblock \bibinfo{journal}{Transportation Research Part C: Emerging
  Technologies} \bibinfo{volume}{58}, \bibinfo{pages}{73--92}.
\bibitem[{Wang et~al.(2019)Wang, Huang and Lo}]{Wang-2019}
\bibinfo{author}{Wang, S.}, \bibinfo{author}{Huang, W.}, \bibinfo{author}{Lo,
  H.K.}, \bibinfo{year}{2019}.
\newblock \bibinfo{title}{Traffic parameters estimation for signalized
  intersections based on combined shockwave analysis and {B}ayesian network}.
\newblock \bibinfo{journal}{Transportation Research Part C: Emerging
  Technologies} \bibinfo{volume}{104}, \bibinfo{pages}{22--37}.
\bibitem[{Wang et~al.(2018)Wang, Jiang, Li, Lin, Zheng and
  Wang}]{NN_SGD_wang2018capturing}
\bibinfo{author}{Wang, X.}, \bibinfo{author}{Jiang, R.}, \bibinfo{author}{Li,
  L.}, \bibinfo{author}{Lin, Y.}, \bibinfo{author}{Zheng, X.},
  \bibinfo{author}{Wang, F.Y.}, \bibinfo{year}{2018}.
\newblock \bibinfo{title}{Capturing car-following behaviors by deep learning}.
\newblock \bibinfo{journal}{IEEE Transactions on Intelligent Transportation
  Systems} \bibinfo{volume}{19}, \bibinfo{pages}{910--920}.
\bibitem[{Waymo(2019)}]{Waymo-2019}
\bibinfo{author}{Waymo}, \bibinfo{year}{2019}.
\newblock \bibinfo{title}{Waymo {O}pen {D}ataset: An autonomous driving
  dataset}.
\newblock \bibinfo{howpublished}{\url{https://waymo.com/open/data/}}.
\bibitem[{Wei et~al.(2017)Wei, Avc{\i}, Liu, Belezamo, Ayd{\i}n, Li and
  Zhou}]{wei2017dynamic}
\bibinfo{author}{Wei, Y.}, \bibinfo{author}{Avc{\i}, C.}, \bibinfo{author}{Liu,
  J.}, \bibinfo{author}{Belezamo, B.}, \bibinfo{author}{Ayd{\i}n, N.},
  \bibinfo{author}{Li, P.T.}, \bibinfo{author}{Zhou, X.}, \bibinfo{year}{2017}.
\newblock \bibinfo{title}{Dynamic programming-based multi-vehicle longitudinal
  trajectory optimization with simplified car following models}.
\newblock \bibinfo{journal}{Transportation research part B: methodological}
  \bibinfo{volume}{106}, \bibinfo{pages}{102--129}.
\bibitem[{Wei et~al.(2019)Wei, Wang, Hao and Barth}]{wei2019vision}
\bibinfo{author}{Wei, Z.}, \bibinfo{author}{Wang, C.}, \bibinfo{author}{Hao,
  P.}, \bibinfo{author}{Barth, M.J.}, \bibinfo{year}{2019}.
\newblock \bibinfo{title}{Vision-based lane-changing behavior detection using
  deep residual neural network}, in: \bibinfo{booktitle}{2019 IEEE Intelligent
  Transportation Systems Conference (ITSC)}, \bibinfo{organization}{IEEE}. pp.
  \bibinfo{pages}{3108--3113}.
\bibitem[{Wierstra et~al.(2010)Wierstra, Forster, Peters and
  Schmidhuber}]{Daan-2010}
\bibinfo{author}{Wierstra, D.}, \bibinfo{author}{Forster, A.},
  \bibinfo{author}{Peters, J.}, \bibinfo{author}{Schmidhuber, J.},
  \bibinfo{year}{2010}.
\newblock \bibinfo{title}{Recurrent policy gradients}.
\newblock \bibinfo{journal}{Logic Journal of the IGPL} \bibinfo{volume}{18},
  \bibinfo{pages}{620--634}.
\bibitem[{Wiest et~al.(2012)Wiest, H{\"o}ffken, Kre{\ss}el and
  Dietmayer}]{wiest2012probabilistic}
\bibinfo{author}{Wiest, J.}, \bibinfo{author}{H{\"o}ffken, M.},
  \bibinfo{author}{Kre{\ss}el, U.}, \bibinfo{author}{Dietmayer, K.},
  \bibinfo{year}{2012}.
\newblock \bibinfo{title}{Probabilistic trajectory prediction with gaussian
  mixture models}, in: \bibinfo{booktitle}{2012 IEEE Intelligent Vehicles
  Symposium}, \bibinfo{organization}{IEEE}. pp. \bibinfo{pages}{141--146}.
\bibitem[{Wolf(2016)}]{Wolf-2016}
\bibinfo{author}{Wolf, I.}, \bibinfo{year}{2016}.
\newblock \bibinfo{title}{The interaction between humans and autonomous
  agents}, in: \bibinfo{booktitle}{Autonomous Driving}.
  \bibinfo{publisher}{Springer}, pp. \bibinfo{pages}{103--124}.
\bibitem[{Woo et~al.(2017)Woo, Ji, Kono, Tamura, Kuroda, Sugano, Yamamoto,
  Yamashita and Asama}]{Woo-2017}
\bibinfo{author}{Woo, H.}, \bibinfo{author}{Ji, Y.}, \bibinfo{author}{Kono,
  H.}, \bibinfo{author}{Tamura, Y.}, \bibinfo{author}{Kuroda, Y.},
  \bibinfo{author}{Sugano, T.}, \bibinfo{author}{Yamamoto, Y.},
  \bibinfo{author}{Yamashita, A.}, \bibinfo{author}{Asama, H.},
  \bibinfo{year}{2017}.
\newblock \bibinfo{title}{Lane-change detection based on vehicle-trajectory
  prediction}.
\newblock \bibinfo{journal}{IEEE Robotics and Automation Letters}
  \bibinfo{volume}{2}, \bibinfo{pages}{1109--1116}.
\bibitem[{Wu(2018)}]{wu2018learning}
\bibinfo{author}{Wu, C.}, \bibinfo{year}{2018}.
\newblock \bibinfo{title}{Learning and Optimization for Mixed Autonomy
  Systems-A Mobility Context}.
\newblock Ph.D. thesis. UC Berkeley.
\bibitem[{Wu et~al.(2018a)Wu, Bayen and Mehta}]{wu2018stabilizing}
\bibinfo{author}{Wu, C.}, \bibinfo{author}{Bayen, A.M.},
  \bibinfo{author}{Mehta, A.}, \bibinfo{year}{2018}a.
\newblock \bibinfo{title}{Stabilizing traffic with autonomous vehicles}, in:
  \bibinfo{booktitle}{2018 IEEE International Conference on Robotics and
  Automation (ICRA)}, \bibinfo{organization}{IEEE}. pp. \bibinfo{pages}{1--7}.
\bibitem[{Wu et~al.(2018b)Wu, Bayen and Mehtan}]{Cathy-2018}
\bibinfo{author}{Wu, C.}, \bibinfo{author}{Bayen, A.M.},
  \bibinfo{author}{Mehtan, A.}, \bibinfo{year}{2018}b.
\newblock \bibinfo{title}{Stabilizing traffic with autonomous vehicles}, in:
  \bibinfo{booktitle}{2018 IEEE International Conference on Robotics and
  Automation (ICRA)}, pp. \bibinfo{pages}{1--7}.
\bibitem[{Wu et~al.(2017a)Wu, Kreidieh, Parvate, Vinitsky and
  Bayen}]{wu2017flow}
\bibinfo{author}{Wu, C.}, \bibinfo{author}{Kreidieh, A.},
  \bibinfo{author}{Parvate, K.}, \bibinfo{author}{Vinitsky, E.},
  \bibinfo{author}{Bayen, A.M.}, \bibinfo{year}{2017}a.
\newblock \bibinfo{title}{Flow: Architecture and benchmarking for reinforcement
  learning in traffic control}.
\newblock \bibinfo{journal}{arXiv preprint arXiv:1710.05465} .
\bibitem[{Wu et~al.(2017b)Wu, Kreidieh, Vinitsky and Bayen}]{wu2017emergent}
\bibinfo{author}{Wu, C.}, \bibinfo{author}{Kreidieh, A.},
  \bibinfo{author}{Vinitsky, E.}, \bibinfo{author}{Bayen, A.M.},
  \bibinfo{year}{2017}b.
\newblock \bibinfo{title}{Emergent behaviors in mixed-autonomy traffic}, in:
  \bibinfo{booktitle}{Conference on Robot Learning}, pp.
  \bibinfo{pages}{398--407}.
\bibitem[{Wu et~al.(2017c)Wu, Kreidieh, Parvate, Vinitsky and Bayen}]{Wu-2017}
\bibinfo{author}{Wu, C.}, \bibinfo{author}{Kreidieh, A.R.},
  \bibinfo{author}{Parvate, K.}, \bibinfo{author}{Vinitsky, E.},
  \bibinfo{author}{Bayen, A.M.}, \bibinfo{year}{2017}c.
\newblock \bibinfo{title}{Flow: A modular learning framework for autonomy in
  traffic}, in: \bibinfo{booktitle}{arXiv preprint arXiv:1710.05465}.
\bibitem[{Wu et~al.(2017d)Wu, Parvate, Kheterpal, Dickstein, Mehta, Vinitsky
  and Bayen}]{wu2017framework}
\bibinfo{author}{Wu, C.}, \bibinfo{author}{Parvate, K.},
  \bibinfo{author}{Kheterpal, N.}, \bibinfo{author}{Dickstein, L.},
  \bibinfo{author}{Mehta, A.}, \bibinfo{author}{Vinitsky, E.},
  \bibinfo{author}{Bayen, A.M.}, \bibinfo{year}{2017}d.
\newblock \bibinfo{title}{Framework for control and deep reinforcement learning
  in traffic}, in: \bibinfo{booktitle}{Intelligent Transportation Systems
  (ITSC), 2017 IEEE 20th International Conference on},
  \bibinfo{organization}{IEEE}. pp. \bibinfo{pages}{1--8}.
\bibitem[{Wu and Work(2018)}]{Wu-2018}
\bibinfo{author}{Wu, F.}, \bibinfo{author}{Work, D.B.}, \bibinfo{year}{2018}.
\newblock \bibinfo{title}{Connections between classical car following models
  and artificial neural networks}, in: \bibinfo{booktitle}{Proceedings of the
  21st International Conference on Intelligent Transportation Systems (ITSC))},
  \bibinfo{address}{Maui, HI, USA}. pp. \bibinfo{pages}{3191--3198}.
\bibitem[{Wymann et~al.(2000)Wymann, Espie, Guionneau, Dimitrakakis, Coulom and
  Sumner}]{wymann}
\bibinfo{author}{Wymann, B.}, \bibinfo{author}{Espie, E.},
  \bibinfo{author}{Guionneau, C.}, \bibinfo{author}{Dimitrakakis, C.},
  \bibinfo{author}{Coulom, R.}, \bibinfo{author}{Sumner, A.},
  \bibinfo{year}{2000}.
\newblock \bibinfo{title}{{TORCS}, the open racing car simulator}.
\newblock \bibinfo{howpublished}{Software available at http://torcs.
  sourceforge.net}.
\bibitem[{Xu et~al.(2017)Xu, Gao, Yu and Darrell}]{Xu-2017}
\bibinfo{author}{Xu, H.}, \bibinfo{author}{Gao, Y.}, \bibinfo{author}{Yu, F.},
  \bibinfo{author}{Darrell, T.}, \bibinfo{year}{2017}.
\newblock \bibinfo{title}{End-to-end learning of driving models from
  large-scale video datasets}, in: \bibinfo{booktitle}{Proceedings of the IEEE
  conference on computer vision and pattern recognition (CVPR)}, pp.
  \bibinfo{pages}{2174--2182}.
\bibitem[{Yang et~al.(2017)Yang, Wang, Liu, Deng and Hedrick}]{yang2017feature}
\bibinfo{author}{Yang, S.}, \bibinfo{author}{Wang, W.}, \bibinfo{author}{Liu,
  C.}, \bibinfo{author}{Deng, W.}, \bibinfo{author}{Hedrick, J.K.},
  \bibinfo{year}{2017}.
\newblock \bibinfo{title}{Feature analysis and selection for training an
  end-to-end autonomous vehicle controller using deep learning approach}, in:
  \bibinfo{booktitle}{2017 IEEE Intelligent Vehicles Symposium (IV)},
  \bibinfo{organization}{IEEE}. pp. \bibinfo{pages}{1033--1038}.
\bibitem[{Yang et~al.(2018)Yang, Luo, Li, Zhou, Zhang and Wang}]{Yaodong-2018}
\bibinfo{author}{Yang, Y.}, \bibinfo{author}{Luo, R.}, \bibinfo{author}{Li,
  M.}, \bibinfo{author}{Zhou, M.}, \bibinfo{author}{Zhang, W.},
  \bibinfo{author}{Wang, J.}, \bibinfo{year}{2018}.
\newblock \bibinfo{title}{Lenient multi-agent deep reinforcement learning}, in:
  \bibinfo{booktitle}{the International Conference on Machine Learning}, pp.
  \bibinfo{pages}{5567--5576}.
\bibitem[{Yang and Perdikaris(2019)}]{Yang-2019}
\bibinfo{author}{Yang, Y.}, \bibinfo{author}{Perdikaris, P.},
  \bibinfo{year}{2019}.
\newblock \bibinfo{title}{Adversarial uncertainty quantification in
  physics-informed neural networks}.
\newblock \bibinfo{journal}{Journal of Computational Physics}
  \bibinfo{volume}{394}, \bibinfo{pages}{136--152}.
\bibitem[{Yao et~al.(2018)Yao, Cui, Li, Wang and An}]{yao2018trajectory}
\bibinfo{author}{Yao, H.}, \bibinfo{author}{Cui, J.}, \bibinfo{author}{Li, X.},
  \bibinfo{author}{Wang, Y.}, \bibinfo{author}{An, S.}, \bibinfo{year}{2018}.
\newblock \bibinfo{title}{A trajectory smoothing method at signalized
  intersection based on individualized variable speed limits with location
  optimization}.
\newblock \bibinfo{journal}{Transportation Research Part D: Transport and
  Environment} \bibinfo{volume}{62}, \bibinfo{pages}{456--473}.
\bibitem[{Yao et~al.(2019)Yao, Hu, Wang, Jiang, Ran and
  Chen}]{yao2019stability}
\bibinfo{author}{Yao, Z.}, \bibinfo{author}{Hu, R.}, \bibinfo{author}{Wang,
  Y.}, \bibinfo{author}{Jiang, Y.}, \bibinfo{author}{Ran, B.},
  \bibinfo{author}{Chen, Y.}, \bibinfo{year}{2019}.
\newblock \bibinfo{title}{Stability analysis and the fundamental diagram for
  mixed connected automated and human-driven vehicles}.
\newblock \bibinfo{journal}{Physica A: Statistical Mechanics and its
  Applications} \bibinfo{volume}{533}, \bibinfo{pages}{121931}.
\bibitem[{Yoo and Langari(2020)}]{yoo2020game}
\bibinfo{author}{Yoo, J.}, \bibinfo{author}{Langari, R.}, \bibinfo{year}{2020}.
\newblock \bibinfo{title}{A game-theoretic model of human driving and
  application to discretionary lane-changes}.
\newblock \bibinfo{journal}{arXiv preprint arXiv:2003.09783} .
\bibitem[{Yoo and Langari(2012)}]{yoo2012stackelberg}
\bibinfo{author}{Yoo, J.H.}, \bibinfo{author}{Langari, R.},
  \bibinfo{year}{2012}.
\newblock \bibinfo{title}{Stackelberg game based model of highway driving}, in:
  \bibinfo{booktitle}{ASME 2012 5th Annual Dynamic Systems and Control
  Conference joint with the JSME 2012 11th Motion and Vibration Conference},
  \bibinfo{organization}{American Society of Mechanical Engineers}. pp.
  \bibinfo{pages}{499--508}.
\bibitem[{Yoo and Langari(2013)}]{yoo2013stackelberg}
\bibinfo{author}{Yoo, J.H.}, \bibinfo{author}{Langari, R.},
  \bibinfo{year}{2013}.
\newblock \bibinfo{title}{A stackelberg game theoretic driver model for
  merging}, in: \bibinfo{booktitle}{ASME 2013 Dynamic Systems and Control
  Conference}, \bibinfo{organization}{American Society of Mechanical
  Engineers}. pp. \bibinfo{pages}{V002T30A003--V002T30A003}.
\bibitem[{Yu et~al.(2018)Yu, Tseng and Langari}]{yu2018human}
\bibinfo{author}{Yu, H.}, \bibinfo{author}{Tseng, H.E.},
  \bibinfo{author}{Langari, R.}, \bibinfo{year}{2018}.
\newblock \bibinfo{title}{A human-like game theory-based controller for
  automatic lane changing}.
\newblock \bibinfo{journal}{Transportation Research Part C: Emerging
  Technologies} \bibinfo{volume}{88}, \bibinfo{pages}{140--158}.
\bibitem[{Zegers et~al.(2017)Zegers, Semsar-Kazerooni, Ploeg, van~de Wouw and
  Nijmeijer}]{zegers2017consensus}
\bibinfo{author}{Zegers, J.C.}, \bibinfo{author}{Semsar-Kazerooni, E.},
  \bibinfo{author}{Ploeg, J.}, \bibinfo{author}{van~de Wouw, N.},
  \bibinfo{author}{Nijmeijer, H.}, \bibinfo{year}{2017}.
\newblock \bibinfo{title}{Consensus control for vehicular platooning with
  velocity constraints}.
\newblock \bibinfo{journal}{IEEE Transactions on Control Systems Technology}
  \bibinfo{volume}{26}, \bibinfo{pages}{1592--1605}.
\bibitem[{Zhang et~al.(2017)Zhang, Vinkhuyzen and Cefkin}]{zhang2017evaluation}
\bibinfo{author}{Zhang, J.}, \bibinfo{author}{Vinkhuyzen, E.},
  \bibinfo{author}{Cefkin, M.}, \bibinfo{year}{2017}.
\newblock \bibinfo{title}{Evaluation of an autonomous vehicle external
  communication system concept: a survey study}, in:
  \bibinfo{booktitle}{International conference on applied human factors and
  ergonomics}, \bibinfo{organization}{Springer}. pp. \bibinfo{pages}{650--661}.
\bibitem[{Zhang et~al.(2019a)Zhang, Yang and Başar}]{zhang_multi-agent_2019}
\bibinfo{author}{Zhang, K.}, \bibinfo{author}{Yang, Z.},
  \bibinfo{author}{Başar, T.}, \bibinfo{year}{2019}a.
\newblock \bibinfo{title}{Multi-{Agent} {Reinforcement} {Learning}: {A}
  {Selective} {Overview} of {Theories} and {Algorithms}} \URLprefix
  \url{https://arxiv.org/abs/1911.10635v1}.
\bibitem[{Zhang et~al.(2019b)Zhang, Langari, Tseng, Filev, Szwabowski and
  Coskun}]{zhang2019game}
\bibinfo{author}{Zhang, Q.}, \bibinfo{author}{Langari, R.},
  \bibinfo{author}{Tseng, H.E.}, \bibinfo{author}{Filev, D.P.},
  \bibinfo{author}{Szwabowski, S.}, \bibinfo{author}{Coskun, S.},
  \bibinfo{year}{2019}b.
\newblock \bibinfo{title}{A game theoretic model predictive controller with
  aggressiveness estimation for mandatory lane change}.
\newblock \bibinfo{journal}{IEEE Transactions on Intelligent Vehicles} .
\bibitem[{Zhang et~al.(2016)Zhang, Kahn, Levine and Abbeel}]{Tianhao-2016}
\bibinfo{author}{Zhang, T.}, \bibinfo{author}{Kahn, G.},
  \bibinfo{author}{Levine, S.}, \bibinfo{author}{Abbeel, P.},
  \bibinfo{year}{2016}.
\newblock \bibinfo{title}{Learning deep control policies for autonomous aerial
  vehicles with {MPC}-guided policy search}, in: \bibinfo{booktitle}{2016 IEEE
  international conference on robotics and automation (ICRA)}, pp.
  \bibinfo{pages}{528--535}.
\bibitem[{Zhang et~al.(2018a)Zhang, Zheng, Wang and
  Fan}]{onboard_lidar_zhang2018background}
\bibinfo{author}{Zhang, Z.Y.}, \bibinfo{author}{Zheng, J.},
  \bibinfo{author}{Wang, X.}, \bibinfo{author}{Fan, X.}, \bibinfo{year}{2018}a.
\newblock \bibinfo{title}{Background filtering and vehicle detection with
  roadside lidar based on point association}, in: \bibinfo{booktitle}{2018 37th
  Chinese Control Conference (CCC)}, \bibinfo{organization}{IEEE}. pp.
  \bibinfo{pages}{7938--7943}.
\bibitem[{Zhang et~al.(2018b)Zhang, Zheng, Wang and
  Fan}]{lidar_zhang2018background}
\bibinfo{author}{Zhang, Z.Y.}, \bibinfo{author}{Zheng, J.},
  \bibinfo{author}{Wang, X.}, \bibinfo{author}{Fan, X.}, \bibinfo{year}{2018}b.
\newblock \bibinfo{title}{Background filtering and vehicle detection with
  roadside lidar based on point association}, in: \bibinfo{booktitle}{2018 37th
  Chinese Control Conference (CCC)}, \bibinfo{organization}{IEEE}. pp.
  \bibinfo{pages}{7938--7943}.
\bibitem[{Zhao et~al.(2020)Zhao, Wang, Xu, Wang, Li and Qu}]{zhao2020field}
\bibinfo{author}{Zhao, X.}, \bibinfo{author}{Wang, Z.}, \bibinfo{author}{Xu,
  Z.}, \bibinfo{author}{Wang, Y.}, \bibinfo{author}{Li, X.},
  \bibinfo{author}{Qu, X.}, \bibinfo{year}{2020}.
\newblock \bibinfo{title}{Field experiments on longitudinal characteristics of
  human driver behavior following an autonomous vehicle}.
\newblock \bibinfo{journal}{Transportation Research Part C: Emerging
  Technologies} \bibinfo{volume}{114}, \bibinfo{pages}{205--224}.
\bibitem[{Zheng et~al.(2016)Zheng, Li, Li and Wang}]{zheng2016stability}
\bibinfo{author}{Zheng, Y.}, \bibinfo{author}{Li, S.E.}, \bibinfo{author}{Li,
  K.}, \bibinfo{author}{Wang, L.Y.}, \bibinfo{year}{2016}.
\newblock \bibinfo{title}{Stability margin improvement of vehicular platoon
  considering undirected topology and asymmetric control}.
\newblock \bibinfo{journal}{IEEE Transactions on Control Systems Technology}
  \bibinfo{volume}{24}, \bibinfo{pages}{1253--1265}.
\bibitem[{Zhou and Laval(2019)}]{zhou2019longitudinal}
\bibinfo{author}{Zhou, H.}, \bibinfo{author}{Laval, J.}, \bibinfo{year}{2019}.
\newblock \bibinfo{title}{Longitudinal motion planning for autonomous vehicles
  and its impact on congestion: A survey}.
\newblock \bibinfo{journal}{arXiv preprint arXiv:1910.06070} .
\bibitem[{Zhou et~al.(2017a)Zhou, Qu and Li}]{RNN_zhou2017recurrent}
\bibinfo{author}{Zhou, M.}, \bibinfo{author}{Qu, X.}, \bibinfo{author}{Li, X.},
  \bibinfo{year}{2017}a.
\newblock \bibinfo{title}{A recurrent neural network based microscopic car
  following model to predict traffic oscillation}.
\newblock \bibinfo{journal}{Transportation research part C: emerging
  technologies} \bibinfo{volume}{84}, \bibinfo{pages}{245--264}.
\bibitem[{Zhou and Ahn(2019)}]{zhou2019robust}
\bibinfo{author}{Zhou, Y.}, \bibinfo{author}{Ahn, S.}, \bibinfo{year}{2019}.
\newblock \bibinfo{title}{Robust local and string stability for a decentralized
  car following control strategy for connected automated vehicles}.
\newblock \bibinfo{journal}{Transportation Research Part B: Methodological}
  \bibinfo{volume}{125}, \bibinfo{pages}{175--196}.
\bibitem[{Zhou et~al.(2017b)Zhou, Ahn, Chitturi and Noyce}]{zhou2017rolling}
\bibinfo{author}{Zhou, Y.}, \bibinfo{author}{Ahn, S.},
  \bibinfo{author}{Chitturi, M.}, \bibinfo{author}{Noyce, D.A.},
  \bibinfo{year}{2017}b.
\newblock \bibinfo{title}{Rolling horizon stochastic optimal control strategy
  for acc and cacc under uncertainty}.
\newblock \bibinfo{journal}{Transportation Research Part C: Emerging
  Technologies} \bibinfo{volume}{83}, \bibinfo{pages}{61--76}.
\bibitem[{Zhou et~al.(2017c)Zhou, Ahn, Chitturi and Noyce}]{zhou-2017}
\bibinfo{author}{Zhou, Y.}, \bibinfo{author}{Ahn, S.},
  \bibinfo{author}{Chitturi, M.}, \bibinfo{author}{Noyce, D.A.},
  \bibinfo{year}{2017}c.
\newblock \bibinfo{title}{Rolling horizon stochastic optimal control strategy
  for acc and cacc under uncertainty}.
\newblock \bibinfo{journal}{Transportation Research Part C: Emerging
  Technologies} \bibinfo{volume}{83}, \bibinfo{pages}{61--76}.
\bibitem[{Zhou et~al.(2020)Zhou, Ahn, Wang and
  Hoogendoorn}]{zhou2020stabilizing}
\bibinfo{author}{Zhou, Y.}, \bibinfo{author}{Ahn, S.}, \bibinfo{author}{Wang,
  M.}, \bibinfo{author}{Hoogendoorn, S.}, \bibinfo{year}{2020}.
\newblock \bibinfo{title}{Stabilizing mixed vehicular platoons with connected
  automated vehicles: An h-infinity approach}.
\newblock \bibinfo{journal}{Transportation Research Part B: Methodological}
  \bibinfo{volume}{132}, \bibinfo{pages}{152--170}.
\bibitem[{Zhou et~al.(2019)Zhou, Wang and Ahn}]{zhou2019distributed}
\bibinfo{author}{Zhou, Y.}, \bibinfo{author}{Wang, M.}, \bibinfo{author}{Ahn,
  S.}, \bibinfo{year}{2019}.
\newblock \bibinfo{title}{Distributed model predictive control approach for
  cooperative car-following with guaranteed local and string stability}.
\newblock \bibinfo{journal}{Transportation research part B: methodological}
  \bibinfo{volume}{128}, \bibinfo{pages}{69--86}.
\bibitem[{Zhu et~al.(2018a)Zhu, Wang, Tarko et~al.}]{onboard_zhu2018modeling}
\bibinfo{author}{Zhu, M.}, \bibinfo{author}{Wang, X.}, \bibinfo{author}{Tarko,
  A.}, et~al., \bibinfo{year}{2018}a.
\newblock \bibinfo{title}{Modeling car-following behavior on urban expressways
  in shanghai: A naturalistic driving study}.
\newblock \bibinfo{journal}{Transportation research part C: emerging
  technologies} \bibinfo{volume}{93}, \bibinfo{pages}{425--445}.
\bibitem[{Zhu et~al.(2018b)Zhu, Wang and Wang}]{NN_DRL_zhu2018human}
\bibinfo{author}{Zhu, M.}, \bibinfo{author}{Wang, X.}, \bibinfo{author}{Wang,
  Y.}, \bibinfo{year}{2018}b.
\newblock \bibinfo{title}{Human-like autonomous car-following planning by deep
  reinforcement learning}.
\newblock \bibinfo{type}{Technical Report}.
\bibitem[{Ziebart et~al.(2008)Ziebart, Maas, Bagnell and
  Dey}]{ziebart2008maximum}
\bibinfo{author}{Ziebart, B.D.}, \bibinfo{author}{Maas, A.L.},
  \bibinfo{author}{Bagnell, J.A.}, \bibinfo{author}{Dey, A.K.},
  \bibinfo{year}{2008}.
\newblock \bibinfo{title}{Maximum entropy inverse reinforcement learning.}, in:
  \bibinfo{booktitle}{AAAI}, \bibinfo{organization}{Chicago, IL, USA}. pp.
  \bibinfo{pages}{1433--1438}.

\end{thebibliography}







\end{document}